\documentclass[11pt]{article}

\usepackage[preprint]{acl}

\usepackage{times}
\usepackage{latexsym}
\usepackage{fontawesome5}
\usepackage{hyperref}

\usepackage[T1]{fontenc}
\usepackage{graphicx}
\usepackage{booktabs}
\usepackage{multirow}

\usepackage[utf8]{inputenc}

\usepackage{microtype}
\usepackage{booktabs}
\usepackage{multirow}
\usepackage{makecell}
\usepackage[table]{xcolor}
\usepackage{graphicx}

\usepackage{inconsolata}

\usepackage{graphicx}
\usepackage{amsmath}
\usepackage{amssymb}

\usepackage[most]{tcolorbox}

\definecolor{projectblue}{HTML}{2563A6}
\usepackage{fvextra}
\newtcolorbox{appendixprompt}[1]{%
  enhanced standard,breakable,colback=black!3,colframe=black!65,
  colbacktitle=black!72,coltitle=white,
  fonttitle=\bfseries\footnotesize,
  fontupper=\fontsize{9}{10.5}\selectfont\raggedright,
  title={#1},title after break={#1 (continued)},
  boxrule=0.6pt,arc=2pt,left=7pt,right=7pt,top=5pt,bottom=5pt,
  before skip=6pt,after skip=6pt,pad at break=1mm,
  toprule at break=0.6pt,bottomrule at break=0.6pt}

\title{Catalyzing Interactive HTML Generation Capabilities \\ via High-Fidelity Reinforcement Learning}

\author{
\textbf{Xujia Wang}\textsuperscript{1,2,*} \quad
\textbf{Yixuan Liu}\textsuperscript{1,*} \quad
\textbf{Shaohan Huang}\textsuperscript{2,$\dagger$} \quad
\textbf{Furu Wei}\textsuperscript{2} \\
\textbf{Lei Hou}\textsuperscript{1} \quad
\textbf{Bin Xu}\textsuperscript{1,$\dagger$} \quad
\textbf{Juanzi Li}\textsuperscript{1}
\\[0.6em]
\textsuperscript{1}Tsinghua University \quad
\textsuperscript{2}Microsoft Research
\\[0.3em]
\texttt{wang-xj22@mails.tsinghua.edu.cn}
\\[0.2em]
{
\textsuperscript{*}Equal contribution. \quad
\textsuperscript{$\dagger$}Corresponding authors.
}
}

\begin{document}
\maketitle
\begin{abstract}
Generating HTML from natural-language specifications is an important application of code large language models (LLMs). Despite producing renderable pages, current models often fail to ensure functionality and visual quality during actual interaction. Existing methods typically target task-specific webpage supervision, leaving demanding interactive artifact optimization underexplored. Using mini-games as a challenging testbed, we expose limitations in interaction experience, reward signal, and credit assignment fidelity that are obscured by shallow webpage interactions. We introduce
\href{https://klozewang.github.io/webcatalyst-project-page/#overview}
{WebCatalyst}, a browser-interaction reinforcement learning framework with three complementary components: a game-level GUI assessor for sustained interaction and evaluation, Visual Anchor for mitigating VLM reward hacking, and Action-Trace Credit for grounding token-level credit assignment in runtime execution traces. Trained on fewer than 1K requests, predominantly for games, WebCatalyst raises the Cookie-Bench Mini Games score from 68.8 to 75.3 and also the WebGen-Bench accuracy from 48.8 to 56.9 on Qwen3.6-35B-A3B, outperforming Claude Opus 4.7 in average rank across evaluated metrics. Further analysis shows that Visual Anchor prevents aesthetic degradation, while runtime-grounded credit assignment accelerates functional improvement. Together, our results show that improving feedback fidelity is the key to substantially advancing interactive HTML generation.
\end{abstract}

\section{Introduction}


With advances in code generation, large language models (LLMs) are widely used to generate HTML webpages from natural-language requests or visual designs \citep{si2024design2code,laurencon2024websight,yun2024web2code,gui2025webcode2m,lu2025webgenbench}. Although large-scale pretraining and supervised fine-tuning enable models to produce syntactically valid HTML \citep{laurencon2024websight,yun2024web2code,gui2025webcode2m}, successful rendering does not guarantee usability: page quality depends on how it looks and behaves during interaction \citep{xiao2025interaction2code,lu2025webgenbench,zhang2025artifactsbench}.



\begin{figure}[t]   
  \centering           
  \includegraphics[width=\linewidth]{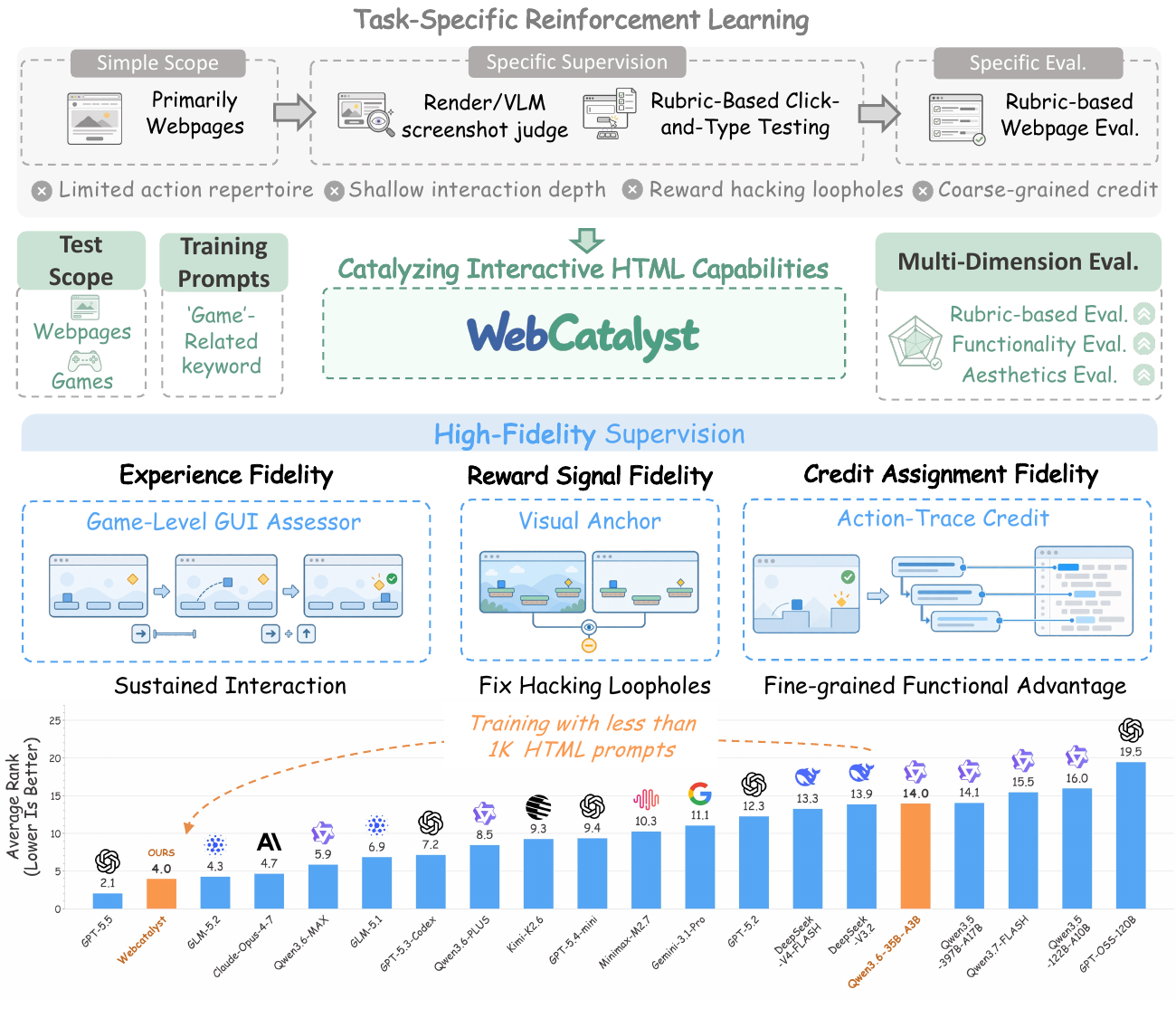} 
  \vspace{-25pt}
  \caption{From task-specific webpage optimization to interactive HTML capability catalysis. WebCatalyst improves fidelity in interaction experience, reward signals, and credit assignment through a Game-Level GUI Assessor, Visual Anchor, and Action-Trace Credit.}
  \label{fig:intro}
  \vspace{-15pt}
\end{figure}

To incorporate these aspects of usability into training, recent approaches developed reinforcement learning (RL) supervision for webpage generation, including executable rubrics \citep{jin2026rcca}, iterative-refinement feedback \citep{lu2025webgenagent,li2026relook,deng2026visrefiner}, GUI-agent functional checks \citep{lu2025webgenagent}, visual-quality rewards \citep{lai2025webrenderbench,li2026relook,jiang2026webgenr1}, and evolving graders \citep{chen2026webgrader}.
While these methods demonstrate the value of task-aligned feedback, their supervision is often tailored to specific webpage tasks and evaluation protocols.

Thus, we seek to move beyond task-specific webpage optimization \textbf{toward catalyzing more demanding interactive HTML capabilities}. This raises a central question: \textit{What design principles enable effective and more generalizable RL supervision for interactive HTML generation?} Alongside conventional webpages with basic click-and-type interactions, we study mini-games as a challenging testbed, whose sustained interactions and dynamic visuals expose fidelity limitations in interaction experience, reward signals, and credit assignment.


To this end, we introduce \textbf{WebCatalyst}, a browser-interaction RL framework for improving the interactive HTML generation capabilities of LLMs. Its high-fidelity supervision addresses three coupled challenges in translating browser interactions into effective learning signals. First, evaluating interactive HTML requires evidence from actual interaction \citep{zhang2026miniappbench,yang2026cookiebench}. Our \textbf{GUI Assessor} (\S \ref{sec:guiassessor}) conducts sustained, game-level browser interactions and hierarchically aggregates action outcomes and screenshot-based evaluations into a page-level reward. Second, we identify a reward-hacking failure mode in GUI-agent evaluation: VLM judges may overlook visual degradation when functionality improves. To mitigate this bias, the \textbf{Visual Anchor} (\S \ref{sec:visualanchor}) penalizes degradation relative to reference screenshots. Third, coarse page-level rewards provide little guidance about which code tokens contribute to individual behaviors in complex interactive environments \citep{dou2024stepcoder,kumar2026egca,jin2026rcca}. We therefore introduce \textbf{Action-Trace Credit} (\S \ref{sec:creditassign}), a runtime-grounded credit-assignment method that captures call stacks for successful actions, maps executed code spans back to generated tokens, and redistributes advantages according to call depth. These components improve interaction experience, reward signal, and credit assignment fidelity enabling more effective training for complex interactive HTML generation.


With fewer than 1K training requests, most involving games with complex interactions, WebCatalyst improves the Cookie-Bench Mini Games overall score from \textbf{68.8 to 75.3} and WebGen-Bench accuracy from \textbf{48.8 to 56.9} on Qwen3.6-35B-A3B. It also achieves a better average rank than Claude Opus 4.7 across webpage and mini-game evaluations. Further analysis reveals that GUI feedback alone improves functionality at the cost of visual quality. Visual Anchor mitigates this degradation, while Action-Trace Credit accelerates gains on functional metrics. By exposing limitations masked by shallow webpage interactions, these results show how high-fidelity feedback can move RL beyond task-specific optimization toward more capable interactive HTML generation.


Our contributions are:
\begin{itemize}
    \setlength{\itemsep}{1pt}
    \setlength{\parsep}{0pt}
    \item We study interactive HTML generation beyond task-specific webpage optimization, using challenging mini-games to reveal three key dimensions of RL supervision fidelity: interaction experience, reward signals, and credit assignment.

    \item We introduce \textbf{WebCatalyst}, a unified browser-interaction RL framework that addresses these limitations through three complementary components: a \textbf{Game-Level GUI Assessor}, \textbf{Visual Anchor}, and \textbf{Action-Trace Credit}.

    \item With fewer than 1K training requests, WebCatalyst improves both game and webpage generation, enabling Qwen3.6-35B-A3B to achieve a better average rank than Claude Opus 4.7 across evaluated metrics.
\end{itemize}

\section{Related Work}




\subsection{Assessing Interactive HTML Artifacts}

Evaluating interactive HTML requires assessing both functional behavior and visual quality during use.
GUI agents provide grounded mouse-and-keyboard interaction \citep{cheng2024seeclick,he2024webvoyager,qin2025uitars}, but existing browser and desktop environments mainly evaluate agents in pre-existing applications \citep{zhou2024webarena,drouin2024workarena,xie2024osworld}. For generated HTML, interaction-aware benchmarks evaluate behavior, functionality, and visual quality \citep{xiao2025interaction2code,lu2025webgenbench,zhang2025artifactsbench}.
Recent exploratory evaluators gather visual and behavioral evidence over interaction trajectories \citep{zhang2026miniappbench,yang2026cookiebench}, while game-oriented evaluators assess runtime gameplay and sometimes support iterative code refinement \citep{zhang2026webgamebench,huang2026continualgamegeneration,peng2026playcoder}. These works use interactive assessors primarily for benchmarking or inference-time refinement, rather than as feedback providers for RL.

\subsection{Training for HTML Generation}

HTML generation covers design reconstruction and web application generation from natural-language requests.
Supervised approaches learn from screenshot--code pairs \citep{beltramelli2017pix2code,laurencon2024websight,yun2024web2code,gui2025webcode2m}.
Browser-guided refinement also supplies SFT data \citep{lu2025webgenagent,wu2026htmlcure}.
Recent RL methods primarily target webpage generation, using feedback aligning with task-specific evaluation protocols \citep{lai2025webrenderbench,li2026relook,jiang2026webgenr1,lu2025webgenagent,chen2026webgrader}, whereas we focus on RL supervision fidelity in sustained HTML interaction.

\subsection{Code Credit Assignment}

Sequence-level rewards obscure the contributions of individual code regions.
Prior work localizes feedback using learned reward models \citep{le2022coderl,fan2026recode}, compiler-reported errors \citep{liu2023rltf}, execution coverage \citep{dou2024stepcoder,kumar2026egca}, or a single related code span \citep{li2026boostapr,jin2026rcca}.
Our \textbf{Action-Trace Credit} enables precise behavior-to-code reward assignment by tracing each successful GUI action to its runtime code path.

\section{Method}

\begin{figure*}[htb]
    \centering
    \includegraphics[width=0.95\linewidth]{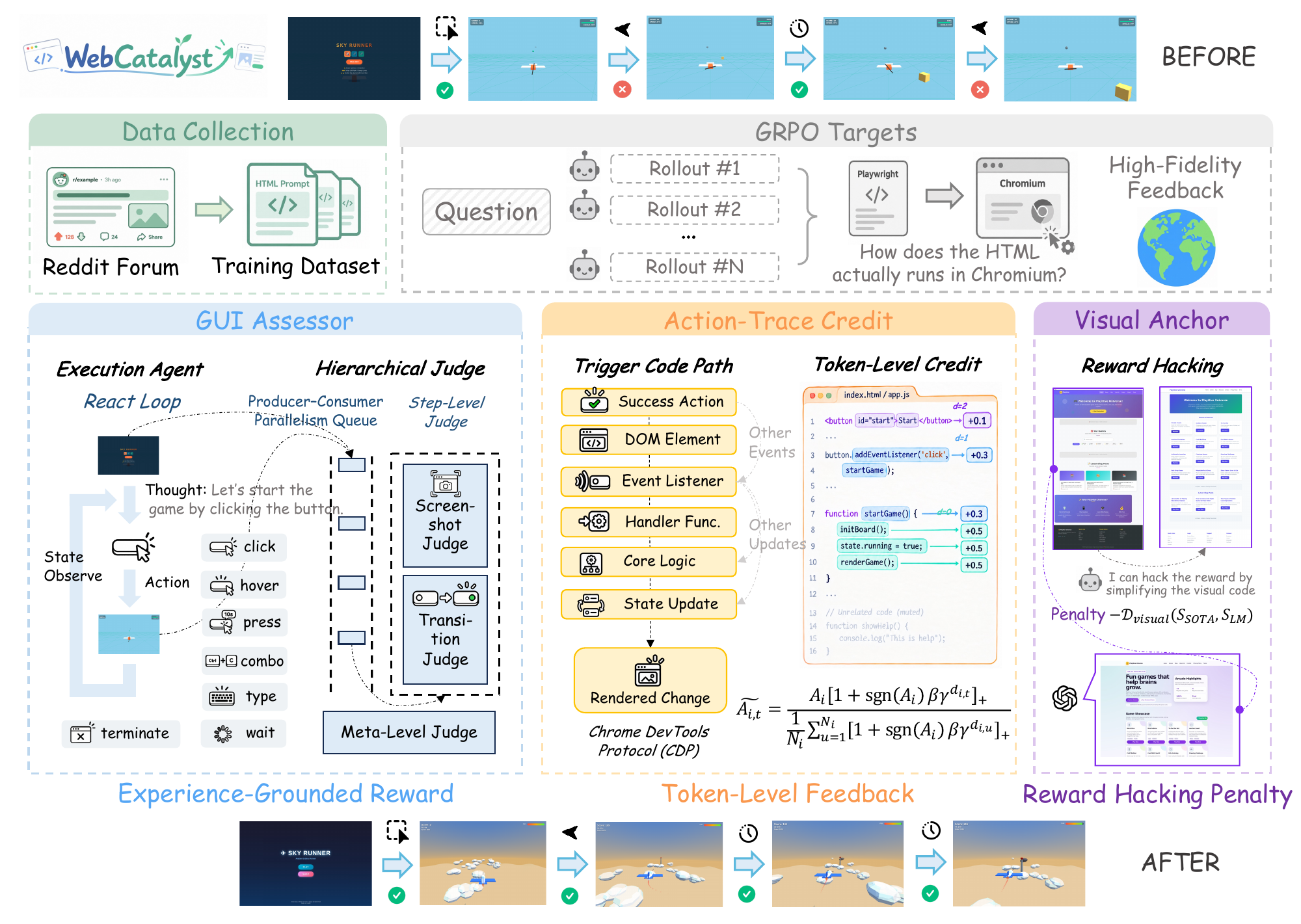} 
    \vspace{-8pt}
    \caption{Overview of WebCatalyst's browser-interaction RL framework.
Training requests come from Reddit, and sampled HTML artifacts are execute in Playwright.
The game-level GUI Assessor combines sustained interaction with hierarchical judgment for page-level rewards.
Visual Anchor penalizes visual degradation relative to reference screenshots, while Action-Trace Credit maps successful actions' runtime call paths to generated tokens and redistributes fine-grained advantages.
GRPO optimizes the generation policy using the calibrated rewards and token-level advantages.
Gameplay examples illustrate functionality and visual quality before and after training.
}
    \label{fig:method}
    \vspace{-10pt}
\end{figure*}

To move beyond task-specific webpage training toward catalyzing transferable capabilities for more demanding interactive HTML generation, we integrate interactive execution, reference calibration, and runtime-grounded credit assignment in one training framework (Figure \ref{fig:method}).
Given a requirement $x$, we sample $G$ responses $y_i=(y_{i,1},\ldots,y_{i,N_i})$ from $\pi_{\mathrm{old}}$ and render them as HTML pages $h_i$.
GUI Assessor produces interaction trajectories $\tau_i$ and page scores $R_i^{\mathrm{GUI}}$, improving \textbf{experience fidelity}.
Visual Anchor estimates visual degradation $D_i$ and calibrates page scores into rewards $r_i$, improving \textbf{reward signal fidelity}.
Action-Trace Credit distributes group-relative advantages $A_i$ into token-level advantages $\widehat A_{i,k}$, improving \textbf{credit-assignment fidelity}.
The policy $\pi_\theta$ is then updated using GRPO \citep{shao2024deepseekmath}.

\subsection{Experience-Grounded Execution and GUI Assessment}
\label{sec:guiassessor}

Short click-and-type tests in existing webpage-specific training cannot support sustained, multi-step interaction and state transitions in interactive HTML artifacts such as games, requiring a rich action space and sufficient interaction depth. To address this, we model $h_i$ as a state machine executed in a browser.
An Execution Agent runs a ReAct loop \citep{yao2023react} conditioned on requirement $x$, current screenshot $S_{i,t}$, prior actions and execution observation. It selects $a_{i,t}$ from
\begin{equation}
\begin{aligned}
\mathcal A=\{&
\texttt{click},\texttt{hover},\texttt{mouseMove},\texttt{press},\\
&\texttt{combo},\texttt{type},\texttt{wait},\texttt{terminate}\}.
\end{aligned}
\end{equation}
Action parameters include coordinates, key combinations, and durations, driving the page from its initial state $V_{i,0}$ to $V_{i,T_i}$:
\begin{equation}
\begin{aligned}
V_{i,t+1}
&=\mathcal T_{h_i}(V_{i,t},a_{i,t}),\\
S_{i,t}
&=\mathcal O(V_{i,t}),\\
E_{i,t}
&=(S_{i,t},a_{i,t},S_{i,t+1},m_{i,t}),\\
\tau_i
&=\{E_{i,t}\}_{t=0}^{T_i-1}.
\end{aligned}
\end{equation}
Here, $S_{i,t}$ is a screenshot, and $m_{i,t}$ is the browser execution record.
Interaction ends when the agent terminates or exhausts its budget.

\paragraph{Hierarchical Judge.}
Local, single-step judgments cannot capture overall quality across sustained interaction. We therefore adopt a two-level hierarchy consisting of step-level assessment and trajectory-level aggregation.
For each transition, the Step-level Judge uses before-and-after screenshots, the action, and the execution record to assess post-action screenshot $S_{i,t+1}$'s visual quality and the resulting behavior, producing a screenshot score $q^{\mathrm{sc}}_{i,t}$ and a transition score $q^{\mathrm{tr}}_{i,t}$.
The Meta-level Judge eventually combines memory with all step-level assessments to evaluate functional behavior and requirement fulfillment across sustained interactions, yielding page-level supervision feedback ranging from 0 to 100:
\begin{equation}
\begin{aligned}
R_i^{\mathrm{GUI}}
&=\mathcal J_{\mathrm{meta}}\left(
x,\{\mathcal J_{\mathrm{step}}(E_{i,t})\}_{t=0}^{T_i-1}
\right)\\
\end{aligned}
\end{equation}
This hierarchy aggregates individual-action evidence to assess overall user experience.
Detailed scoring criteria appear in the appendix \ref{sec:appendix-gui-assessor}.

\paragraph{Producer--Consumer Queue.}
Sustained interaction makes sequential execution and assessment prohibitively slow.
After each action, the Execution Agent enqueues the resulting transition $E_{i,t}$ and immediately proceeds to the next step, while the Step-level Judge evaluates queued transitions concurrently.
Once execution terminates, we collect all step-level assessments and pass them to the Meta-level Judge for aggregation.
This producer--consumer pipeline overlaps interaction with assessment and reduces evaluation latency.

\subsection{Reference-Calibrated Visual Anchor}
\label{sec:visualanchor}

When evaluating HTML artifacts involving richer interactions, we observe that relying solely on a VLM-based GUI judge creates a reward loophole: the policy can improve interaction scores by simplifying visual structure or omitting diverse features.
As illustrated in Figure~\ref{fig:method}, such artifacts may preserve a few easily verified interactions while degrading visual dimensions to which the judge is insensitive, yielding high-reward yet visually inferior outputs.
Thus, we introduce Visual Anchor, which compares the generated artifact's screenshot $S_{\mathrm{LM}}$ against a well-organized reference screenshot $S_{\mathrm{SOTA}}$ to estimate one-sided visual degradation:
\begin{equation}
D_i=
\mathcal D_{\mathrm{visual}}
(S_{\mathrm{SOTA}},S_{\mathrm{LM}};x)
\in[0,100].
\end{equation}

\paragraph{Reference Construction and Calibration.}
To provide a stable visual baseline for otherwise reference-free assessment, a capable model generates reference HTML pages before training, which the GUI Assessor executes.
For both generated and reference pages, we select the highest-scoring screenshot to avoid unrepresentative states such as startup or failure screens:
\begin{equation}
\begin{aligned}
t_i^\star
&=\arg\max_t q^{\mathrm{sc}}_{i,t},
& S_{\mathrm{LM}}
&=S_{i,t_i^\star+1},\\
t_x^\star
&=\arg\max_t q^{\mathrm{sc,ref}}_{x,t},
& S_{\mathrm{SOTA}}
&=S^{\mathrm{ref}}_{x,t_x^\star+1}.
\end{aligned}
\end{equation}
The pairwise visual judge evaluates degradation in layout, visual hierarchy, and design completeness.
It assigns $D_i=0$ when the generated page is comparable to or better than the reference, and applies a penalty to the reward when degradation is clear.
\subsection{Runtime-Grounded Credit Assignment}
\label{sec:creditassign}

Even with calibrated page-level rewards, optimizers such as GRPO uniformly assign advantage $A_i$ to all generated tokens, obscuring which code regions contribute to individual interactive behaviors.
Directly penalizing failed actions can cause misattribution: a failure may arise from incorrect or missing logic,  whose responsible code can be ambiguous or absent from the runtime trace. Thinking in reverse, we derive \textbf{Action-Trace Credit (ATC)} from the runtime call paths of successful actions.

\paragraph{Runtime Code Tracing.}
To identify the code responsible for individual interactions, we capture JavaScript call stacks via the Chrome DevTools Protocol during traceable actions, retaining only traces from actions classified as successful by the Step-level Judge.
Each stack contains the active leaf frame followed by its callers.
Following this leaf-to-caller order, we assign depth $0$ to the leaf and progressively larger depths to its callers.
We map locatable inline JavaScript function spans to generated tokens.
For each token $k$, set $\mathcal F_{i,k}$ records the traced depth of the functions involving it.

\paragraph{Minimum Call Depth and Depth Decay.}
A token may occur in multiple traced functions and call paths from different actions.
We take the minimum call depth $d_{i,k}^{\min}$ across its traces and modulate credit by $\gamma^{d_{i,k}^{\min}}$, analogous to discounting in Proximal Policy Optimization \citep{schulman2017ppo}. 
\begin{equation}
\begin{aligned}
d_{i,k}^{\min}
&=
\begin{cases}
\displaystyle\min_{(f,d)\in\mathcal F_{i,k}}d,
&\mathcal F_{i,k}\neq\varnothing,\\[4pt]
+\infty,
&\mathcal F_{i,k}=\varnothing
\end{cases}\\
c_{i,k}
&=\gamma^{d_{i,k}^{\min}}
\end{aligned}
\end{equation}
where $\gamma\in(0,1)$.
Under the implemented leaf-to-caller convention, leaf-aligned tokens, which typically executing core action logic and state updates, receive the largest trace coefficient, decaying toward outer callers such as event handlers.
We adopt $\gamma^{+\infty}=0$, preserving baseline weights for tokens outside successful actions' traces.
Using the sign of page-level advantage, we define the modulation weight and token-level advantage as
\begin{equation}
\begin{aligned}
\widetilde w_{i,k}
&=
\left[
1+\beta\,\operatorname{sgn}(A_i)c_{i,k}
\right]_+,\\
\widehat A_{i,k}
&=
\frac{A_i\widetilde w_{i,k}}
{\frac{1}{N_i}\sum_{u=1}^{N_i}\widetilde w_{i,u}}.
\end{aligned}
\end{equation}
Here, $[z]_+=\max(z,0)$, and $0\leq\beta<1$ controls modulation strength.
Relative to untraced tokens, tokens involved in successful actions receive stronger positive updates when $A_i>0$ and weaker negative updates when $A_i<0$.
The normalization preserves the mean token-level advantage at $A_i$.
See appendix for further details.

\subsection{Joint Reinforcement Learning Objective}

We integrate experience assessment, visual calibration, and runtime-grounded credit assignment into a unified GRPO objective. For each candidate $i$, Visual Anchor calibrates the GUI Assessor score into the final reward, from which we compute advantages within each generated-candidate group:
\begin{equation}
\begin{aligned}
r_i
&=\frac{R_i^{\mathrm{GUI}}-D_i}{100},\\
A_i
&=\frac{r_i-\operatorname{mean}_{j}(r_j)}
{\operatorname{std}_{j}(r_j)+\epsilon}.
\end{aligned}
\end{equation}
With ATC-derived token-level advantages $\widehat A_{i,k}$, we maximize the following training objective
{
\small
\begin{equation}
\begin{aligned}
\mathcal J(\theta)
&=\mathbb E\Bigg[
\frac{1}{G}
\sum_{i=1}^{G}
\frac{1}{N_i}
\sum_{k=1}^{N_i}
\Bigl(
\min\bigl\{
\rho_{i,k}\widehat A_{i,k},\\
\qquad
\operatorname{clip}&
(\rho_{i,k},1-\varepsilon_c,1+\varepsilon_c)
\widehat A_{i,k}
\bigr\}
-\eta\,\mathrm{KL}_{i,k}
\Bigr)\Bigg].
\end{aligned}
\end{equation}
}
where
\begin{equation}
\rho_{i,k}
=\frac{\pi_\theta(y_{i,k}\mid x,y_{i,<k})}
{\pi_{\mathrm{old}}(y_{i,k}\mid x,y_{i,<k})}.
\end{equation}
The expectation is taken over requirements and candidate responses.
$\mathrm{KL}_{i,k}$ estimates token-level divergence between the current policy \(\pi_\theta\) and fixed reference policy \(\pi_0\), where $\eta$ is the KL coefficient.

\section{Experiments}
\label{sec:experiments}


\subsection{Experimental Setup}
\definecolor{rankbg}{gray}{0.97}
\newcommand{\lockicon}{%
  \raisebox{0.1ex}{\scriptsize\faLock}%
}
\definecolor{rankgreen}{HTML}{137A3A}
\newcommand{\topfour}[1]{\textcolor{rankgreen}{#1}}
\definecolor{darkgreen}{HTML}{176B3A}
\newcommand{\deltaval}[1]{\textcolor{darkgreen}{\textbf{#1}}}

\begin{table*}[htbp]
\centering
\caption{
Main results on WebGen-Bench and Cookie-Bench Mini Games.
Acc., Aesth., and Func. denote accuracy, aesthetics, and functionality.
Size gives amount of total/active parameters, Avg. Rank averages ranks across evaluation dimensions, and $\Delta$ gives changes from Qwen3.6-35B-A3B.
Arrows indicate preferred directions.
Bold marks group-wise best results, with WebCatalyst included in Other LLMs.
WebCatalyst, with only 640 game development requests, improves both game and webpage generation and surpasses Claude Opus 4.7 in average rank.
}
\label{tab:main_results}

\small
\setlength{\tabcolsep}{2.8pt}
\renewcommand{\arraystretch}{1.0}

\resizebox{\textwidth}{!}{%
\begin{tabular}{lccccccccccccc>{\columncolor{rankbg}}c}
\toprule
\multirow{3}{*}{\textbf{Model}}
& \multirow{3}{*}{\textbf{Date}}
& \multirow{3}{*}{\textbf{Size}}
& \multicolumn{8}{c}{\textbf{WebGen-Bench}}
& \multicolumn{3}{c}{\textbf{Cookie-Bench Mini Games}}
& \multicolumn{1}{c}{%
    \multirow{3}{*}{%
      \makecell{\textbf{Avg.}\\\textbf{Rank} $\downarrow$}
    }
  } \\
\cmidrule(lr){4-11}
\cmidrule(lr){12-14}
& & &
\makecell{\textbf{Start Fail.}\\\textbf{Rate} $\downarrow$}
& \textbf{Acc.} $\uparrow$
& \makecell{\textbf{Content}\\\textbf{Pres.} $\uparrow$}
& \makecell{\textbf{User}\\\textbf{Inter.} $\uparrow$}
& \makecell{\textbf{Data}\\\textbf{Mgmt.} $\uparrow$}
& \makecell{\textbf{Func.}\\\textbf{Test} $\uparrow$}
& \makecell{\textbf{Data}\\\textbf{Display} $\uparrow$}
& \makecell{\textbf{Design}\\\textbf{Val.} $\uparrow$}
& \textbf{Aesth.} $\uparrow$
& \textbf{Func.} $\uparrow$
& \textbf{Total} $\uparrow$
& \multicolumn{1}{c}{} \\

\midrule
\multicolumn{15}{c}{\textit{\textbf{Other LLMs}}} \\

GPT-OSS-120B
& 25/08 & 116B/5B
& 1.4 & 37.6 & 49.7 & 31.3 & 36.6 & 26.0 & 43.0 & 61.5
& 40.3 & 54.4 & 47.4 & 19.5 \\

Qwen3.5-122B-A10B
& 26/02 & 122B/10B
& 2.6 & 49.8 & 57.5 & 46.0 & 49.1 & 37.0 & 55.9 & 76.2
& 66.0 & 60.5 & 63.2 & 16.0 \\

Qwen3.5-397B-A17B
& 26/02 & 397B/17B
& 0.5 & 48.7 & 59.2 & 43.5 & 47.5 & 33.6 & 57.0 & 77.9
& 69.9 & 65.0 & 67.5 & 14.1 \\

Qwen3.6-35B-A3B
& 26/04 & 35B/3B
& \textbf{0.0} & 48.8 & 59.2 & 43.0 & 47.1 & 33.9 & 61.8 & 70.5
& 81.5 & 56.1 & 68.8 & 14.0 \\

Qwen3.6-Plus
& 26/04 & N/A
& \textbf{0.0} & 53.6 & 61.5 & 50.0 & 52.2 & 41.4 & 65.9 & 68.9
& 84.8 & 63.6 & 74.2 & 8.5 \\

Qwen3.6-Max
& 26/04 & N/A
& \textbf{0.0} & 55.4 & 64.7 & 51.1 & 53.8 & \textbf{45.3} & 64.0 & 70.5
& \textbf{87.8} & 64.4 & 76.1 & 5.9 \\

Qwen3.7-Flash
& 26/07 & N/A
& 1.1 & 48.5 & 54.6 & 45.2 & 48.1 & 33.6 & 65.1 & 64.3
& 84.0 & 54.2 & 69.1 & 15.5 \\

GLM-5.1
& 26/04 & 754B/40B
& \textbf{0.0} & 54.0 & 63.2 & 47.4 & 56.9 & 39.4 & 64.5 & \textbf{78.7}
& \textbf{87.8} & 61.2 & 74.5 & 6.9 \\

GLM-5.2
& 26/06 & 754B/40B
& \textbf{0.0} & \textbf{57.5} & 62.1 & 52.4 & \textbf{62.5} & 45.1
& 68.6 & 73.4
& 87.2 & \textbf{65.3} & \textbf{76.3} & 4.3 \\

Minimax-M2.7
& 26/03 & 229B/10B
& 0.9 & 53.4 & 62.4 & 49.0 & 52.2 & 38.8 & 66.4 & 74.2
& 81.4 & 60.3 & 70.8 & 10.3 \\

DeepSeek-V3.2
& 25/12 & 671B/37B
& \textbf{0.0} & 49.7 & 56.0 & 45.2 & 51.6 & 35.1 & 60.2 & 74.2
& 69.0 & 56.5 & 62.8 & 13.9 \\

DeepSeek-V4-Flash
& 26/07 & 284B/13B
& 2.6 & 51.3 & 58.0 & 45.4 & 55.6 & 37.9 & 58.3 & 77.9
& 72.2 & 57.9 & 65.1 & 13.3 \\

Kimi-K2.6
& 26/04 & 1T/32B
& 0.9 & 54.9 & \textbf{68.4} & 49.8 & 50.3 & 42.2 & 66.7 & 72.5
& 83.5 & 58.7 & 68.6 & 9.3 \\

\midrule
\multicolumn{15}{c}{\textit{\textbf{Proprietary LLMs}}} \\

Gemini-3.1-Pro
& 26/02 & \lockicon
& \textbf{0.0} & 53.2 & 60.9 & 50.2 & 50.6 & 39.1 & 65.6 & 73.4
& 62.9 & 60.9 & 61.9 & 11.1 \\

GPT-5.2
& 25/12 & \lockicon
& 2.9 & 50.5 & 65.2 & 47.8 & 39.7 & 36.4 & 62.9 & 70.5
& 85.0 & 63.7 & 74.3 & 12.3 \\

GPT-5.3-codex
& 26/02 & \lockicon
& \textbf{0.0} & 54.4 & 65.8 & 52.4 & 45.9 & 40.9 & 65.9 & 74.6
& 79.4 & \textbf{69.6} & 74.5 & 7.2 \\

GPT-5.4-mini
& 26/03 & \lockicon
& \textbf{0.0} & 52.6 & 62.6 & 47.6 & 51.2 & 37.6 & 61.6 & \textbf{80.3}
& 80.8 & 64.1 & 72.4 & 9.4 \\

Claude-Opus-4.7
& 26/04 & \lockicon
& \textbf{0.0} & 57.0 & \textbf{68.4} & 48.1 & \textbf{62.2} & 41.6
& \textbf{72.8} & 75.8
& 80.7 & 69.5 & 75.1 & 4.7 \\

GPT-5.5
& 26/04 & \lockicon
& \textbf{0.0} & \textbf{59.3} & 66.1 & \textbf{54.6} & 60.9
& \textbf{45.7} & 70.2 & \textbf{80.3}
& \textbf{86.4} & 66.9 & \textbf{76.7} & \textbf{2.1} \\

\midrule
\multicolumn{15}{c}{\textit{\textbf{Ours}}} \\
\rowcolor{green!12}
\textbf{WebCatalyst}
& 26/04 & 35B/3B
& \textbf{0.0}
& 56.9
& 68.1
& \textbf{52.9}
& 54.2
& 42.2
& \textbf{68.8}
& 77.9
& 87.5
& 63.2
& 75.3
& \textbf{4.0} \\
\addlinespace[2pt]
\textbf{$\Delta$ from Base}
& --
& --
& \deltaval{0.0}
& \deltaval{+8.1}
& \deltaval{+8.9}
& \deltaval{+9.9}
& \deltaval{+7.1}
& \deltaval{+8.3}
& \deltaval{+7.0}
& \deltaval{+7.4}
& \deltaval{+5.9}
& \deltaval{+7.0}
& \deltaval{+6.5}
& \deltaval{-10.0} \\

\bottomrule
\end{tabular}%
}
\vspace{-13pt}
\end{table*}

\paragraph{Benchmarks.}
We evaluate HTML generation at different levels of interaction complexity using WebGen-Bench~\citep{lu2025webgenbench} and the Mini Games split of Cookie-Bench~\citep{yang2026cookiebench}.
WebGen-Bench contains 101 webpage-generation requests and 647 test cases, each specifying an operation and expected outcome to verify a functional or presentation requirement. For Cookie-Bench Mini Games, we evaluate 156 deduplicated requests. The benchmark uses UI-TARS~\citep{qin2025uitars} to continuously interact with browser, producing visual and behavioral traces used to assess aesthetics and functionality which tests more demanding interactive HTML generation.

\paragraph{Metrics.}
For WebGen-Bench, an evaluation agent executes each test case and compares the observed behavior with its expected outcome, assigning a verdict of \textsc{Yes}, \textsc{Partial}, or \textsc{No}.
These verdicts receive scores of $1$, $0.5$, and $0$, respectively, and \textbf{Accuracy} is their mean across all test cases.
We also report Accuracy by category, Content Presentation, User Interaction, Data Management, Functional Testing, Data Display Testing, and Design Validation Testing.
\textbf{Start Failure Rate} is the percentage of generated websites that fail to launch. Lower is better.

For Cookie-Bench Mini Games, the evaluator scores each artifact's visual quality (\textbf{Aesthetics}) and its behavior robustness over sustained interaction (\textbf{Functionality}).
We average request-level scores and report them on a $0$--$100$ scale.
\textbf{Total} aggregates performance across both dimensions.


\paragraph{Implementation Details.}
We construct the training set by collecting Reddit posts about HTML development, filtering them with game-related keywords, and rewriting them as direct user requests. This allows us to test whether interactive HTML training with high-fidelity feedback transfers to simpler downstream tasks such as conventional webpages. We train Qwen3.6-35B-A3B \citep{qwen36_35b_a3b} on NVIDIA GB200 GPUs with a batch size of 16 and group size $G=5$. For ATC, we set $\beta=0.5$ and the decay factor $\gamma=0.9$. We use GPT-5.6 Terra as the judging model in the GUI Assessor, allowing up to eight interaction turns per page. We use GPT-5.5 to generate Visual Anchor's reference artifacts. We evaluate the checkpoint after $40$ training steps, corresponding to $640$ processed requests. Further implementation details are provided in the Appendix~\ref{sec:implementdetails}.


\subsection{Main Results}

Table~\ref{tab:main_results} compares WebCatalyst on Cookie-Bench Mini Games and WebGen-Bench, representing complex interactive artifacts and conventional webpages, respectively.
Because our training requests are collected using game-related keywords, Mini Games provides a task-aligned but demanding evaluation.
WebGen-Bench instead uses rubric-based test cases for conventional webpages, providing a cross-setting evaluation of whether capabilities learned from interactive game generation transfer to broader webpage tasks.

First, \textbf{high-fidelity feedback enables competitive mini-game generation.}
Using 640 training requests, WebCatalyst raises the Mini Games Total score of Qwen3.6-35B-A3B from 68.8 to 75.3.
The improvement spans both evaluation dimensions: Aesthetics increases from 81.5 to 87.5, and Functionality from 56.1 to 63.2.
Despite using a relatively compact backbone with 35B total parameters and 3B activated per token, WebCatalyst outperforms several contemporary frontier models, including MiniMax-M2.7, GLM-5.1, and Kimi-K2.6, and matches Claude Opus 4.7 on the Total metric.

Second, \textbf{the gains transfer to conventional webpages with simpler operations and design requirements.}
Despite being trained primarily on game-focused requests, WebCatalyst improves overall WebGen-Bench Accuracy from 48.8 to 56.9, with improvements across all request- and test-case categories.
Notable gains appear in User Interaction (43.0 to 52.9), Content Presentation (59.2 to 68.1), and Functional Testing (33.9 to 42.2), indicating that the transfer covers both executable behavior and visual presentation.
Its overall Accuracy exceeds Kimi-K2.6 (54.9) and nearly matches Claude Opus 4.7 (57.0).
These results suggest that high-fidelity training on complex interactive artifacts develops capabilities that generalize beyond game tasks to webpage generation.

To aggregate performance across the heterogeneous metrics of both benchmarks, we compute each model's mean rank, reported as Average Rank.
WebCatalyst achieves an Average Rank of 4.0, ranking first among the evaluated open-weight models and ahead of most proprietary models, including Claude Opus 4.7 (4.7).
Among all evaluated models, only GPT-5.5 ranks higher (2.1).

Overall, high-fidelity reinforcement learning supervision with only 640 game-focused requests enables WebCatalyst to achieve competitive cross-benchmark performance, transferring gains from mini-games to conventional webpages and narrowing the gap to leading proprietary models.

\subsection{Ablation Study}

\begin{table}[t]
\centering
\caption{
Ablation study of components in WebCatalyst.
Results show the contributions of experience,
reward-signal, and credit-assignment fidelity.
}
\label{tab:webcatalyst-ablation}
\resizebox{0.9\columnwidth}{!}{%
\begin{tabular}{lcccc}
\toprule
\multirow{2}{*}{\textbf{Variant}}
& \multicolumn{1}{c}{\textbf{WebGen}}
& \multicolumn{3}{c}{\textbf{Mini-Games}} \\
\cmidrule(lr){2-2}
\cmidrule(lr){3-5}
& Acc. & Aesth. & Func. & Total \\
\midrule
\rowcolor{gray!5}
Qwen3.6-35B-A3B
& 48.8 & 81.5 & 56.1 & 68.8 \\

\hspace*{1em}+GUI Assessor
& 52.2 & 71.3 & 60.5 & 65.9 \\

\hspace*{2em}+Self Anchor
& 55.5 & 82.7 & 57.3 & 70.0 \\

\hspace*{2em}+GPT-5.5 Anchor
& 56.8 & 82.3 & 58.2 & 70.3 \\

\rowcolor{green!12}
\hspace*{3em}\textbf{+ATC}
& \textbf{56.9}
& \textbf{87.5}
& \textbf{63.2}
& \textbf{75.3} \\
\bottomrule
\end{tabular}%
}
\vspace{-13pt}
\end{table}

Table~\ref{tab:webcatalyst-ablation} isolates the complementary contributions of the three feedback-fidelity components.
Adding the GUI Assessor increases WebGen Accuracy from 48.8 to 52.2 and Mini Games Functionality from 56.1 to 60.5.
However, Aesthetics drops from 81.5 to 71.3, reducing Total from 68.8 to 65.9.
This trade-off shows that unanchored GUI feedback can reward functional improvements while overlooking visual degradation, exposing the reward loophole.

Visual Anchor mitigates this degradation.
Self Anchor uses screenshots from the base model's pre-RL outputs as references, whereas GPT-5.5 Anchor uses screenshots from GPT-5.5-generated artifacts.
Both variants restore Aesthetics above the backbone level and increase Total from 65.9 to 70.0 and 70.3, respectively, while further improving WebGen Accuracy to 55.5 and 56.8.
These results show that reference-based calibration discourages visual simplification while preserving functional gains. Both Visual Anchor variants mitigate reward-induced visual degradation, while further analysis suggests that higher-quality anchors help maintain training stability (\S \ref{sec:stronger}).

Finally, adding the action-trace credit raises Mini Games aesthetics, functionality, and Total from 82.3, 58.2, and 70.3 to 87.5, 63.2, and 75.3, while preserving WebGen accuracy at 56.9.
Thus, Visual Anchor and ATC play complementary roles: the former calibrates what makes a high-quality artifact, whereas the latter assigns the resulting feedback more precisely to responsible tokens.

\subsection{Analysis}

\subsubsection{Looking into Training Dynamics}

\begin{figure}[htbp]
  \vspace{-10pt}
  \centering
  \includegraphics[width=\columnwidth]{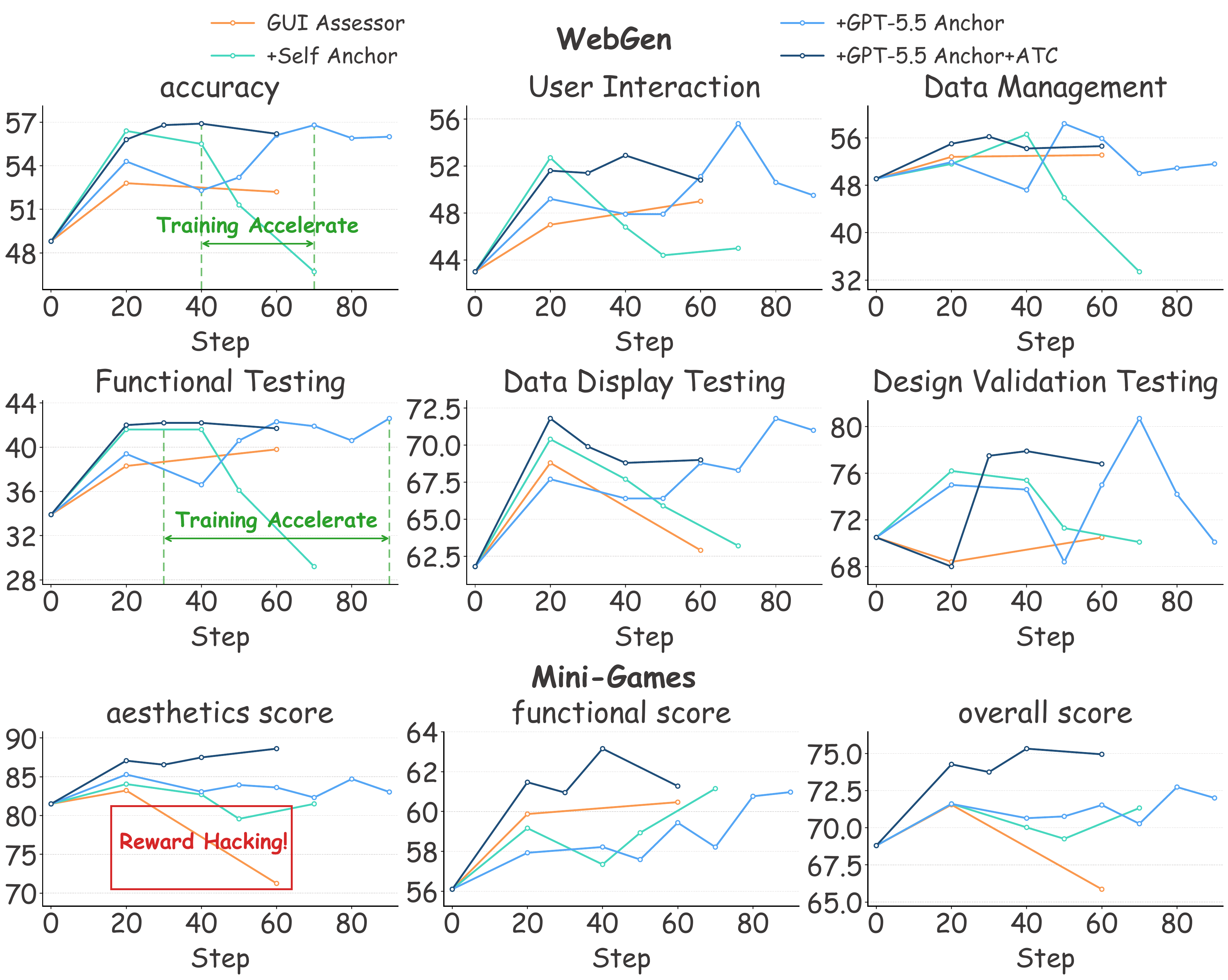}
  \vspace{-20pt}
  \caption{
    Evaluation performance across training steps and feedback configurations.
    The top two rows show WebGen-Bench performance,
    while the bottom row shows Mini Games aesthetics, functionality, and overall scores.
    Anchor denotes Visual Anchor, and ATC denotes Action-Trace Credit.
  }
  \label{fig:experiments}
  \vspace{-15pt} 
\end{figure}

\paragraph{WebGen Gains Can Mask Feedback-Fidelity Failures.}
Figure~\ref{fig:experiments} illustrates why improvements on WebGen-Bench alone do not establish feedback fidelity across interaction regimes.
When training with only the GUI Assessor, several WebGen metrics improve steadily, whereas Mini Games Aesthetics deteriorates even as Functionality increases.
Conventional webpage evaluations can therefore capture improvements in local operation success while masking visual and feature simplification under sustained interaction.
The results show that WebGen gains alone do not demonstrate that the supervision faithfully reflects the quality of more demanding interactive artifacts.

\label{sec:stronger}
\paragraph{Higher-Quality Visual References Stabilize Training.}
Figure~\ref{fig:experiments} further shows that reference quality affects training stability.
Although Self Anchor yields early improvements, several metrics deteriorate during later training.
In contrast, GPT-5.5 Anchor remains comparatively stable and retains its gains throughout training.
This comparison suggests that weaker references still leave room for one-sided visual degradation, whereas higher-quality reference artifacts provide a more reliable calibration baseline that limits such reward exploitation and stabilizes optimization.
\vspace{-2.5pt}

\paragraph{Action-Trace Credit Accelerates Optimization.}
While improving complex interactive HTML artifacts' quality with fine-grained reward signals, Action-Trace Credit also changes how quickly the model converts interaction feedback into measurable improvements.
In Figure~\ref{fig:experiments}, ATC reaches approximately 56--57\% WebGen accuracy at step 30, whereas the same GPT-5.5 Anchor without ATC reaches comparable accuracy only around step 70. Functional Testing shows a similar pattern.

\begin{figure}[htbp]
  \vspace{-10pt}
  \centering
  \includegraphics[width=0.9\columnwidth]{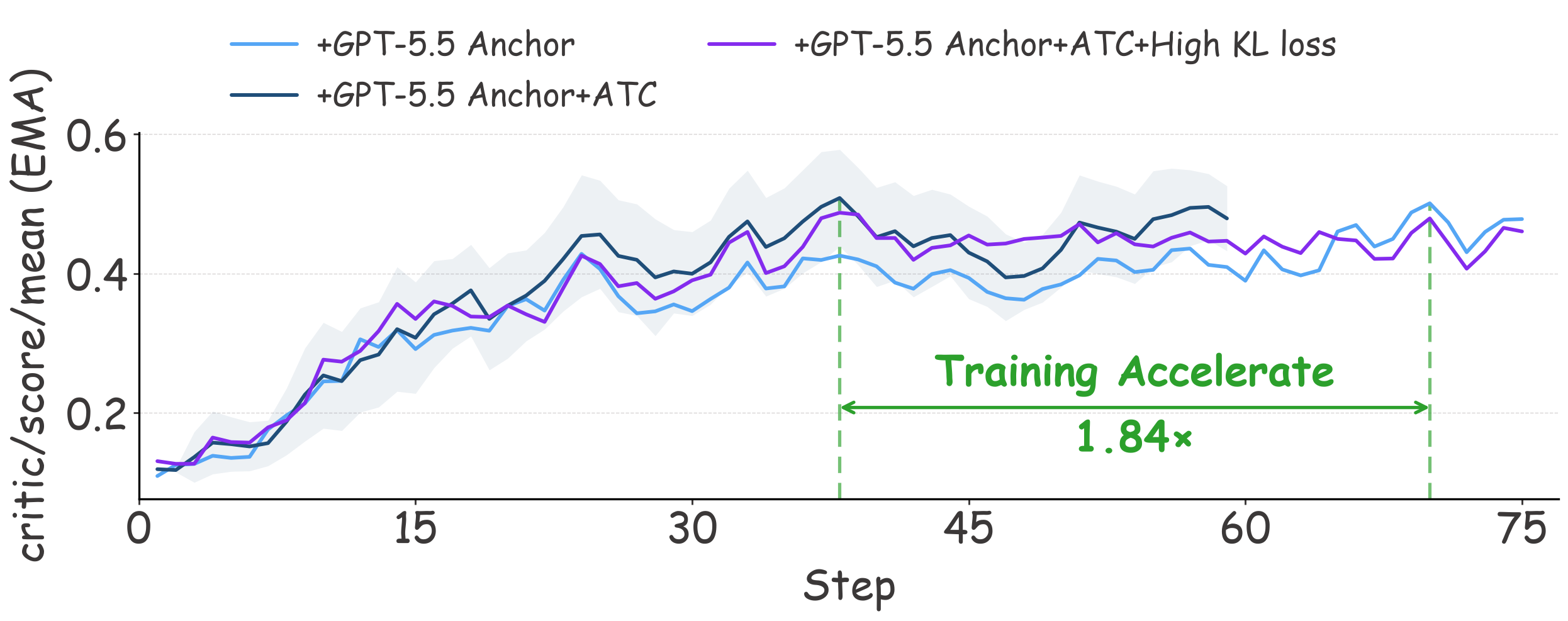}
  \vspace{-7.5pt}
  \caption{
    Reward trajectories with and without runtime-grounded credit assignment.
The vertical axis shows the EMA of the mean critic score, and High KL denotes stronger KL regularization.
Dashed markers indicate when variants reach comparable score levels.
All settings exhibit diminishing gains after approximately 40 steps. Optimization is approaching convergence.
  }
  \vspace{-10pt}
  \label{fig:score_change}
\end{figure}

Figure~\ref{fig:score_change} provides consistent optimization evidence: ATC reduces training steps to reach a comparable EMA critic score by approximately $1.84\times$.
Together, evaluation and training curves show that runtime-grounded credit assignment accelerates convergence by directing interaction feedback to code responsible for successful actions.

\subsubsection{Experience Diagnostics and Iterative Refinement}

\begin{figure}[htbp]
  \centering
  \includegraphics[width=\columnwidth]{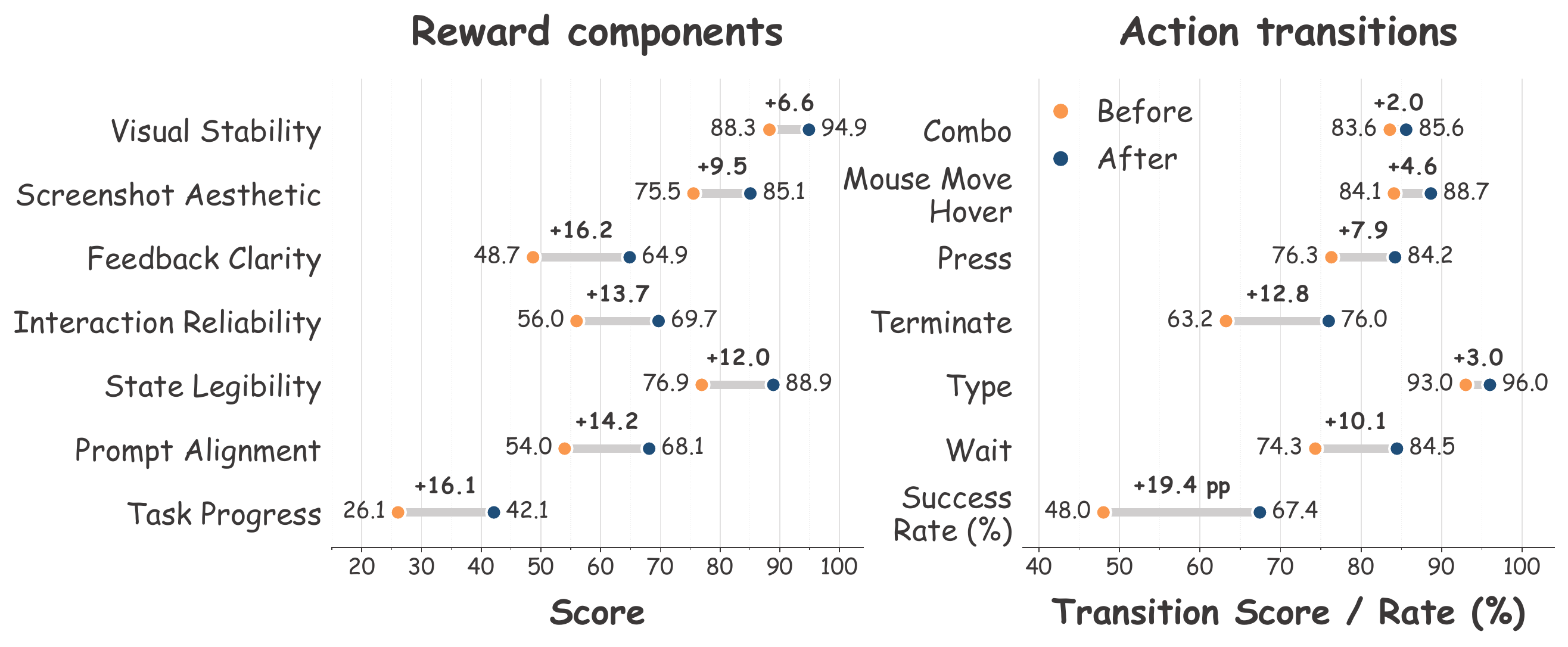}
  \vspace{-20pt}
  \caption{
    User-experience and action-level gains before and after training.
    Before and After denote Qwen3.6-35B-A3B and WebCatalyst, respectively.
    The left panel shows experience-dimension scores,
    while the right shows action-transition scores.
  }
  \label{fig:experience_diagnostics}
  \vspace{-15pt} 
\end{figure}

\paragraph{Gains Extend Across Experience Dimensions.}
We further analyze the Meta-level Judge's experience dimensions and Step-level Judge outcomes.
Figure~\ref{fig:experience_diagnostics} shows WebCatalyst improves all six evaluated dimensions, including visual stability, feedback clarity, interaction reliability, state legibility, prompt alignment, and task progress.
Gains also span all action types, raising the overall action success rate from 48.0\% to 67.4\%.
These broad gains indicate that the aggregate benchmark improvements are not driven by a narrow subset of behaviors, but reflect more general improvements in interaction quality.
\vspace{-2.5pt}
\paragraph{Single-Shot Gains Persist under Iterative Refinement.}
We examine whether WebCatalyst's single-shot improvements persist in an iterative refinement pipeline that emulates human-in-the-loop development, where feedback grounded in actual use of the generated artifact guides successive HTML revisions. Each round, the GUI Assessor interacts with the generated artifact, identifies experiential weaknesses and issues, and provides this feedback for revision.
As shown in Table~\ref{tab:refinement}, WebCatalyst achieves a Total score of 75.3 without refinement, matching the backbone's 75.2 after two refinement rounds.
It also maintains a higher Total score at every corresponding round and reaches 76.9 after three rounds, exceeding the backbone's best of 75.2.
Thus, WebCatalyst reduces feedback rounds to comparable quality and improves attainable quality within the evaluated refinement budget.

\begin{table}[htbp]
\centering
\small
\vspace{-2.5pt}
\caption{
Mini Games performance under iterative refinement.
Round 0 denotes the initial generation, and subsequent rounds revise the artifact using feedback from the GUI Assessor.
Bold indicates the better result for each metric within each round.
}
\vspace{-5pt}
\resizebox{\linewidth}{!}{%
\begin{tabular}{ccccccc}
\toprule
\multirow{2}{*}{\textbf{Round}}
& \multicolumn{3}{c}{\textbf{Qwen3.6-35B-A3B}}
& \multicolumn{3}{c}{\textbf{+WebCatalyst}} \\
\cmidrule(lr){2-4}
\cmidrule(lr){5-7}
& Aes. & Func. & Total
& Aes. & Func. & Total \\
\midrule
0
& 81.5 & 56.1 & 68.8
& \textbf{87.5} & \textbf{63.2} & \textbf{75.3} \\

1
& 85.1 & 57.9 & 71.5
& \textbf{88.1} & \textbf{63.4} & \textbf{75.7} \\

2
& 87.7 & \textbf{62.8} & 75.2
& \textbf{89.5} & 61.8 & \textbf{75.7} \\

3
& 86.2 & 62.1 & 74.2
& \textbf{91.1} & \textbf{62.7} & \textbf{76.9} \\
\bottomrule
\end{tabular}%
}
\label{tab:refinement}
\vspace{-10pt} 
\end{table}

\section{Conclusion}

We introduce \textbf{WebCatalyst}, a unified browser-interaction RL framework that improves supervision fidelity across interaction experience, reward signals, and credit assignment.
Trained on fewer than 1K requests, primarily mini-games, WebCatalyst substantially improves mini-game generation, with gains transferring to conventional webpage tasks.
These results underscore the importance of high-fidelity feedback for moving beyond task-specific webpage optimization toward more capable and general interactive HTML generation.

\clearpage
\section{Limitations}

While WebCatalyst enhances the interactive HTML generation capabilities of LLMs, several aspects remain to be explored. First, this work focuses on direct generation. Although our experiments show that stronger direct generation also benefits multi-turn refinement, we do not specifically optimize more complex pipelines involving planning, iterative debugging, or tool use. Second, WebCatalyst studies how RL activates and strengthens interactive HTML capabilities already present in the base model. It complements rather than replaces SFT, which establishes the foundation for visual style, animation, and artifact complexity. The effectiveness of RL therefore remains dependent on the capabilities acquired before training. Finally, we train our model on game prompts to examine whether capabilities learned from interaction-intensive tasks transfer to simpler settings such as conventional webpages. We leave the optimization of task mixtures and data composition, as well as their effects on performance and generalization, to future work.

More broadly, stronger interactive HTML generation capabilities may introduce misuse risks, including the generation of deceptive interfaces, phishing websites, malicious client-side behaviors, or other harmful web content. Although our work focuses on benign generation tasks, future deployment should incorporate appropriate safety safeguards and misuse-aware evaluation.

\bibliography{custom}

\clearpage
\appendix

\section{Appendix}
\label{sec:appendix}

\subsection{GUI Assessor Pipeline}
\label{sec:appendix-gui-assessor}

This section includes the details of our GUI Assessor.  Given a
requirement prompt $x$ and a sampled HTML artifact $h_i$, the assessor (i) executes
$h_i$ under controlled browser budget, (ii) chooses and performs a sequence of
user actions, (iii) evaluates every transition as soon as it is produced, and
(iv) aggregates the completed trajectory into the page-level score
$R_i^{\mathrm{GUI}}$. We present the overall workflow of the GUI Assessor below, together with the key design considerations that guide its construction.

\subsubsection{Controlled Browser Execution}

We load $h_i$ in a fresh Chromium context\footnote{The Chromium context is implemented within Playwright. \url{https://github.com/microsoft/playwright}} with a fixed $1440\!\times\!900$
viewport and wait for network idleness before interaction.  Playwright's
virtual clock is then paused and advanced only by the executor.  This prevents
uncontrolled wall-clock delays from changing animations or game state while a
model call is in flight.  At turn $t$, the page has latent browser state
$V_{i,t}$ and observable screenshot $S_{i,t}=\mathcal O(V_{i,t})$.  After
choosing and executing $a_{i,t}$, the browser produces
$V_{i,t+1}=\mathcal T_{h_i}(V_{i,t},a_{i,t})$, the post-action screenshot
$S_{i,t+1}$, and the execution record $m_{i,t}$.  Together they form the
transition $E_{i,t}$ defined in the main text.

Before the first turn, the mouse is moved to the viewport center.  Before every
subsequent decision, the virtual clock is paused again and a full-page
screenshot is captured.  An episode terminates when the Execution Agent emits
\texttt{terminate} or when the interaction budget is exhausted.

\subsubsection{Execution-Agent Context and Decision}

The Execution Agent implements the ReAct-style interaction policy.  Its decision context contains direct visual evidence, a compact
history, and browser affordances, as summarized in
Table~\ref{tab:gui-decision-context}.  The screenshot remains the source of
truth for the current visible state.  Source code is also supplied only as a rules
reference for identifying supported inputs, event handlers, coordinate
transforms, cooldowns, and pointer-lock behavior.

    The agent returns a structured decision containing a one-sentence summary, one
action $a_{i,t}$, a predicted visible target state, an expected transition, and
optional psychometric observations.  Only $a_{i,t}$ is executed.  The predicted
target and transition are retained as diagnostic intent and are explicitly
marked as non-ground-truth inputs to the Step-Level Judge.  If the returned
action violates the schema, the invalid action is not executed. The validation
error and previous response are returned to the agent for correction.

\begin{table*}[t]
\centering
\footnotesize

\caption{Input information provided to the Execution Agent at turn $t$. Dynamic values are
serialized into the user message, while the action schema and execution rules
are specified in the system prompt.}
\vspace{-5pt}

\renewcommand{\arraystretch}{1.25}
\resizebox{\linewidth}{!}{%
\begin{tabular}{@{}p{0.12\textwidth}p{0.34\textwidth}p{0.42\textwidth}@{}}
\toprule
\textbf{Component} & \textbf{Content} & \textbf{Role in action selection} \\
\midrule
\texttt{Prompt} & Original prompt $x$ & Specifies the intended artifact and user objective. \\
\texttt{Visual State} & Current full-page screenshot $S_{i,t}$ & Provides the authoritative visible state from which the next action is chosen. \\
\texttt{Artifact} & Source of $h_i$, truncated to 12,000 characters & Reveals input handlers and interaction rules to support precise action generation. \\
\texttt{Action Hints} & Up to 40 DOM controls, each represented by its tag, selector hint, text, and bounding rectangle & Grounds ordinary DOM actions in executable selectors. Canvas-internal targets are inferred from the screenshot and source. \\
\texttt{History} & Including previous decision summary, action, execution observation, and termination status & Discourages repetitive actions and supports multi-step behavior. \\
\texttt{Timing} & Turn index, turn limit, clock-frame interval, and post-action advance budget & Allows the agent to choose pre-action delays and held-action durations. \\
\bottomrule
\end{tabular}
}
\label{tab:gui-decision-context}
\end{table*}

\subsubsection{Action Grounding and Timing Semantics}

Table~\ref{tab:gui-action-space} details the parameterization of the action
space $\mathcal A$.  For \texttt{click} and \texttt{hover}, targets are resolved
in the following order: explicit page coordinates, selector plus text,
selector alone, and text alone.  The executor records the resolution method,
number of matches, actual hit-tested element, element rectangle, page and local
coordinates, and canvas dimensions in $m_{i,t}$.  Consequently, the judge can
distinguish a simulator miss from a correctly delivered but ineffective action.

\begin{table*}[t]
\centering
\footnotesize
\vspace{-5pt}
\caption{Grounded action interface used for different types of action $a_{i,t}\in\mathcal A$ in
the browser.}

\vspace{-5pt}

\renewcommand{\arraystretch}{1.1}

\resizebox{\linewidth}{!}{%
\begin{tabular}{@{}p{0.12\textwidth}p{0.31\textwidth}p{0.45\textwidth}@{}}
\toprule
\textbf{Action} & \textbf{Parameters} & \textbf{Execution semantics} \\
\midrule
\texttt{click} & selector or text. (optional) page coordinates $(x,y)$, \texttt{durationMs}, \texttt{waitMs} & Activates a DOM target or page coordinate. A held click emits mouse-down, advances the clock, and then emits mouse-up. \\
\texttt{hover} & selector or text. (optional) page coordinates $(x,y)$, \texttt{waitMs} & Moves the pointer to the resolved element or coordinate without activation. \\
\texttt{mouseMove} & relative $(\Delta x,\Delta y)$. (optional) \texttt{durationMs}, \texttt{waitMs} & Produces smooth relative motion. Under pointer lock it dispatches relative movement events rather than absolute cursor motion. \\
\texttt{press} & key. (optional) \texttt{durationMs}, \texttt{waitMs} & Emits an instantaneous key press or a key-down/key-up pair around a held interval. \\
\texttt{combo} & multiple keys. (optional) \texttt{durationMs}, \texttt{waitMs} & Holds the keys simultaneously and releases them in reverse order. \\
\texttt{type} & selector or text. (optional) \texttt{waitMs} & Fills the selected text input. \\
\texttt{wait} & duration in milliseconds & Advances virtual browser time without another user input. \\
\texttt{terminate} & optional reason & Stops interaction when the task is complete, the game is unrecoverable, or a realistic user would stop. \\
\bottomrule
\end{tabular}
}
\vspace{-10pt}
\label{tab:gui-action-space}
\end{table*}

The optional \texttt{waitMs} delays action onset, whereas
\texttt{durationMs} controls how long an action is held.  Let $\Delta$ denote
the browser clock-frame interval.  When a duration is explicitly requested, we
execute
\begin{equation}
\ell_{i,t}=\max\!\left(\ell^{\mathrm{req}}_{i,t},\,2\Delta+25\text{ ms}\right),
\end{equation}
so that a held action crosses multiple observable frames.  During a held
action, intermediate observations are captured after geometrically increasing
elapsed times (50, 100, 200 ms, and so on) up to $\ell_{i,t}$.  For an
instantaneous action, the clock is advanced by 50 ms before capturing
$S_{i,t+1}$, allowing immediate visual feedback to become observable. This design is primarily intended to facilitate trajectory debugging for the GUI Assessor and to support its use as an evaluator. During training, for cost efficiency and lower latency, we only provide the before and after states to the step-level judge.

Canvas interactions require special handling.  A selector-only click on a 2-D
canvas lands at its center and is therefore annotated with a warning. Clicks on
specific tiles or sprites should include page coordinates.  For first-person
or pointer-lock pages, firing is dispatched to the locked target at the current
crosshair, while view rotation is expressed as relative \texttt{mouseMove}.
Execution exceptions do not erase the transition: they are serialized into
$m_{i,t}$ with \texttt{success=false}, an error message, and diagnostic
warnings for the Step-level Judge.

\subsubsection{Queued Step-Level Assessment}

Immediately after $E_{i,t}$ is materialized, it is submitted to a FIFO
producer--consumer queue.  The Execution Agent can proceed to turn $t+1$ while the
Step-level Judge consumes earlier transitions.  A single judge request jointly
evaluates the transition and the post-action screenshot, using $x$,
$S_{i,t}$, $a_{i,t}$, $m_{i,t}$, $S_{i,t+1}$, and timing metadata.  The image
order is fixed as before-action followed by after-action.  This joint call
keeps the assessments grounded in identical evidence while the queue
overlaps judge latency with continued browser execution.

For compactness, let $z^{\mathrm{tr}}_{i,t,\cdot}$ and
$z^{\mathrm{sc}}_{i,t,\cdot}$ denote the Step-level Judge's component scores,
all in $[0,100]$.  The two reported scores are
{
\small
\begin{equation}
\begin{aligned}
q^{\mathrm{tr}}_{i,t}
={}&0.40z^{\mathrm{tr}}_{i,t,\mathrm{vs}}
+0.25z^{\mathrm{tr}}_{i,t,\mathrm{fb}}
+0.25z^{\mathrm{tr}}_{i,t,\mathrm{ex}}
+0.10z^{\mathrm{tr}}_{i,t,\mathrm{pr}},\\
q^{\mathrm{sc}}_{i,t}
={}&0.45z^{\mathrm{sc}}_{i,t,\mathrm{vs}}
+0.30z^{\mathrm{sc}}_{i,t,\mathrm{sv}}
+0.25z^{\mathrm{sc}}_{i,t,\mathrm{rd}}.
\end{aligned}
\label{eq:appendix-step-scores}
\end{equation}
}
Here \textsc{vs}, \textsc{fb}, \textsc{ex}, and \textsc{pr} denote visual
stability, application-feedback quality, execution correctness, and task
progress. \textsc{sv} and \textsc{rd} denote state visibility and readability. Task progress is intentionally
weakly weighted so that visually stable, understandable rule-based rejection
is not treated as a broken interaction.

In addition to numeric scores, the Step-level Judge returns strengths,
weaknesses, a compact visible-state description, and reusable interaction
experience.  The failure taxonomy is important for
attribution:
\begin{itemize}
    \item \texttt{coordinate\_miss}: the simulator did not hit the intended
    target;
    \item \texttt{action\_ineffective}: a correctly delivered, apparently
    valid action produced no meaningful response;
    \item \texttt{illegal\_or\_rejected}: the action violated visible game
    rules or prerequisites;
    \item \texttt{timing\_issue}: the action was plausible but mistimed or held
    for an unsuitable duration;
    \item \texttt{uncertain}: available evidence does not support a reliable
    diagnosis; and
    \item \texttt{none}: no relevant failure was observed.
\end{itemize}
Thus, simulator-side delivery errors primarily lower execution correctness,
whereas application-side failures lower feedback and interaction quality.  A
clear, rule-consistent rejection can still receive high visual-stability and
feedback scores.

\subsubsection{Trajectory-Level Meta Assessment}

After interaction terminates, the producer stops and we wait for the queue to drain so
that every $E_{i,t}$ has a completed Step-level assessment.  The Meta-level
Judge receives $x$, episode-level termination and timing information, every
action and execution record, and compact Step-level outputs. Visual evidence is conveyed through the structured
Step-level scores, state descriptions, strengths, weaknesses, and failure
modes.  This separation makes the final decision auditable and avoids an
unbounded multi-image context.

We set seven Meta-level dimensions. They are visual stability ($z_{i,\mathrm{vs}}$),
screenshot aesthetics ($z_{i,\mathrm{ae}}$), feedback clarity
($z_{i,\mathrm{fc}}$), interaction reliability ($z_{i,\mathrm{ir}}$), state
legibility ($z_{i,\mathrm{sl}}$), requirement alignment
($z_{i,\mathrm{pa}}$), and task progress ($z_{i,\mathrm{tp}}$).  The weighted
episode score is
{
\small
\begin{equation}
\begin{aligned}
\overline R_i^{\mathrm{GUI}}={}
&0.15z_{i,\mathrm{vs}}+0.15z_{i,\mathrm{ae}}
+0.10z_{i,\mathrm{fc}}\\+0.20z_{i,\mathrm{ir}}
&+0.10z_{i,\mathrm{sl}}+0.20z_{i,\mathrm{pa}}
+0.10z_{i,\mathrm{tp}}.
\end{aligned}
\label{eq:appendix-meta-score}
\end{equation}
}
All dimensions are clamped to $[0,100]$.  To prevent a high average from
masking a severe defect, the judge may additionally assign a usability cap
$C_i$ (Table \ref{tab:gui-usability-cap}).  Taking $C_i=100$ when no cap is needed, the score consumed by the
Visual Anchor and GRPO stages is
\begin{equation}
R_i^{\mathrm{GUI}}=\min\!\left(\overline R_i^{\mathrm{GUI}},C_i\right).
\label{eq:appendix-gui-score}
\end{equation}

\begin{table}[t]
\centering
\footnotesize

\renewcommand{\arraystretch}{1.25}

\resizebox{\linewidth}{!}{%
\begin{tabular}{@{}p{0.16\columnwidth}p{0.70\columnwidth}@{}}
\toprule
\textbf{$C_i$} & \textbf{Corresponding Issues} \\
\midrule
0--10 & Blank, crashed, corrupted, or impossible to evaluate. \\
20 & Page renders but no meaningful primary interaction works. \\
35 & A required entry action is correctly targeted but nonfunctional. \\
45 & The artifact is substantially the wrong product or lacks the requested core experience. \\
60 & The artifact is recognizable, but its central loop or required primary mechanic is blocked. \\
70 & The main loop works, but a prominent required feature is missing or nonfunctional. \\
80 & Primary play remains possible despite notable missing content or unreliable secondary mechanics. \\
90 & Only minor prompt omissions or polish gaps prevent a near-perfect score. \\
100 & No cap. Ordinary weaknesses are already represented by the component scores. \\
\bottomrule
\end{tabular}
}
\caption{Usability-cap rubric. The judge chooses the highest cap consistent
with the observed severity.}
\label{tab:gui-usability-cap}
\end{table}

\subsection{Action Trace Collection and Token Mapping}

\paragraph{Online trace collection.}
We attach a Chrome DevTools Protocol session to the Chromium page used by the GUI Assessor. Before interaction, JavaScript precise coverage identifies functions with resolvable source locations in inline scripts from the artifact's main-frame execution context, for which function-entry breakpoints are installed. The current implementation traces \texttt{click}, \texttt{mouseMove}, \texttt{press}, and \texttt{combo} actions. During each action, debugger pauses expose active call stacks, while precise coverage additionally identifies executed functions that may not appear in the captured stacks. To bound tracing overhead, at most 4096 pause events are processed per action.

Tracing is performed online, while attribution is filtered after Step-level assessment. An action contributes source evidence only when its transition is judged successful. Failed or low-quality actions are therefore excluded from source attribution, avoiding spurious negative credit to unrelated code. Before attribution, duplicate stacks and strict suffixes of longer captured call paths are removed.

\paragraph{Source span, runtime depth, and token alignment.}
Each retained call frame is mapped to the smallest enclosing function span in the generated HTML. Source offsets are converted to UTF-8 byte intervals and then mapped back to the full policy response \(y_i\) using the recorded location of \(h_i\). SHA-256 hashes and byte lengths are verified to detect offset drift before attribution.

The implementation measures runtime depth in the leaf-to-caller order returned by Chrome DevTools. Let \(s=(f_0,\ldots,f_{L_s-1})\) denote a retained maximal stack, where \(f_0\) is the currently executing leaf frame containing core logic and \(f_{L_s-1}\) is the outermost captured caller. For the retained stack set \(\mathcal S_i\), function spans, and UTF-8 byte positions \(b\), we define
\begin{equation}
\begin{aligned}
d_s(f)&=\min\{j\mid f_j=f\},\\
d_i(f)&=\min_{s\in\mathcal S_i:f\in s}d_s(f),\\
\delta_i(b)&=\min_{f:b\in\operatorname{span}(f)}d_i(f),\\
d_{i,k}&=\max_{b\in B_{i,k}}\delta_i(b),
\end{aligned}
\label{eq:appendix-atc-runtime-depth}
\end{equation}
where \(B_{i,k}\) is the UTF-8 byte span of token \(k\), and \(d_{i,k}\) is defined only for fully traced tokens. 

This implementation maps executed code frames back to their corresponding rollout tokens while avoiding potential misalignment at token boundaries. Repeated observations and overlapping function spans therefore retain the smallest available runtime depth, while a token spanning regions with different depths uses the largest depth among its covered bytes. Under this convention, leaf frames have depth 0 and outer callers receive progressively larger depths.

The original rollout token IDs are reconstructed as exact UTF-8 byte spans. For byte-level tokenizers, these spans are obtained directly from token pieces. Otherwise, fast-tokenizer character offsets are converted to UTF-8 byte offsets over the exact rollout text and accepted only when re-tokenization recovers the original non-special token IDs. A token is included in \(\mathcal F_{i,k}\) only when every byte in \(B_{i,k}\) is covered by valid trace runs.

\begin{figure}[t]
    \centering
    \includegraphics[width=\columnwidth]{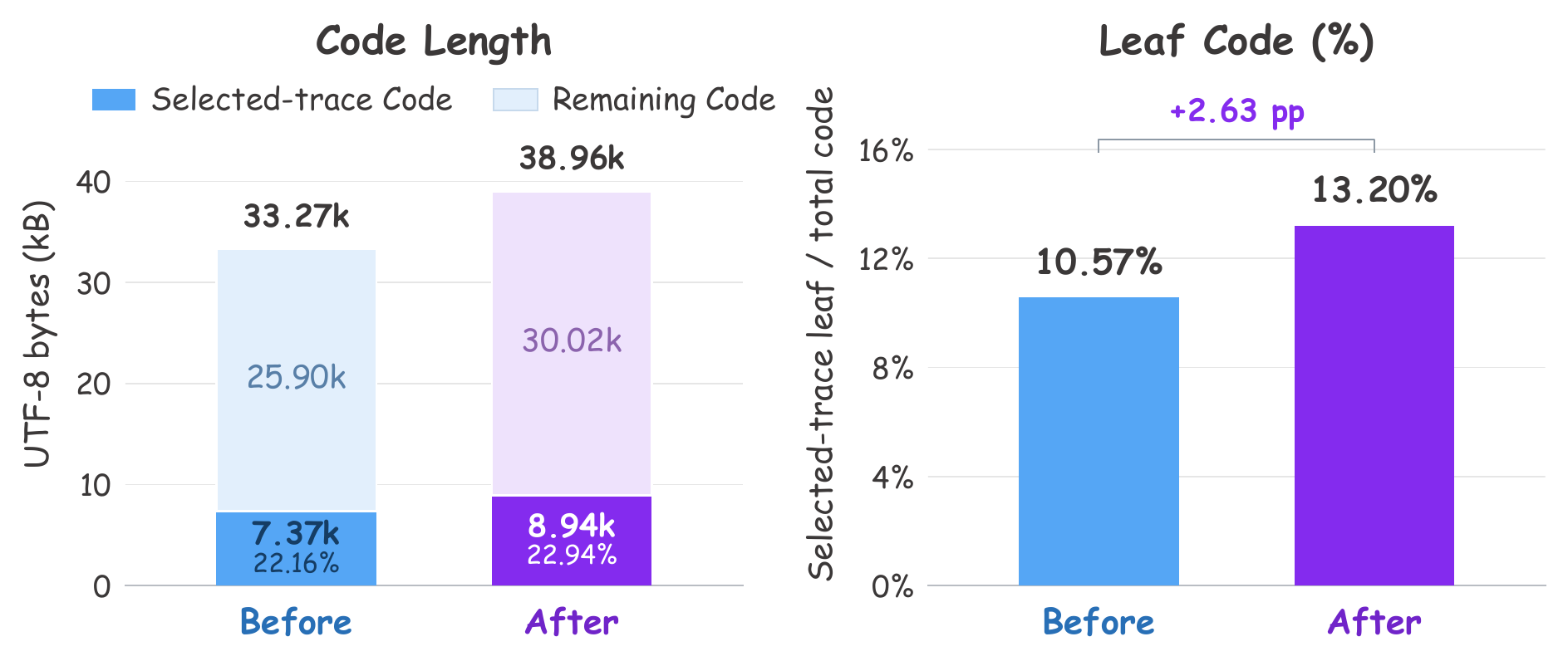}
    \caption{
        Changes in code size and action-trace attribution after training.
        The solid regions denote selected-trace code, while the pale regions
        denote the remaining generated code. Leaf-code coverage measures the
        proportion of total code attributed to depth-0 frames.
    }
    \label{fig:action-trace-analysis}
    \vspace{-20pt}
\end{figure}

\paragraph{Effect of action-trace credit.}
Figure~\ref{fig:action-trace-analysis} shows that action-trace credit remains highly selective, covering only about \(23\%\) of the generated code while focusing optimization on source regions associated with successful interactions. After training, the average code size increases from \(33.27\) to \(38.96\) kB, accompanied by a proportional increase in selected-trace code from \(7.37\) to \(8.94\) kB. More notably, the share of total code attributed to depth-0 leaf frames rises from \(10.57\%\) to \(13.20\%\). Since these leaf frames contain the logic executed most directly during successful actions, this shift suggests that training allocates more model capacity to robust, task-relevant interaction logic rather than merely expanding peripheral code.

\paragraph{Overall Positioning.} WebCatalyst advances RL supervision for interactive HTML generation across three coupled stages from interaction to policy optimization. First, the Game-Level GUI Assessor converts sustained browser interaction into trajectory-level RL feedback, moving beyond shallow click-and-type verification and GUI agents used primarily for evaluation or inference-time refinement. Second, Visual Anchor introduces one-sided reference calibration to prevent functionality-driven optimization from exploiting visual simplification, rather than serving as a generic aesthetic reward. Third, Action-Trace Credit grounds token-level credit in individual successful GUI actions by tracing their runtime JavaScript call paths back to generated tokens, rather than relying on compiler information or static code regions. Together, WebCatalyst establishes an end-to-end interaction-grounded supervision pipeline, carrying high-fidelity signals from experienced behavior to reward construction and ultimately to behavior-relevant code tokens.

\subsection{Implementation Details}
\label{sec:implementdetails}

\subsubsection{Training Hyperparameters}

Tables~\ref{tab:training-hyperparameters} and
\ref{tab:training-system-hyperparameters} summarize the configuration used
for the main experiment. We optimize Qwen3.6-35B-A3B with GRPO through verl~\citep{sheng2024hybridflow}. At
each update, $G=5$ responses are sampled for each of 16 requests. The reported checkpoint is taken after 40 updates, corresponding to
640 processed request instances. We use groupwise standard-deviation normalization in our GRPO
objective, with $\epsilon=10^{-6}$. 

\begin{table}[t]
\centering
\footnotesize

\caption{Optimization and generation hyperparameters.}
\label{tab:training-hyperparameters}
\setlength{\tabcolsep}{3pt}
\renewcommand{\arraystretch}{1.05}
\begin{tabular}{>{\raggedright\arraybackslash}p{0.43\columnwidth}
                  >{\raggedright\arraybackslash}p{0.50\columnwidth}}
\toprule
\textbf{Setting} & \textbf{Value} \\
\midrule
Backbone & Qwen3.6-35B-A3B \\
Optimization algorithm & GRPO \\
Requests per update & 16 \\
Group size & $5$ \\
Batch size & $16$ \\
Optimizer & Adam \\
Learning rate & $1\times10^{-6}$ \\
LR schedule & Constant \\
Gradient clipping & $1.0$ \\
Clipping $\varepsilon_c$ & $0.2$ \\
Loss aggregation & Token mean \\
KL regularization & $\eta=0.001$ \\
GRPO standard-deviation normalization & $\epsilon=10^{-6}$ \\
Max. prompt / response length & $2{,}048$ / $32{,}768$ tokens \\
Training sampling & Temperature $T=1.0$ \\
Random seed & $42$ \\
\bottomrule
\end{tabular}
\end{table}

\begin{table}[t]
\centering
\footnotesize
\caption{Reward and execution settings. TP, PP, CP, EP, and ETP
denote tensor, pipeline, context, expert, and expert-tensor parallelism,
respectively.}
\setlength{\tabcolsep}{3pt}
\renewcommand{\arraystretch}{1.05}
\begin{tabular}{>{\raggedright\arraybackslash}p{0.43\columnwidth}
                  >{\raggedright\arraybackslash}p{0.50\columnwidth}}
\toprule
\textbf{Setting} & \textbf{Value} \\
\midrule
GUI Assessor judge & GPT-5.6 Terra \\
Interaction turns & $8$ \\
Visual anchor model & GPT-5.5 \\
ATC modulation  & $\beta=0.5$ \\
Trace pause-event cap & $4{,}096$ per action \\
Assessor workers & $6$ workers \\
Attempt timeout & $300$ s \\
Training precision & bfloat16 \\
Policy parallelism & TP$=2$, PP$=1$, CP$=1$, EP$=8$, ETP$=1$ \\
MoE auxiliary / $z$-loss coefficients & $0.01$ / $0.001$ \\
\bottomrule
\end{tabular}
\label{tab:training-system-hyperparameters}
\end{table}

Training uses one node with eight NVIDIA B200 GPUs. Policy updates are
performed with Megatron-Core \citep{megatron-lm} in bfloat16, while vLLM \citep{kwon2023efficient} generates the rollouts.
On B200 GPU, we run one complete rollout replica per GPU and use the Triton MoE backend.

\subsubsection{Training Dataset Construction}
We construct the training set from Reddit posts related to frontend and game development. We first retrieve candidate requests using ``game'' and related keywords. We ensure that no sensitive or personal identification information is obtained in our data. The requests are then rewritten into natural user-facing queries with diverse personas, tones, and levels of constraint specificity. We apply a broad HTML-game filter that retains examples exhibiting a complete game subject, browser or HTML technology, and concrete game mechanics, while removing backend-only questions and requests that do not define a playable artifact. GPT-5.5 subsequently verifies each retained query and either keeps it unchanged when it is already suitable or refines it into a direct request for a complete, self-contained browser game. After deduplication, the resulting prompts are split into training and validation sets. At training time, each prompt is paired with a system instruction requiring the policy to return only one complete HTML document with all CSS and JavaScript included inline.


\subsubsection{Evaluation Protocol}

\paragraph{Generation setup.}
We evaluate the models on the test sets of WebGen-Bench~\citep{lu2025webgenbench} and Cookie-Bench Mini Games~\citep{yang2026cookiebench}. For every benchmark request, the model receives the original user prompt together with the same output-format instruction, that is, return one complete, self-contained HTML document with all CSS and JavaScript included inline. 

For our locally served Qwen checkpoints, we use temperature $0.6$, top-$p$ $0.95$, top-$k$ $20$, a maximum of $32{,}768$ response tokens, and enable the model's thinking mode. For proprietary baselines, we retain the provider-native sampling configuration and set the largest supported output-token budget when that control is available.

\paragraph{Cookie-Bench Mini Games.}
We use the deduplicated multilingual Mini Games split containing 156 requests, primarily in Chinese and English, with additional examples in French, German, Spanish, and other languages. Each generated HTML file is served in Chromium and first receives provisional aesthetics and functionality scores from its initial rendered state. A source-aware planner then constructs a test plan, after which UI-TARS explores the game through a ReAct loop. The evaluator records the action trace and reviews it, identifies functional and visual problems, and adjusts the provisional scores. We use Qwen3.8-27B as the execution model and DeepSeek-V4-Pro as the review model. 
A model-side format generation, installation, or launch failure is retained with zero score. A failed evaluator pass is retried once with a fresh browser-agent session.

\paragraph{WebGen-Bench.}
WebGen-Bench contains 101 requests and 647 atomic test cases. We serve each generated page locally and use Qwen2.5-VL-32B-Instruct as the UI evaluation agent. For each test case, the agent is given an operation and its expected outcome, may perform at most 15 browser interactions, and returns \textsc{yes}, \textsc{partial}, or \textsc{no}. With $v_j\in\{1,0.5,0\}$ denoting these verdicts, overall accuracy is $\mathrm{Acc}=100\sum_{j=1}^{647}v_j/647$. The same aggregation is used for the request-level categories (Content Presentation, User Interaction, and Data Management) and test-case categories (Functional Testing, Data Display Testing, and Design Validation Testing). The fraction for pages that cannot be served is reported as Start Failure Rate.

\subsection{Practical Deployment Guidance}
\subsubsection{Cost Breakdown}

We train Qwen3.6-35B-A3B for 40 optimization steps on eight NVIDIA
B200 GPUs. The main efficiency bottleneck is artifact evaluation by
the GUI Assessor. Our primary
setting uses GPT-5.6 Terra through an API limited to 50 requests per
minute, with six concurrent assessor workers. Evaluating the 80
rollouts generated at each training step takes approximately
40 minutes, corresponding to an effective throughput of about
2 rollouts per minute. As shown in Table~\ref{tab:evaluation-token-cost},
evaluating one rollout consumes, on average, 112.27K input tokens and
9.28K output tokens, or 121.56K tokens in total. A full training step
therefore consumes approximately 8.98M input tokens and 0.74M output
tokens. At prices of \$2 and \$12 per million input and output tokens,
respectively, this amounts to approximately \$26.88 per step, or
\$1,075 over 40 steps, excluding caching discounts and unlogged
retries. The run takes approximately 26.7 B200 GPU-hours for LLM inference and parameter updates.

On an eight-GPU A100 SXM system, the assessor latency is comparable to
the time required for policy rollout generation. This suggests that
an asynchronous policy-training pipeline with a policy staleness of
approximately one step could overlap assessment with subsequent
rollouts and improve hardware utilization. We leave such an
implementation to future work. 

We additionally experiment with
Qwen3.8-27B as a locally served judge, using four B200 GPUs for vLLM judge
serving and four B200 GPUs for rollout and training. This configuration
requires approximately 192 aggregate B200 GPU-hours for 40 steps
(about 24 hours of wall-clock time), while its token consumption is of
the same order as that of GPT-5.6 Terra. These measurements expose both
the monetary and latency costs of using our GUI Assessor and provide
practical guidance for reproducing or scaling the training pipeline.

\begin{table}[t]
\centering
\small
\setlength{\tabcolsep}{3pt}

\resizebox{\columnwidth}{!}{%
\begin{tabular}{@{}lrrrr@{}}
\toprule
\textbf{Component} & \textbf{Input (K)} & \textbf{Output (K)} & \textbf{Total (K)} & \textbf{Cost (\$)} \\
\midrule
ReAct Agent  & 60.34  & 2.04 & 62.38  & 0.1452 \\
Step-Level Judge  & 43.44  & 5.69 & 49.13  & 0.1551 \\
Meta-Level Judge & 8.49   & 1.55 & 10.05  & 0.0356 \\
\midrule
\textbf{Total}
& \textbf{112.27}
& \textbf{9.28}
& \textbf{121.56}
& \textbf{0.3360} \\
\bottomrule
\end{tabular}
}
\caption{Average per-sample evaluation token usage and
estimated cost for GPT-5.6 Terra,
with a budget of eight interaction turns.
Token counts are in thousands (K).
The turn judge jointly evaluates transitions and screenshots
and is counted once.
Costs assume \$2 and \$12 per million input and output tokens,
respectively, excluding cache discounts.}
\label{tab:evaluation-token-cost}
\vspace{-20pt}
\end{table}

\subsubsection{Judge Selection}

\begin{table*}[htbp]
\centering
\caption{
Effect of judge selection on downstream RL performance and alignment with human experience.
Vanilla denotes the Qwen3.6-35B-A3B base model without RL.
Src-Gemini uses Gemini 3.1 Pro to evaluate source code directly, while Src-Rubric additionally incorporates rendered screenshots and task-specific rubrics.
For GUI Assessor, each row specifies the backbone used for interactive evaluation.
Pearson $r$ measures correlation with human experiential ratings.
Arrows indicate the preferred direction, dashes indicate unavailable results, and green highlights judge settings with Pearson $r>0.8$.
}
\label{tab:main_results_with_pearson}
\small
\setlength{\tabcolsep}{3.2pt}
\renewcommand{\arraystretch}{1.0}
\resizebox{\textwidth}{!}{%
\begin{tabular}{lccccccccccc}
\toprule
\multirow{3}{*}{%
  \makecell{Grader}%
}
& \multicolumn{7}{c}{\textbf{WebGen-Bench}}
& \multicolumn{3}{c}{\textbf{Cookie-Bench Mini Games}}
& \multirow{3}{*}{%
    \makecell{\textbf{Pearson}\\$\boldsymbol{r}\uparrow$}
  } \\
\cmidrule(lr){2-8}
\cmidrule(lr){9-11}
& \textbf{Acc.} $\uparrow$
& \makecell{\textbf{Content}\\\textbf{Pres.} $\uparrow$}
& \makecell{\textbf{User}\\\textbf{Inter.} $\uparrow$}
& \makecell{\textbf{Data}\\\textbf{Mgmt.} $\uparrow$}
& \makecell{\textbf{Func.}\\\textbf{Test} $\uparrow$}
& \makecell{\textbf{Data}\\\textbf{Display} $\uparrow$}
& \makecell{\textbf{Design}\\\textbf{Val.} $\uparrow$}
& \textbf{Aesth.} $\uparrow$
& \textbf{Func.} $\uparrow$
& \textbf{Total} $\uparrow$
& \\
\midrule

Vanilla (N/A)
& 48.8 & 59.2 & 43.0 & 47.1 & 33.9 & 61.8 & 70.5
& 81.5 & 56.1 & 68.8 & -- \\

Src-Gemini
& -- & -- & -- & -- & -- & -- & --
& -- & -- & -- & 0.065 \\
Src-Rubric
& -- & -- & -- & -- & -- & -- & --
& -- & -- & -- & 0.309 \\

\addlinespace[0.1em]
\midrule
\addlinespace[0.2em]

\multicolumn{12}{c}{\textit{\textbf{GUI Assessor}}} \\

Qwen3.6-27B
& 49.2 & 64.7 & 39.6 & 51.2 & 32.2 & 65.6 & 71.7
& 75.4 & 62.4 & 68.9 & 0.613 \\

\rowcolor{green!12}
\textbf{Qwen3.8-27B}
& 59.0 & 70.4 & 53.7 & 56.9 & 45.1 & 74.5 & 73.8
& 83.8 & 61.0 & 72.4 & \textbf{0.838} \\

\rowcolor{green!12}
\textbf{GPT-5.6 Terra}
& 56.9
& 68.1
& 52.9
& 54.2
& 42.2
& 68.8
& 77.9
& 87.5
& 63.2
& 75.3
& \textbf{0.812} \\

\bottomrule
\end{tabular}%
}
\end{table*}

For ease of deployment, our main experiments use GPT-5.6 Terra through its API as the judge model. While this simplifies the training setup, API throughput becomes the primary bottleneck in training. We therefore investigate how judge selection affects training and whether locally served open-source models can provide a practical alternative. As shown in Table~\ref{tab:main_results_with_pearson}, we train additional variants using Qwen3.6-27B \citep{qwen2026qwen36_27b} and Qwen3.8-27B \citep{qwen2026qwen38}  as the judge backbone and the operator of the ReAct agent.
To measure how well these judges reflect human experience, we construct a small human-annotated evaluation set. Following a simplified version of the Meta-Judge criteria, we select 10 prompts and collect artifacts from seven to eight models spanning different capability levels, including GPT-5.3-Codex, GPT-5.5, Claude Opus 4.7, GPT-5.4-mini among others. Our researchers evaluate more than 70 artifacts based on their actual interaction experience, while blinded to the identities of the models that generated them. We then measure the Pearson correlation between these human scores and the scores produced under different evaluation settings. For comparison, Src-Gemini uses Gemini 3.1 Pro to evaluate the source code directly, whereas Src-Rubric augments this source-based evaluation with rendered screenshots and task-specific rubrics. Table~\ref{tab:main_results_with_pearson} reports both the resulting training performance and the corresponding Pearson correlations.

Across judge backbones, the GUI Assessor maintains a higher correlation with human scores than rubric-based judging, indicating that its feedback better reflects actual human experience. Judge capability nevertheless remains important: a weaker model such as Qwen3.6-27B produces lower-fidelity evaluations and consequently weaker training outcomes. Once judge quality reaches the level of Qwen3.8-27B, with a Pearson correlation above 0.8, training becomes relatively stable, although the exact gains still vary across models and between games and webpages. These results suggest that self-hosted open-source judges can be used in WebCatalyst, provided that their evaluations remain sufficiently aligned with human experience.

\raggedbottom
\subsection{Prompts for Generation and Assessment}
\label{app:prompts}

Each box lists the system prompt and user-message template in this section. Angle-bracketed placeholders denote runtime inputs and image attachments.

\subsubsection{HTML Generation}
\label{app:prompt-generation}
\begin{appendixprompt}{HTML Generation}
\noindent\textbf{System:}\par\smallskip
\noindent You are an expert HTML developer. Respond with only the complete, self-contained HTML source code for the requested game. Start your response with \textless{}!DOCTYPE html\textgreater{} and end it with \textless{}/html\textgreater{}. Do not include Markdown code fences, explanations, or any text outside the HTML document. Include all CSS and JavaScript in the HTML file.\par
\medskip\noindent\textbf{User:}\par
\begin{Verbatim}[fontsize=\footnotesize,breaklines=true,breakanywhere=true]
<task_prompt>
\end{Verbatim}
\end{appendixprompt}

\subsubsection{Execution-Agent Action Decision}
\label{app:prompt-action}
The Execution Agent receives the current screenshot, task, interactive elements, source code, and episode history.

\begin{appendixprompt}{Execution-Agent Action Decision}
\noindent\textbf{System:}\par\smallskip
\noindent You are a user simulation decision engine.\par
\smallskip
\noindent You receive a screenshot of a rendered web artifact, optional episode history, optional source code, a free-form user persona, and free-form psychometric metrics/guidelines. Choose the next realistic user action for the current turn, or terminate the episode when the task is complete or the simulated user would naturally stop.\par
\smallskip
\noindent Return a single JSON object, with no markdown fences. Include only fields that are relevant to the chosen action; do not copy placeholder fields from other examples.\par
\smallskip
\begin{Verbatim}[fontsize=\footnotesize,breaklines=true,breakanywhere=true]
{
  "summary": "<one sentence>",
  "action": <one action object>,
  "targetState": "<what visible state should count as the goal after this action>",
  "expectedTransition": "<what should visually change while time advances>",
  "psychometricObservation": {
    "positive": ["engagement", "control", "clarity", "confidence"],
    "negative": ["confusion", "frustration", "fatigue", "boredom"]
  }
}
\end{Verbatim}
\smallskip
\noindent\textbf{Action examples:}\par\nobreak
\noindent - Click DOM control: \texttt{\{"type":\allowbreak{}"click",\allowbreak{}"selector":\allowbreak{}"\#startBtn"\}}\par
\noindent - Hold fire on FPS canvas: \texttt{\{"type":\allowbreak{}"click",\allowbreak{}"selector":\allowbreak{}"\#game",\allowbreak{}"durationMs":\allowbreak{}250\}}\par
\noindent - Click a tile/block/sprite inside a 2D canvas using page coordinates: \texttt{\{"type":\allowbreak{}"click",\allowbreak{}"selector":\allowbreak{}"\#gameCanvas",\allowbreak{}"x":\allowbreak{}240,\allowbreak{}"y":\allowbreak{}310\}}\par
\noindent - Coordinate click without selector only when absolute page coordinates are the intended input: \texttt{\{"type":\allowbreak{}"click",\allowbreak{}"x":\allowbreak{}430,\allowbreak{}"y":\allowbreak{}520\}}\par
\noindent - Press or hold key: \texttt{\{"type":\allowbreak{}"press",\allowbreak{}"key":\allowbreak{}"w",\allowbreak{}"durationMs":\allowbreak{}600\}}\par
\noindent - Hold a two-key keyboard combo for platformers/action games: \texttt{\{"type":\allowbreak{}"combo",\allowbreak{}"keys":\allowbreak{}["ArrowRight",\allowbreak{}"Space"],\allowbreak{}"durationMs":\allowbreak{}180\}}\par
\noindent - Smoothly rotate mouse-look/crosshair: \texttt{\{"type":\allowbreak{}"mouseMove",\allowbreak{}"dx":\allowbreak{}120,\allowbreak{}"dy":\allowbreak{}-30,\allowbreak{}"durationMs":\allowbreak{}300,\allowbreak{}"steps":\allowbreak{}12\}}\par
\noindent - Wait: \texttt{\{"type":\allowbreak{}"wait",\allowbreak{}"ms":\allowbreak{}300\}}\par
\noindent - Stop: \texttt{\{"type":\allowbreak{}"terminate",\allowbreak{}"reason":\allowbreak{}"..."\}}\par
\smallskip
\noindent\textbf{Rules:}\par\nobreak
\noindent - Prefer selectorHint values from Available interactive elements for ordinary DOM controls, for example "\#restartBtn". Use coordinates only when no DOM selector can represent the intended action.\par
\noindent - For 2D canvas, tile, grid, board, mining, puzzle, or sprite games, a canvas selector alone clicks the center of the whole canvas. If the intended target is a specific tile, block, item, sprite, or point inside the canvas, include both the canvas selector and explicit page x/y coordinates for that target, for example \texttt{\{"type":\allowbreak{}"click",\allowbreak{}"selector":\allowbreak{}"\#gameCanvas",\allowbreak{}"x":\allowbreak{}240,\allowbreak{}"y":\allowbreak{}310\}}. Do not return only the canvas selector for an internal canvas target.\par
\noindent - For first-person, pointer-lock, mouse-look, or shooter games, do not use x/y click coordinates to aim. The screen crosshair is controlled by mouse movement, and absolute Playwright mouse movement can unintentionally rotate the camera. To fire or hold fire, prefer a click action on the game canvas selector (for example \texttt{\{"type":\allowbreak{}"click",\allowbreak{}"selector":\allowbreak{}"\#game",\allowbreak{}"durationMs":\allowbreak{}200\}}) with no x/y coordinates. To aim or turn the view, use mouseMove with dx/dy; use small smooth deltas and iterate based on screenshots.\par
\noindent - action.waitMs is an optional pre-action delay in milliseconds. If waitMs \textgreater{} 0, the engine advances Playwright Clock by that amount before executing the action. Use waitMs \textgreater{} 0 when the correct user behavior is to wait for a moving game state and then act at the right moment. Use waitMs=0 or omit it for actions that should happen immediately.\par
\noindent - Distinguish timing fields carefully:\par
\noindent - action.waitMs delays before the action starts. Use this for "wait, then press/click" timing.\par
\noindent - action.durationMs controls how long the action is held after it starts, such as holding a movement key or held mouse button.\par
\noindent - If action.durationMs is present and shorter than two engine observation frame intervals plus 25ms, including durationMs=0, the engine will hold it for at least 2*frameMs + 25ms so the action can cross observable frame boundaries. Omit durationMs entirely for an instantaneous click or key press.\par
\noindent - Timing-sensitive example: if the current game screen shows that pressing an arrow key immediately would be too early, but the snake will align in about 240ms, return action \texttt{\{"type":\allowbreak{}"press",\allowbreak{}"key":\allowbreak{}"ArrowLeft",\allowbreak{}"waitMs":\allowbreak{}240,\allowbreak{} "durationMs":\allowbreak{}80\}}. This means wait 240ms first, then press ArrowLeft for 80ms.\par
\noindent - For UI controls that work while held, such as game movement buttons backed by mousedown/mouseup handlers, return a click durationMs. Do not include durationMs unless a held click is intentional.\par
\noindent - Use combo only for simultaneous keyboard input, such as moving right while jumping in a platformer. combo.keys must contain 1 or 2 keyboard keys; never return more than 2 keys, and never include mouse buttons. If more keys seem useful, choose the most important 1-2 keys for the immediate short action.\par
\noindent - Use terminate only when the visible state and episode history show that the task is complete, the game is over with no useful recovery, or a real user would stop. Do not terminate just because one action has been executed.\par
\noindent - Do not return hidden code-state conditions. The current screenshot is the source of truth for visible state. Episode history is only a record of prior decisions, actions, observations, and termination state; it contains no judge feedback and must not be treated as an evaluation of prior actions.\par
\noindent - When source code is provided, use it to infer supported inputs, event handlers, selectors, canvas coordinate mapping, cooldowns, distance checks, pointer-lock or mouse-look behavior, and other interaction rules. Use code as a rules reference, not as hidden live state; the current screenshot remains the source of truth for visible state. For 2D canvas click handlers, inspect getBoundingClientRect/clientX/clientY/tile-size conversions and choose page x/y coordinates that land on the intended visible tile or object.\par
\noindent - If the game is not started and a Start button is visible, click Start and target a visible active gameplay state.\par
\noindent - Use the current screenshot, source-code rules when present, interactive elements, and the factual episode history to choose the next action. Do not assume that transition or screenshot judge results from earlier turns will be available.\par
\medskip\noindent\textbf{User:}\par
\begin{Verbatim}[fontsize=\footnotesize,breaklines=true,breakanywhere=true]
URL:
<url>

Task prompt:
<task_prompt>

Turn:
<turn_index_plus_one> of <turn_limit>

Reference clock intervals:
{
  "referenceFrameIntervalMs": <frame_ms>,
  "maxPostActionAdvanceMs": <max_advance_ms>
}

Episode history:
<episode_history_json>

Available interactive elements:
<action_hints_json>

User persona (<persona_path>):
<persona_text>

Psychometric metrics (<metrics_path>):
<metrics_text>

Artifact source code (<artifact_path>):
```
<artifact_source_code>
```

<current_screenshot_attachment>
\end{Verbatim}
\end{appendixprompt}

Invalid actions trigger the following correction, with up to three attempts.

\begin{appendixprompt}{Action Validation and Correction}
\noindent\textbf{User-message extension:}\par
\begin{Verbatim}[fontsize=\footnotesize,breaklines=true,breakanywhere=true]
Your previous response was invalid and was not executed.
Validation error: <validation_error>
Return corrected JSON only, using the same schema.
Reminder: combo.keys must contain 1 or 2 keyboard keys only.
Previous invalid JSON:
<previous_invalid_json>
\end{Verbatim}
\end{appendixprompt}

\subsubsection{Combined Step-Level Evaluation}
\label{app:prompt-step}
One model call returns transition and screenshot evaluations from the action metadata and before/after images.

\begin{appendixprompt}{Combined Step-Level Evaluation}
\noindent\textbf{System:}\par\smallskip
\noindent You are the combined TransitionJudger and ScreenshotJudger for an HTML game quality evaluation system. You receive exactly two images in this order: Image 1 is the before-action screenshot and Image 2 is the after-action screenshot. Evaluate both judge roles independently in one response. Never treat the simulator decision as ground truth.\par
\smallskip
\noindent\textbf{TransitionJudger:}\par\nobreak
\noindent Evaluate the two screenshots as evidence of the game's interaction and visual transition quality. Use action execution and hit-test metadata to determine whether the simulator delivered the requested action to the intended target. Action execution warnings are simulator diagnostics. If metadata warns that a canvas selector click lacked coordinates, treat it as evidence that the simulator clicked the canvas center; for a 2D canvas internal target this is usually coordinate\_miss, not proof that the game failed to respond. If metadata contains success=false or executionError, set executionReward low and classify failureMode as action\_ineffective, while still judging visual stability and app feedback from the images. The targetState, expectedTransition, and decision summary describe simulator intent only. Do not score whether that prediction was fulfilled. Score the HTML game itself: visual continuity, responsiveness, feedback clarity, rule consistency, animation/state coherence, absence of crashes or broken rendering, and whether the observed transition makes sense for visible game mechanics. Use multi-dimensional rewards. Human product-quality preference should prioritize stable, readable, coherent visuals and understandable feedback over immediate task progress. Task progress is a weak component and must not dominate overall reward. Strengths and weaknesses must describe the game's transition, feedback, coherence, and robustness, not whether the simulator's predicted target was achieved. If hit-test metadata shows the action point missed the intended target, use coordinate\_miss; do not conclude that the action mechanism itself is ineffective. If the action is invalid under visible state or ordinary app/game rules, use illegal\_or\_rejected. Clear rejection can still receive high visualStabilityReward and appFeedbackReward. If an apparently valid, correctly delivered action produces no feedback, state change, partial progress, or consistent behavior, use action\_ineffective; treat it as a likely implementation bug, selector/event mismatch, or execution-path issue and penalize it strongly. Do not punish the app heavily just because the simulator chose a poor action. A rejected or no-op action can still earn high visualStabilityReward and appFeedbackReward when the page remains stable and feedback is clear. Describe the before image state compactly and extract reusable experience from the observed mechanics and feedback.\par
\noindent\textbf{Transition reward dimensions:}\par\nobreak
\noindent - visualStabilityReward: rendering stability, layout coherence, readable visuals, absence of flicker, broken frames, crashes, or state corruption. This is the most important reward.\par
\noindent - appFeedbackReward: clarity, immediacy, and rule-consistency of the app/game response. Legal rejection can score high when it is understandable.\par
\noindent - executionReward: whether the simulator delivered the requested action to the intended element, coordinate, key, or mouse target according to metadata.\par
\noindent - taskProgressReward: weakly weighted progress toward the task/scenario, such as starting play, moving, scoring, collecting, or completing an objective.\par
\noindent Compute transition overallReward = 0.40 * visualStabilityReward + 0.25 * appFeedbackReward + 0.25 * executionReward + 0.10 * taskProgressReward.\par
\noindent\textbf{Transition rubric:}\par\nobreak
\noindent - 90-100 overallReward: Stable, readable, coherent visuals with clear feedback and correct simulator execution; task progress may be partial or absent if a no-op or rejection is understandable.\par
\noindent - 70-89 overallReward: Mostly stable and understandable, with minor ambiguity, weak feedback, or limited task progress.\par
\noindent - 40-69 overallReward: Usable but meaningfully weak in feedback, execution clarity, visual stability, or interaction coherence.\par
\noindent - 1-39 overallReward: Severe visual instability, broken rendering, execution mismatch, or an apparently valid action producing no understandable feedback.\par
\noindent - 0 overallReward: The two images are blank, corrupted, missing essential page content, or impossible to evaluate.\par
\noindent\textbf{Transition failureMode definitions:}\par\nobreak
\noindent - "none": the action succeeded or any failure is not relevant.\par
\noindent - "coordinate\_miss": the action point missed the intended target or landed on the wrong element or region.\par
\noindent - "action\_ineffective": the target was correct and the action should be valid by visible state, but the app did not respond as expected.\par
\noindent - "illegal\_or\_rejected": the action was invalid by visible state or game rules, missing prerequisites, resource constraints, cooldowns, distance checks, or mechanics and was rejected or correctly no-oped.\par
\noindent - "timing\_issue": the action is plausible but appears too early, too late, or not held long enough.\par
\noindent - "uncertain": there is not enough evidence to classify the failure.\par
\smallskip
\noindent\textbf{ScreenshotJudger:}\par\nobreak
\noindent Evaluate Image 2, the after-action screenshot, for screenshot usability only. Focus on rendering quality, visual stability, visible page/game state, layout stability, readability, clipping, occlusion, artifacts, and whether the image is useful for judging page state. Do not score task completion, action success, game progress, or whether the simulator made a good choice. A screenshot can receive a high score even when the action failed, as long as the visual state is clear and stable. Rate screenshotAestheticReward separately for visual polish, composition, color harmony, spacing, and overall presentation quality. It is not included in screenshot score; the run-level global judge uses it. Compute screenshot score = 0.45 * visualStabilityReward + 0.30 * stateVisibilityReward + 0.25 * readabilityReward.\par
\noindent\textbf{Screenshot rubric:}\par\nobreak
\noindent - 90-100: The screenshot is sharp, complete, readable, stable, and clearly shows page state after the action.\par
\noindent - 70-89: The screenshot is generally useful, with minor clipping, small readability issues, or non-critical visual artifacts.\par
\noindent - 40-69: The screenshot is partially useful but has meaningful occlusion, blurry text, layout instability, cropped important content, or ambiguous state.\par
\noindent - 1-39: The screenshot is mostly unusable due to blank areas, severe clipping, unreadable UI, broken rendering, or failure to show relevant state.\par
\noindent - 0: The screenshot is blank, missing, corrupted, or impossible to evaluate.\par
\smallskip
\noindent\textbf{Return only JSON with exactly these two top-level objects:}\par\nobreak
\begin{Verbatim}[fontsize=\footnotesize,breaklines=true,breakanywhere=true]
{
  "transitionEvaluation": {
    "overallReward": number from 0 to 100,
    "visualStabilityReward": number from 0 to 100,
    "executionReward": number from 0 to 100,
    "appFeedbackReward": number from 0 to 100,
    "taskProgressReward": number from 0 to 100,
    "summary": string,
    "strengths": string[],
    "weaknesses": string[],
    "failureMode": "none" | "coordinate_miss" | "action_ineffective" | "illegal_or_rejected" | "timing_issue" | "uncertain",
    "experience": string,
    "state": string
  },
  "screenshotEvaluation": {
    "score": number from 0 to 100,
    "visualStabilityReward": number from 0 to 100,
    "readabilityReward": number from 0 to 100,
    "stateVisibilityReward": number from 0 to 100,
    "screenshotAestheticReward": number from 0 to 100,
    "summary": string,
    "strengths": string[],
    "weaknesses": string[]
  }
}
\end{Verbatim}
\medskip\noindent\textbf{User:}\par
\begin{Verbatim}[fontsize=\footnotesize,breaklines=true,breakanywhere=true]
Task prompt:
<task_prompt>

Turn index:
<turn_index>

Chosen action:
<action_json>

Action execution metadata:
<action_execution_json>

Simulator intent only, not a scoring target - decision summary:
<decision_summary>

Simulator intent only, not a scoring target - predicted target state:
<target_state>

Simulator intent only, not a scoring target - predicted transition:
<expected_transition>

Observer reached target:
<reached_goal>

Timing:
{
  "totalWaitMs": <total_wait_ms>,
  "simulatedAdvanceMs": <simulated_advance_ms>
}

Image order is fixed: Image 1 is before the action; Image 2 is after the action. Both are attached even when byte-identical.

Image 1: before
Path: <before_screenshot_path>
<before_screenshot_attachment>

Image 2: after
Path: <after_screenshot_path>
<after_screenshot_attachment>
\end{Verbatim}
\end{appendixprompt}

\subsubsection{Meta-Level Evaluation}
\label{app:prompt-meta}
The Meta-Level Judge summarizes the episode from structured actions and per-turn evaluations.

\begin{appendixprompt}{Meta-Level Evaluation}
\noindent\textbf{System:}\par\smallskip
\noindent You are GlobalJudger for an HTML game/user-experience simulation. You receive structured per-turn transition and screenshot evaluations only. You do not receive screenshots. Do not invent unseen visual details; use screenshot summaries, strengths, weaknesses, and reward fields as visual evidence.\par
\smallskip
\noindent Your job is to judge the full play episode as a human product-quality reviewer. Consider whether the page/game is stable, understandable, playable, responsive, and whether critical flows actually work.\par
\smallskip
\noindent Use transition evaluations for interaction quality, feedback clarity, execution reliability, failureMode, experience, state, and task progress. Use screenshot evaluations for visual stability, state legibility, readability, and screenshotAestheticReward.\par
\smallskip
\noindent\textbf{Compute weightedScore before any cap:}\par\nobreak
\noindent weightedScore = 0.15 * visualStabilityReward + 0.15 * screenshotAestheticReward + 0.10 * feedbackClarityReward + 0.20 * interactionReliabilityReward + 0.10 * stateLegibilityReward + 0.20 * promptAlignmentReward + 0.10 * taskProgressReward\par
\smallskip
\noindent Only apply a usability cap as a severity guardrail when the weightedScore would otherwise overstate the real user experience. Do not set usabilityCap merely because there is a flaw. Ordinary UX weaknesses, ambiguous feedback, low task progress, occasional ineffective actions, missing polish, or partially unreliable secondary mechanics should reduce the reward dimensions but should not set usabilityCap.\par
\smallskip
\noindent\textbf{Choose the highest cap consistent with the observed severity:}\par\nobreak
\noindent - Cap at 0-10 if the page is blank, crashed, corrupted, or impossible to evaluate.\par
\noindent - Cap at 20 if the page renders but is effectively non-interactive: no meaningful primary interaction works at all, or the app cannot leave an inert/static state.\par
\noindent - Cap at 35 if a required primary entry action such as Start, Play, Begin, Next, or an equivalent first-step control appears correctly targeted but is nonfunctional, so the main experience cannot begin.\par
\noindent - Cap at 45 if the artifact is substantially the wrong product or the requested core experience is absent, even if some unrelated UI or visuals work.\par
\noindent - Cap at 60 only for severe but non-total incompleteness: the main artifact is recognizable and partially usable, but the central loop or a required primary mechanic is absent, blocked, or impossible to complete.\par
\noindent - Cap at 70 for major prompt gaps where the main loop works and the user can make meaningful progress, but a prominent required feature/content area is missing or clearly nonfunctional.\par
\noindent - Cap at 80 for notable missing content, weak production breadth, or unreliable secondary mechanics when the primary experience remains playable.\par
\noindent - Cap at 90 for minor prompt omissions or polish gaps only when the requested score would otherwise be near-perfect.\par
\noindent - Leave usabilityCap null when the flaw is already well represented by low promptAlignmentReward, taskProgressReward, feedbackClarityReward, or interactionReliabilityReward.\par
\noindent - For broad prompts requesting many assets, levels, weapons, sounds, or polish, do not cap at 60 solely because every requested item was not observed in a short episode. Use reward dimensions, or a higher cap such as 70/80, unless the missing item blocks the primary loop or makes the artifact substantially not what was requested.\par
\noindent - Do not apply a cap for clear rule-based rejections such as Too Far, Need key, No ammo, cooldown, or invalid move.\par
\noindent - Do not apply a cap if failures are mostly coordinate\_miss or otherwise caused by simulator/model execution; lower confidence instead.\par
\smallskip
\noindent Classify a critical blocker only when evidence shows a severe defect or required-content failure that blocks the main requested experience, not merely a weak or confusing interaction. In general, criticalBlocker should be true only for caps of 60 or lower; higher caps usually describe non-critical limitations.\par
\smallskip
\noindent\textbf{Return only JSON with:}\par\nobreak
\begin{Verbatim}[fontsize=\footnotesize,breaklines=true,breakanywhere=true]
{
  "weightedScore": number from 0 to 100,
  "score": number from 0 to 100 after applying usabilityCap,
  "usabilityCap": number from 0 to 100 or null,
  "criticalBlocker": boolean,
  "criticalBlockerReason": string or null,
  "confidence": number from 0 to 100,
  "visualStabilityReward": number from 0 to 100,
  "screenshotAestheticReward": number from 0 to 100,
  "feedbackClarityReward": number from 0 to 100,
  "interactionReliabilityReward": number from 0 to 100,
  "stateLegibilityReward": number from 0 to 100,
  "promptAlignmentReward": number from 0 to 100,
  "taskProgressReward": number from 0 to 100,
  "summary": string,
  "strengths": string[],
  "weaknesses": string[],
  "recurringIssues": [{"issue": string, "severity": "low" | "medium" | "high", "turns": number[], "evidence": string}],
  "failureModeSummary": [{"failureMode": "coordinate_miss" | "action_ineffective" | "illegal_or_rejected" | "timing_issue" | "uncertain" | "none", "count": number, "impact": string}],
  "experience": string,
  "recommendations": string[]
}
\end{Verbatim}
\medskip\noindent\textbf{User:}\par
\begin{Verbatim}[fontsize=\footnotesize,breaklines=true,breakanywhere=true]
Task prompt:
<task_prompt>

Artifact prompt (<artifact_prompt_path><truncated_marker>):
<artifact_prompt_text>

Structured simulation evidence:
{
  "run": {
    "url": <url>,
    "turnCount": <turn_count>,
    "turnLimit": <turn_limit>,
    "terminated": <terminated>,
    "terminationReason": <termination_reason>,
    "totalSimulatedAdvanceMs": <total_simulated_advance_ms>,
    "visualEvaluationSummary": <visual_evaluation_summary_json>
  },
  "turns": [
    {
      "turnIndex": <turn_index>,
      "turnLabel": <turn_label>,
      "action": <action_json>,
      "actionExecution": <action_execution_json>,
      "timing": {
        "totalWaitMs": <total_wait_ms>,
        "simulatedAdvanceMs": <simulated_advance_ms>
      },
      "reachedGoal": <reached_goal>,
      "terminated": <turn_terminated>,
      "terminationReason": <turn_termination_reason>,
      "decisionSummary": <decision_summary>,
      "transition": <compact_transition_evaluation_json>,
      "screenshot": <compact_screenshot_evaluation_json>,
      "visualEvaluationErrors": <visual_evaluation_errors_json>
    },
    <remaining_turns>
  ]
}
\end{Verbatim}
\medskip\noindent\textbf{Compact evaluation fields:}\par
\begin{Verbatim}[fontsize=\footnotesize,breaklines=true,breakanywhere=true]
{
  "judgeType": <judge_type>,
  "turnIndex": <turn_index>,
  "score": <score>,
  "overallReward": <overall_reward>,
  "visualStabilityReward": <visual_stability_reward>,
  "executionReward": <execution_reward>,
  "appFeedbackReward": <app_feedback_reward>,
  "taskProgressReward": <task_progress_reward>,
  "readabilityReward": <readability_reward>,
  "stateVisibilityReward": <state_visibility_reward>,
  "screenshotAestheticReward": <screenshot_aesthetic_reward>,
  "summary": <summary>,
  "strengths": <strengths>,
  "weaknesses": <weaknesses>,
  "experience": <experience>,
  "state": <state>,
  "failureMode": <failure_mode>
}
\end{Verbatim}
\end{appendixprompt}

\subsubsection{Reference-Based Visual Comparison}
\label{app:prompt-anchor}
The Visual Anchor receives the reference screenshot followed by the current rendering.

\begin{appendixprompt}{Reference-Based Visual Comparison}
\noindent\textbf{System:}\par\smallskip
\noindent You compare two screenshots of HTML games created for the exact same task prompt. Image 1 is the retained pre-rollout reference. Image 2 is the current training rollout being evaluated.\par
\smallskip
\noindent Judge only visual presentation: scene quality, beauty, composition, visual hierarchy, layout organization, spacing, typography, color harmony, coherence, polish, and readability. Do not penalize ordinary differences in gameplay progress, animation timing, selected controls, scores, or transient state unless they materially harm visual organization.\par
\smallskip
\noindent If the screenshots do not show an obvious visual-quality difference, assign degradationLevel 0. Treat the provided screenshot scores only as a calibration signal: when their absolute delta is at most 8 points, default to no penalty unless Image 2 shows a clear, material visual degradation. Do not penalize small, ambiguous, or merely numerical differences.\par
\smallskip
\noindent\textbf{Return degradationLevel from 0 to 100 measuring how much worse Image 2 is than Image 1:}\par\nobreak
\noindent - 0: Image 2 is visually equal to or better than Image 1.\par
\noindent - 1-20: Image 2 is only slightly worse.\par
\noindent - 21-50: Image 2 is noticeably worse in multiple visual dimensions.\par
\noindent - 51-80: Image 2 is substantially less polished, coherent, or organized.\par
\noindent - 81-100: Image 2 is dramatically worse or visually unusable relative to Image 1.\par
\smallskip
\noindent\textbf{Do not assign a positive penalty merely because the images differ. Return only JSON with:}\par\nobreak
\begin{Verbatim}[fontsize=\footnotesize,breaklines=true,breakanywhere=true]
{
  "degradationLevel": number from 0 to 100,
  "summary": string,
  "referenceAdvantages": string[],
  "currentAdvantages": string[],
  "currentDeficiencies": string[]
}
\end{Verbatim}
\medskip\noindent\textbf{User:}\par
\begin{Verbatim}[fontsize=\footnotesize,breaklines=true,breakanywhere=true]
Task prompt:
<task_prompt>

Reference selection metadata:
{
  "promptId": <reference_prompt_id>,
  "turnIndex": <reference_turn_index>,
  "screenshotScore": <reference_screenshot_score>,
  "selection": <reference_selection>
}

Current selection metadata:
{
  "turnIndex": <current_turn_index>,
  "screenshotScore": <current_screenshot_score>
}

Image 1: pre-rollout reference
Path: <reference_screenshot_path>
<reference_screenshot_attachment>

Image 2: current evaluated HTML
Path: <current_screenshot_path>
<current_screenshot_attachment>
\end{Verbatim}
\end{appendixprompt}

\clearpage
\providecommand{\caseGameHeader}{%
  \begin{minipage}{0.49\textwidth}\centering\small\textbf{Earlier state}\end{minipage}\hfill
  \begin{minipage}{0.49\textwidth}\centering\small\textbf{After interaction}\end{minipage}\par\medskip}

\begin{figure*}[p]
  \centering
  \begin{minipage}{\textwidth}
\subsection{Case Study}
\label{app:case-study}

\subsubsection{Game Generation Tasks}
\label{app:game-cases}

  \end{minipage}\par\medskip
  \caseGameHeader
  \noindent\textbf{Qwen3.6-35B-A3B}\hfill{\small\color{black!70}Rendering breaks after launch}\par\smallskip
\includegraphics[width=0.49\textwidth]{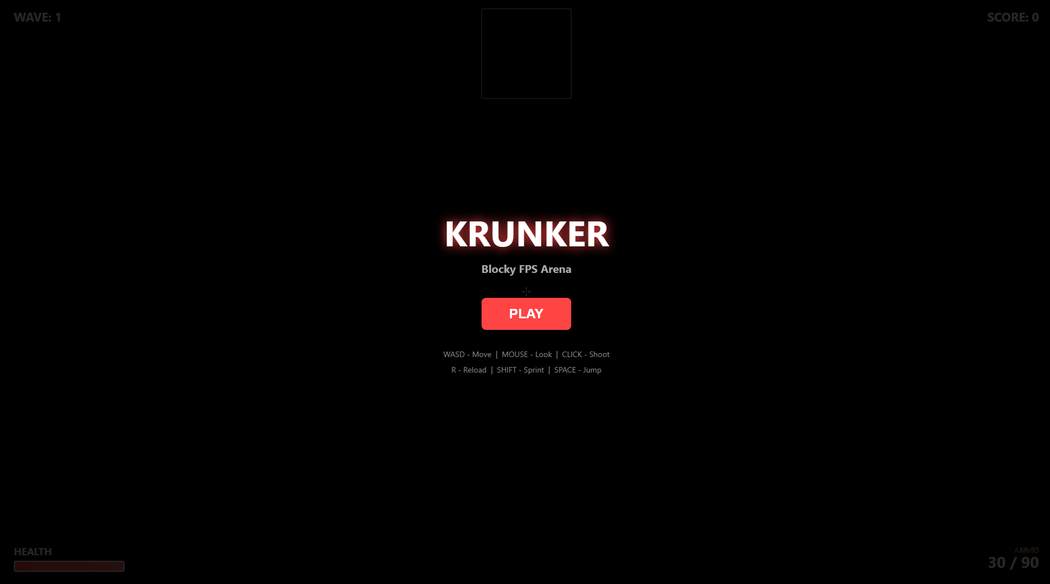}\hfill
\includegraphics[width=0.49\textwidth]{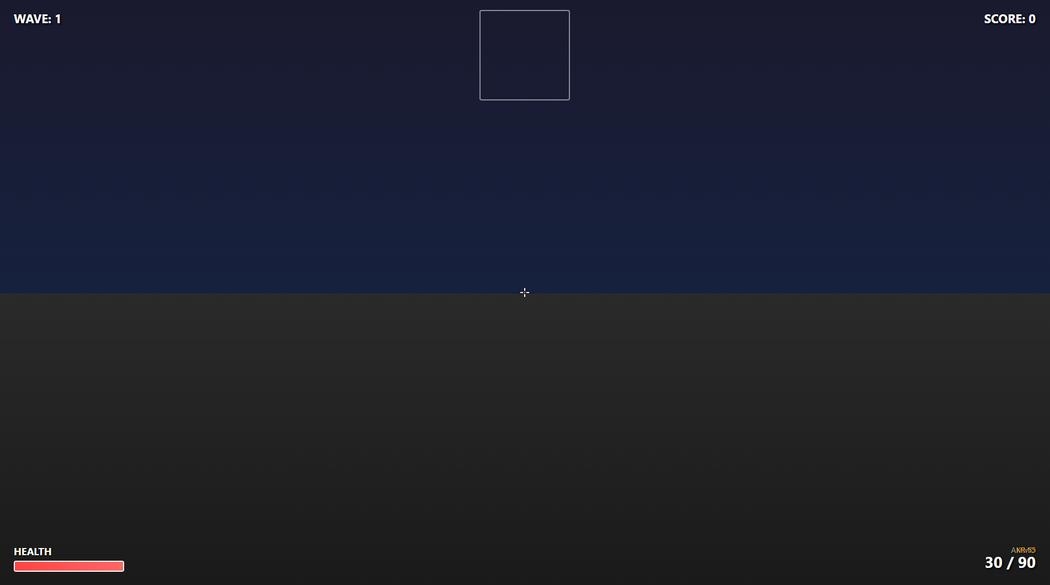}\par\medskip
  \noindent\textbf{Gemini-3.1-Pro}\hfill{\small\color{black!70}}\par\smallskip
\includegraphics[width=0.49\textwidth]{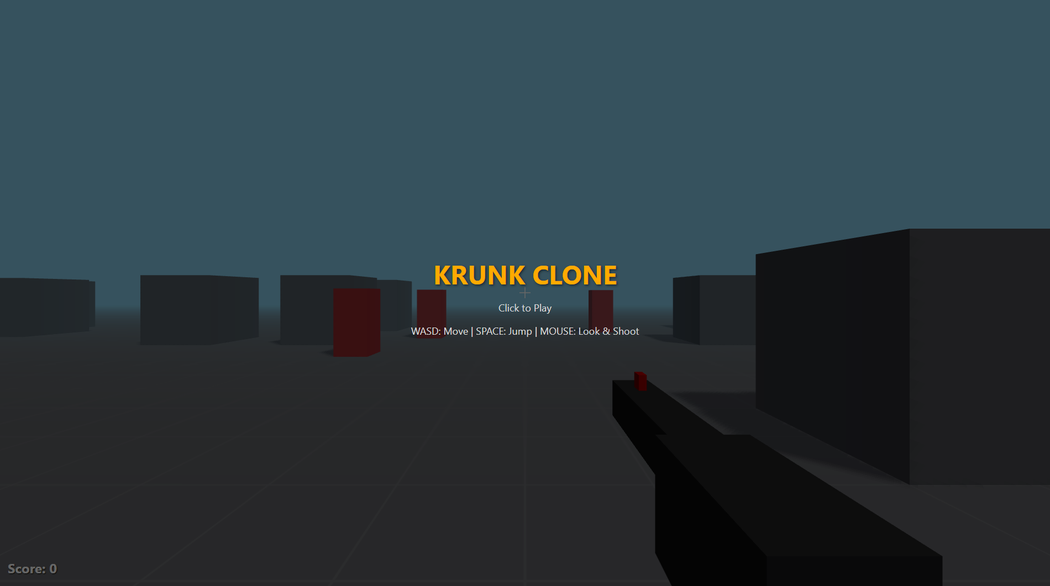}\hfill
\includegraphics[width=0.49\textwidth]{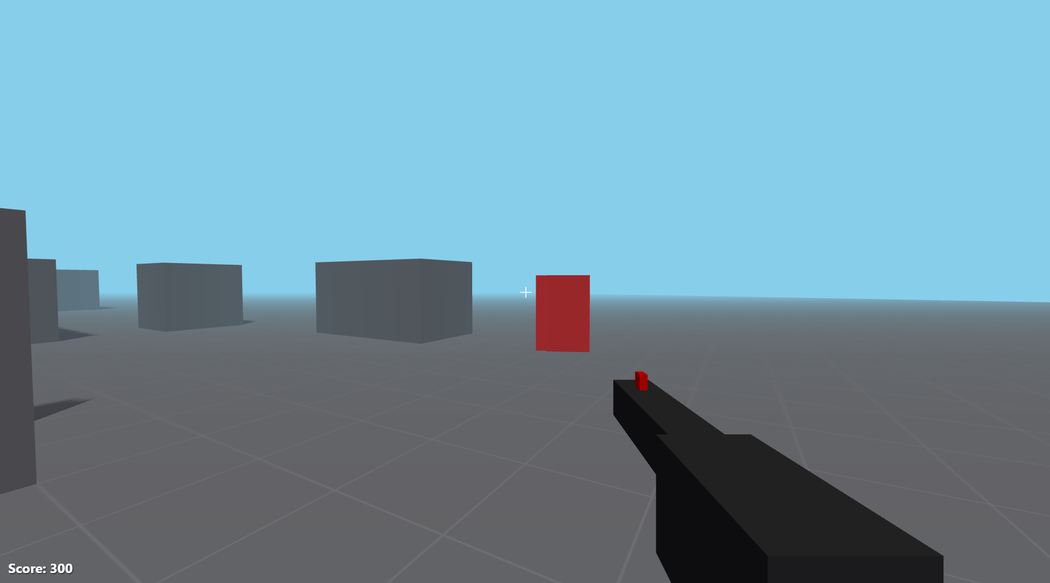}\par\medskip
  \noindent\textbf{GPT-5.3-Codex}\hfill{\small\color{black!70}No enemies in the play view}\par\smallskip
\includegraphics[width=0.49\textwidth]{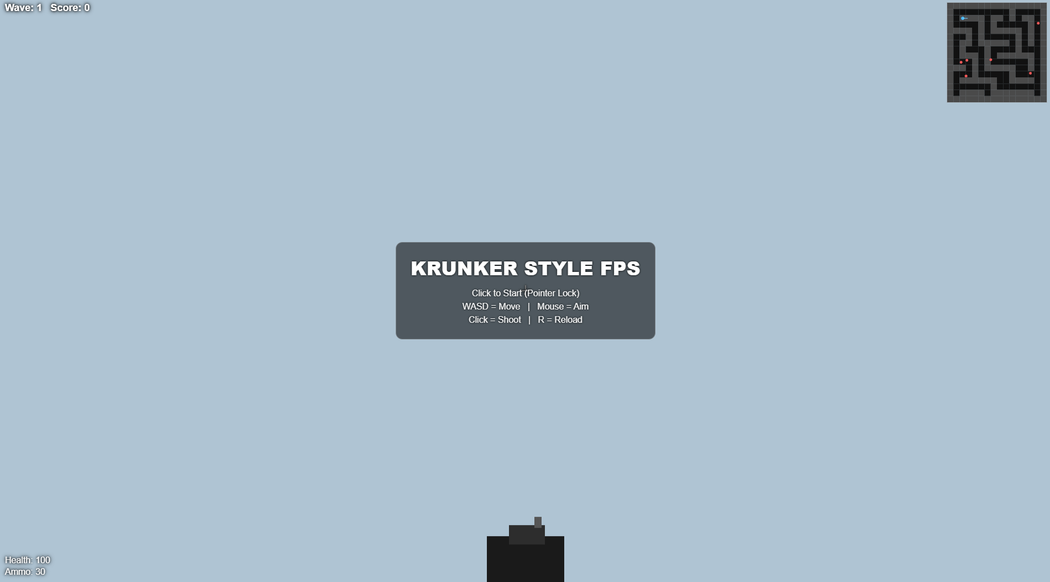}\hfill
\includegraphics[width=0.49\textwidth]{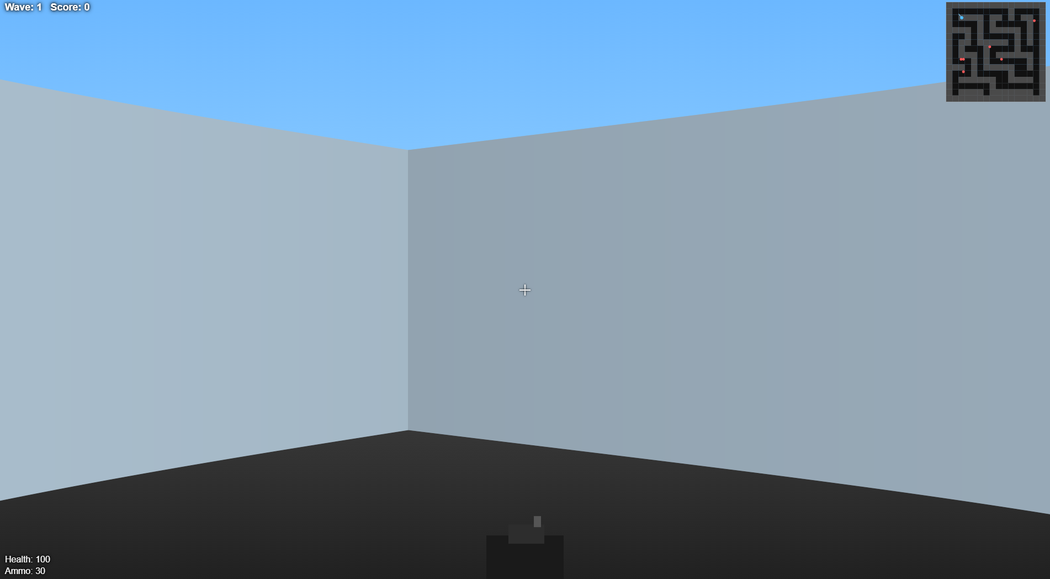}\par\medskip
  \caption{Compared with the Qwen3.6-35B-A3B backbone, WebCatalyst shows markedly stronger robustness after interaction begins. While the backbone often produces plausible initial interfaces, its artifacts may break during execution—for example, rendering fails after launching Krunker, the 3D collection game turns black, Minecraft terrain disappears, and movement is blocked in the tank game. In contrast, WebCatalyst preserves coherent interactive states, with complete FPS scenes, visible collection effects and score updates, stable voxel terrain, and responsive projectile effects. These examples suggest that the gains extend beyond initial appearance to maintaining functionality and visual coherence throughout interaction. \textbf{Krunker comparison (1/2).} The request is ``Krunker Game''. The paired views show the launch interface and the first-person scene reached during interaction.}
  \label{fig:case-krunker-a}
\end{figure*}

\begin{figure*}[p]
  \centering
  \caseGameHeader
  \noindent\textbf{Claude-Opus-4.7}\hfill{\small\color{black!70}}\par\smallskip
\includegraphics[width=0.49\textwidth]{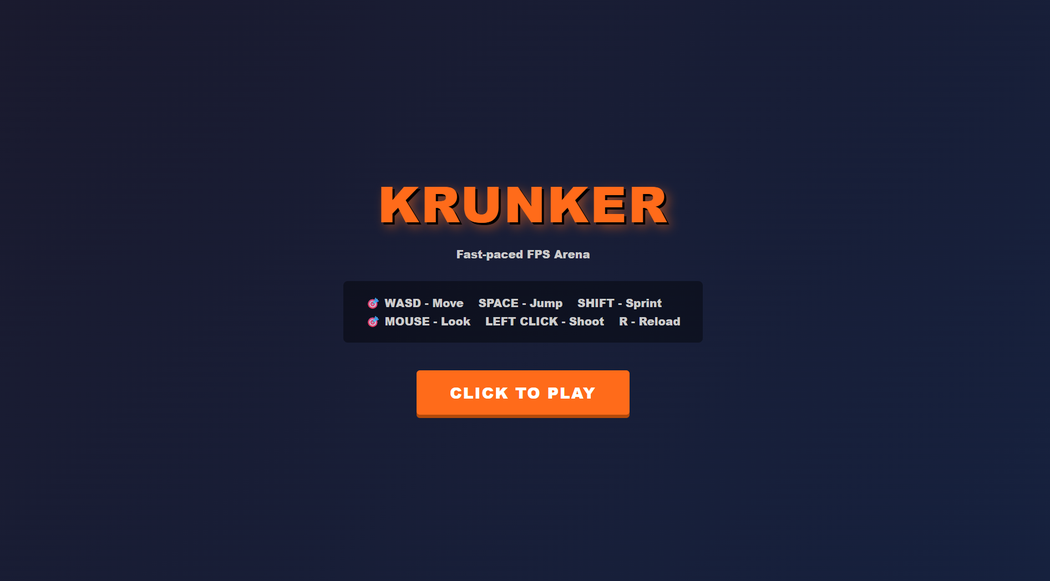}\hfill
\includegraphics[width=0.49\textwidth]{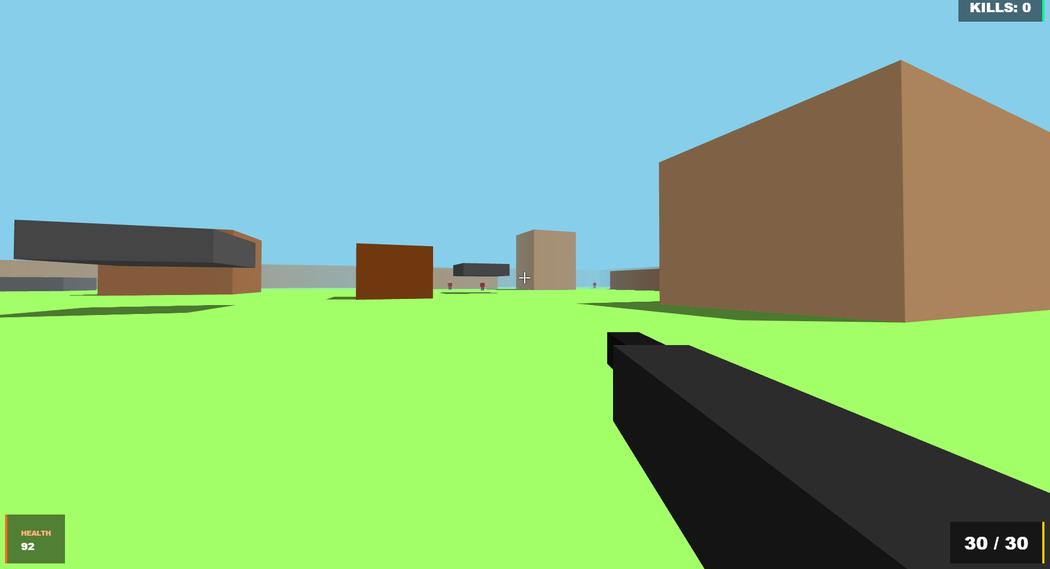}\par\medskip
  \noindent\textbf{GPT-5.4-mini}\hfill{\small\color{black!70}Rendering breaks after launch}\par\smallskip
\includegraphics[width=0.49\textwidth]{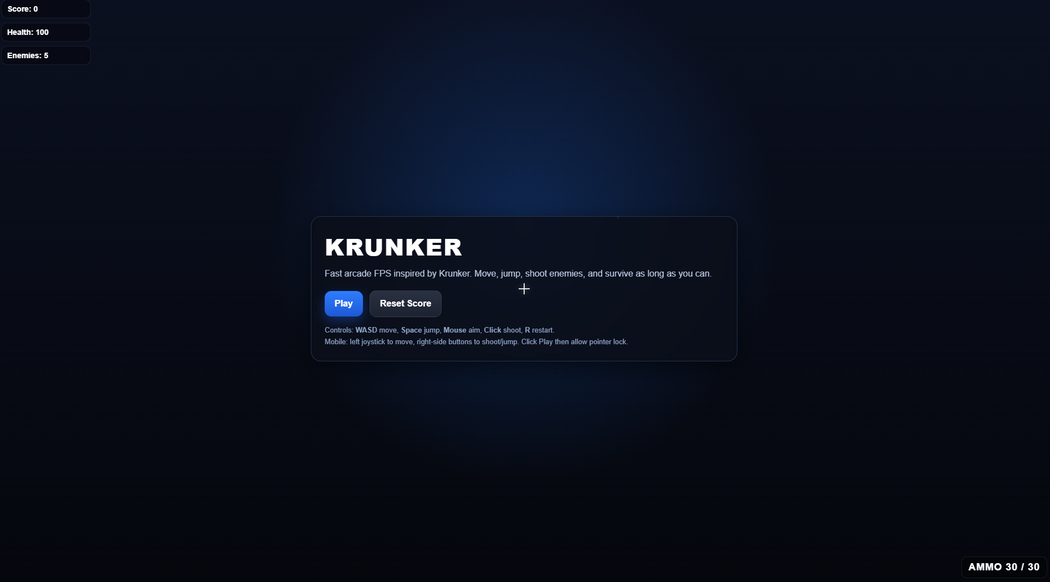}\hfill
\includegraphics[width=0.49\textwidth]{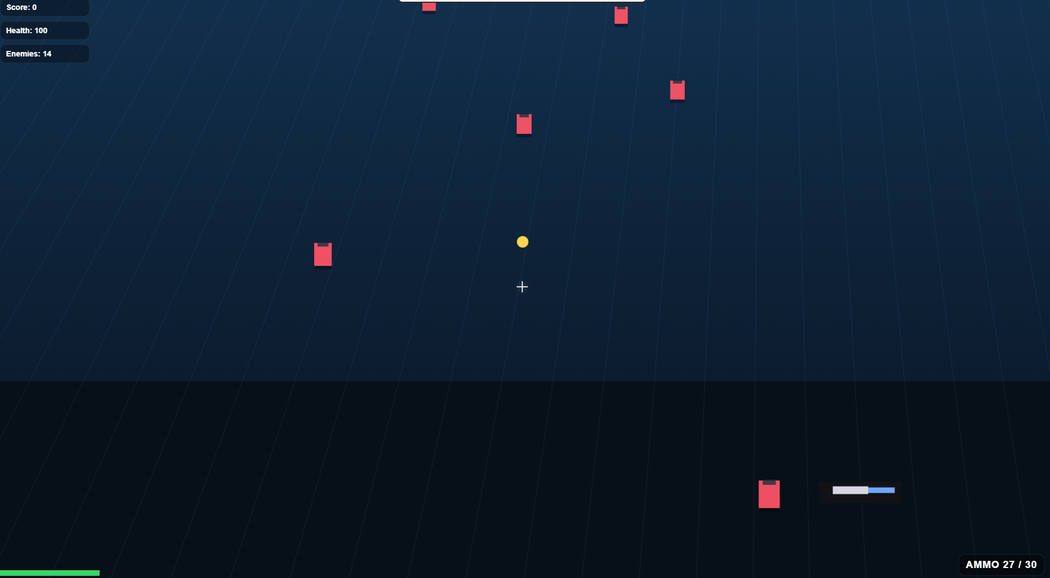}\par\medskip
  \noindent\textbf{WebCatalyst}\hfill{\small\color{black!70}Buildings, enemies, and weapon HUD}\par\smallskip
\includegraphics[width=0.49\textwidth]{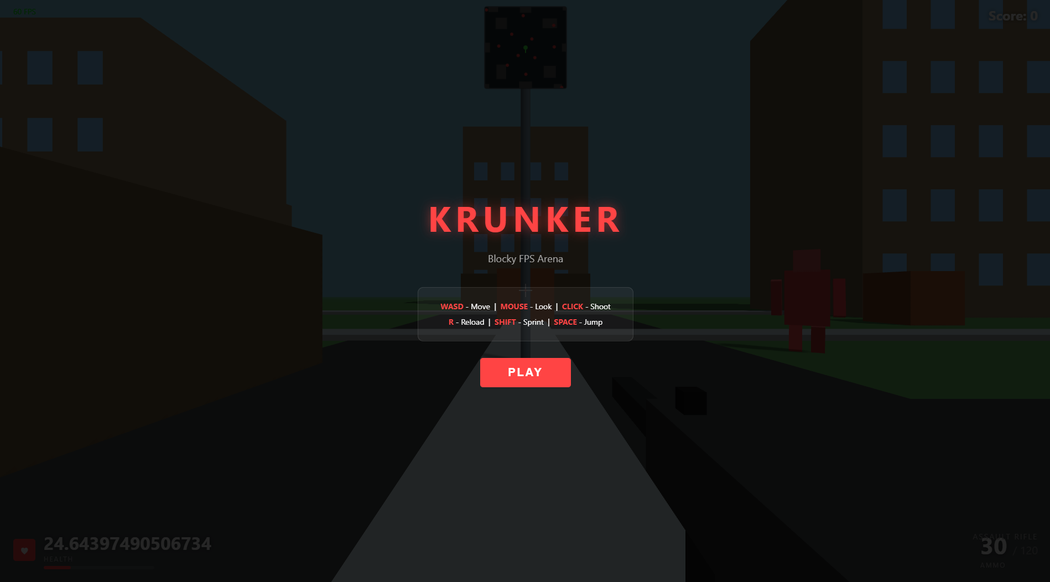}\hfill
\includegraphics[width=0.49\textwidth]{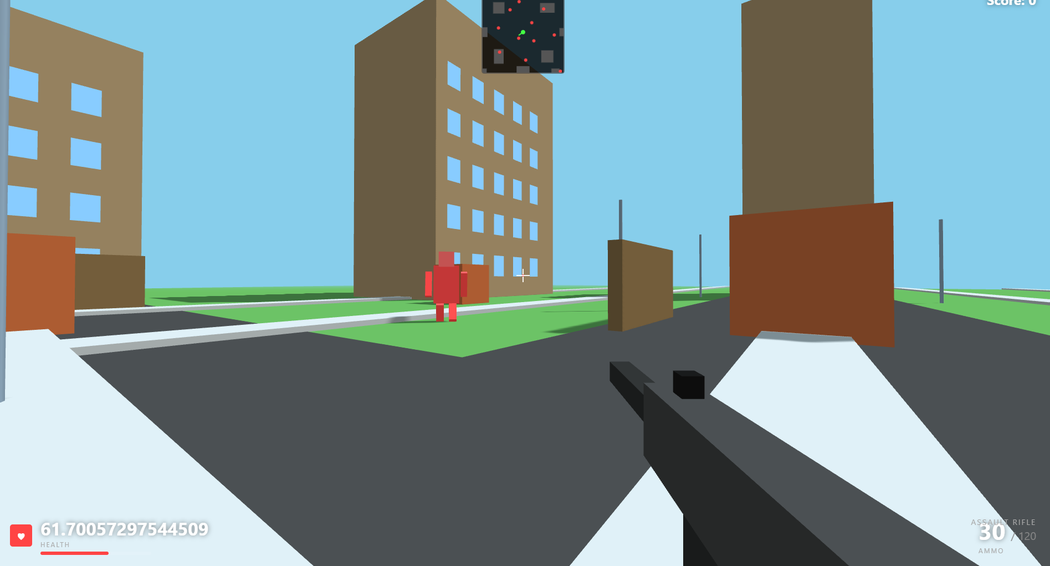}\par\medskip
  \caption{\textbf{Krunker comparison (2/2).} Continuation of Figure~\ref{fig:case-krunker-a}.}
  \label{fig:case-krunker-b}
\end{figure*}

\begin{figure*}[p]
  \centering
  \caseGameHeader
  \noindent\textbf{Qwen3.6-35B-A3B}\hfill{\small\color{black!70}Black play view after launch}\par\smallskip
\includegraphics[width=0.49\textwidth]{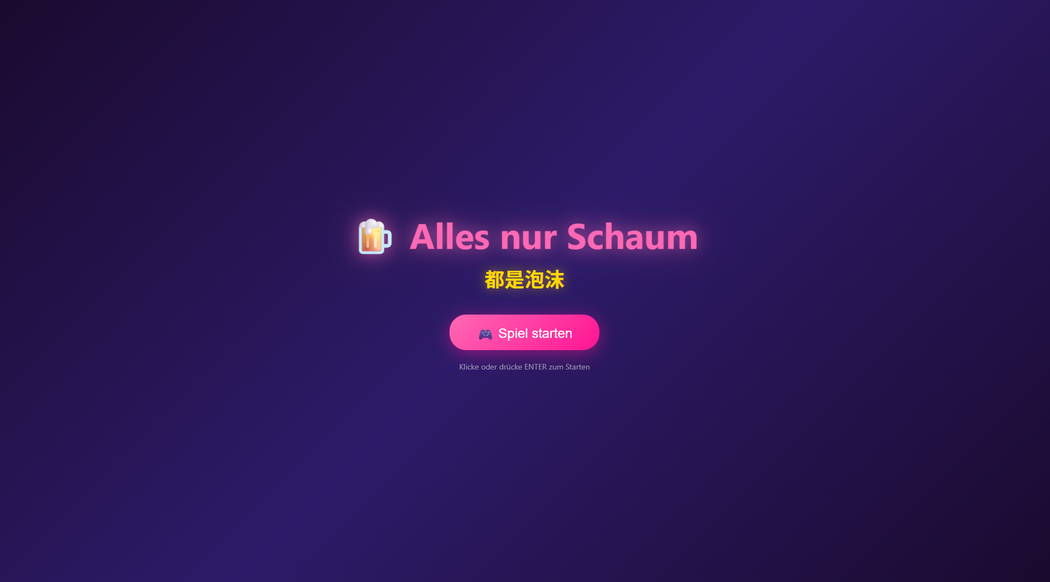}\hfill
\includegraphics[width=0.49\textwidth]{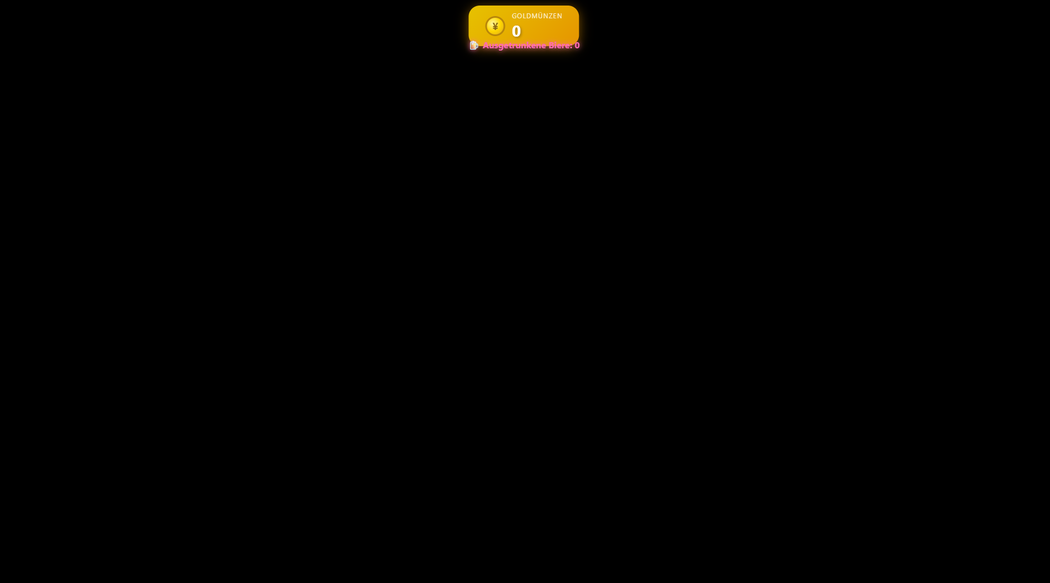}\par\medskip
  \noindent\textbf{GPT-5.3-Codex}\hfill{\small\color{black!70}Backward movement is blocked}\par\smallskip
\includegraphics[width=0.49\textwidth]{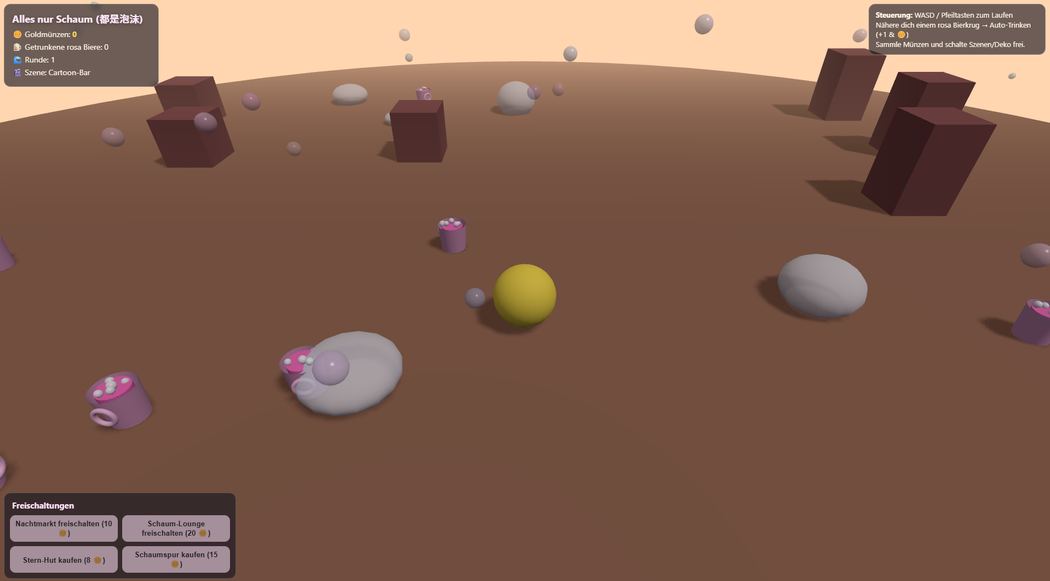}\hfill
\includegraphics[width=0.49\textwidth]{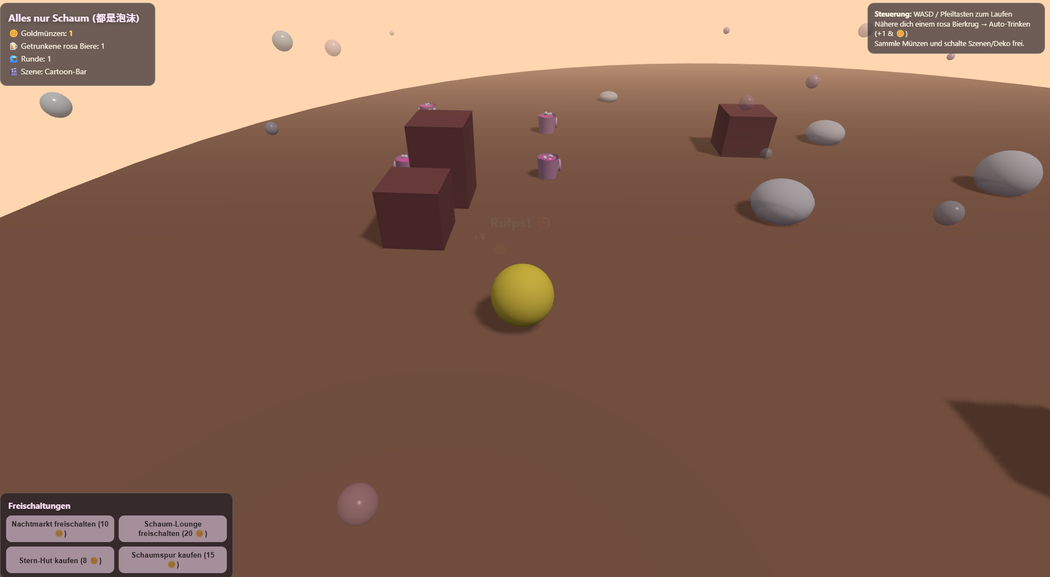}\par\medskip
  \noindent\textbf{Claude-Opus-4.7}\hfill{\small\color{black!70}}\par\smallskip
\includegraphics[width=0.49\textwidth]{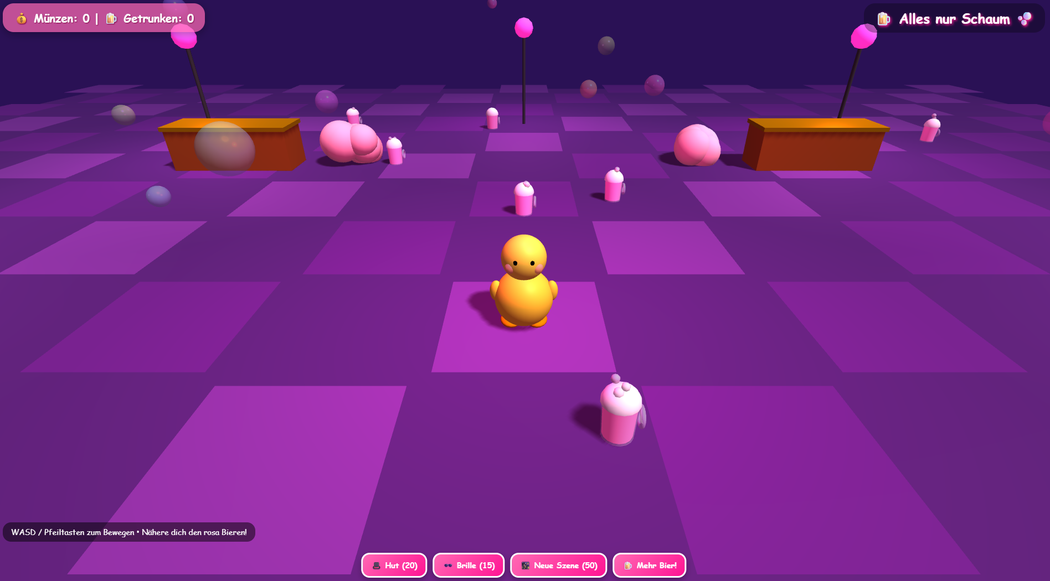}\hfill
\includegraphics[width=0.49\textwidth]{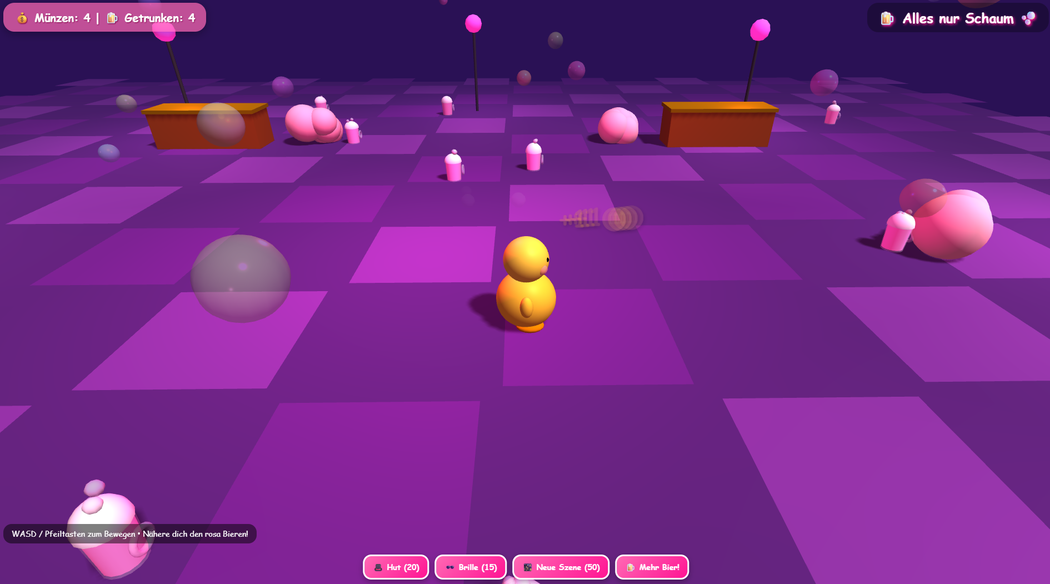}\par\medskip
  \caption{\textbf{3D collection-game comparison (1/2).} The German request for ``Alles nur Schaum'' asks for a small yellow character, pink beer mugs, automatic collection on approach, a \texttt{+1} score effect, and coin rewards in a cheerful 3D environment.}
  \label{fig:case-foam-a}
\end{figure*}

\begin{figure*}[p]
  \centering
  \caseGameHeader
  \noindent\textbf{GPT-5.4-mini}\hfill{\small\color{black!70}Items do not respond to interaction}\par\smallskip
\includegraphics[width=0.49\textwidth]{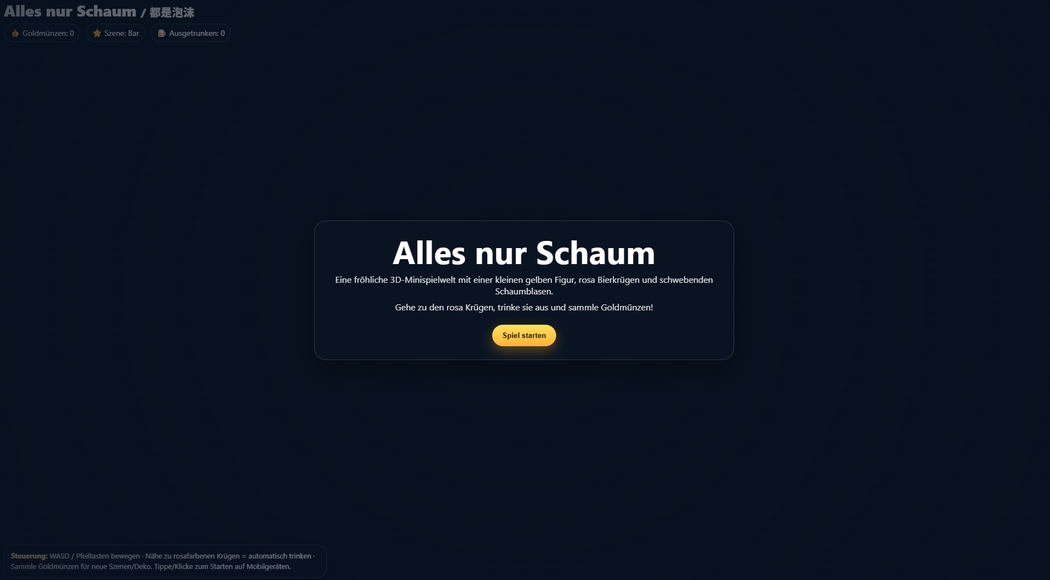}\hfill
\includegraphics[width=0.49\textwidth]{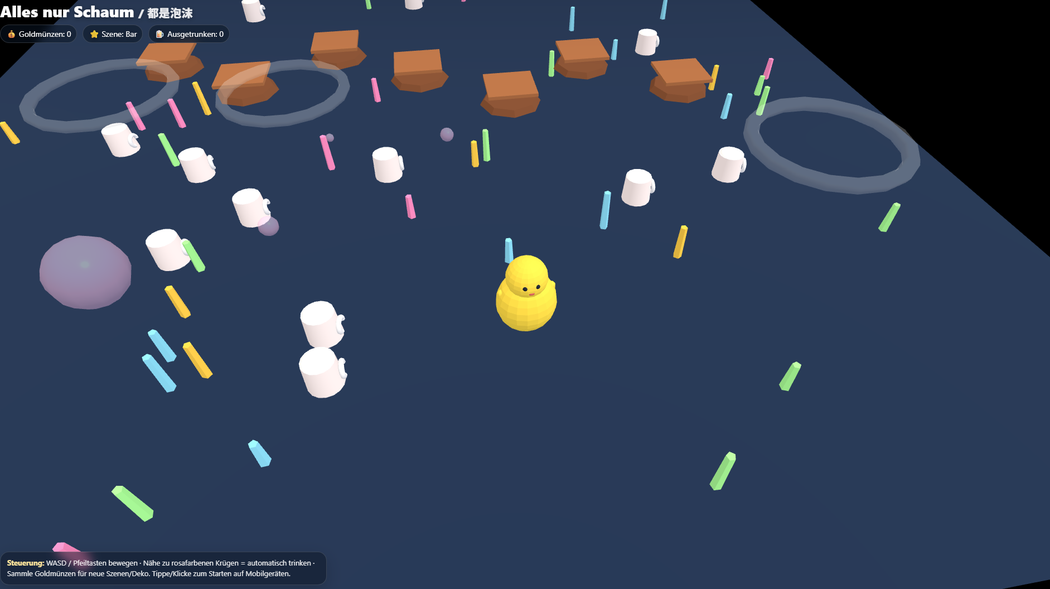}\par\medskip
  \noindent\textbf{WebCatalyst}\hfill{\small\color{black!70}Collection effect and updated score}\par\smallskip
\includegraphics[width=0.49\textwidth]{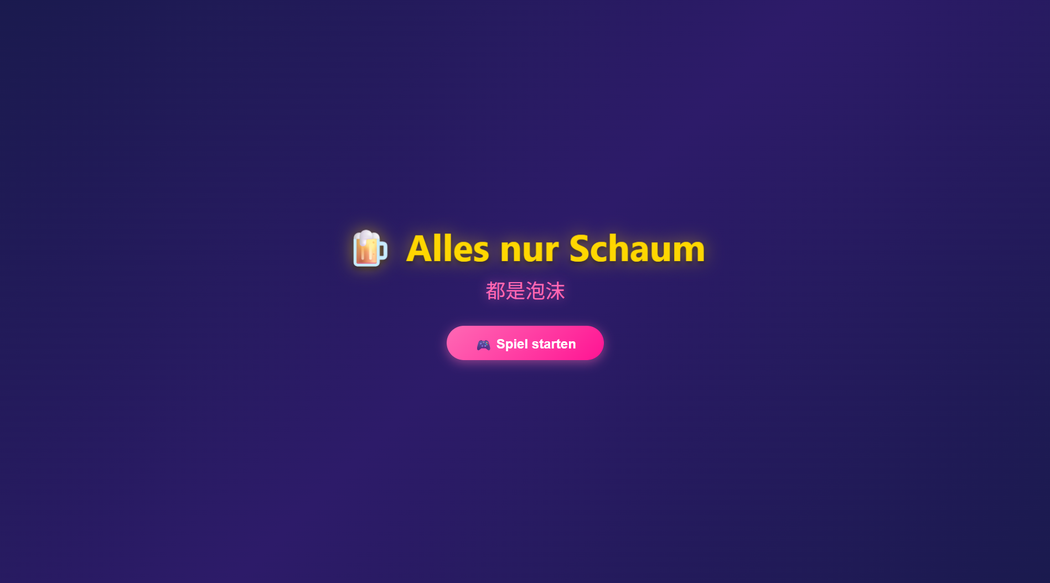}\hfill
\includegraphics[width=0.49\textwidth]{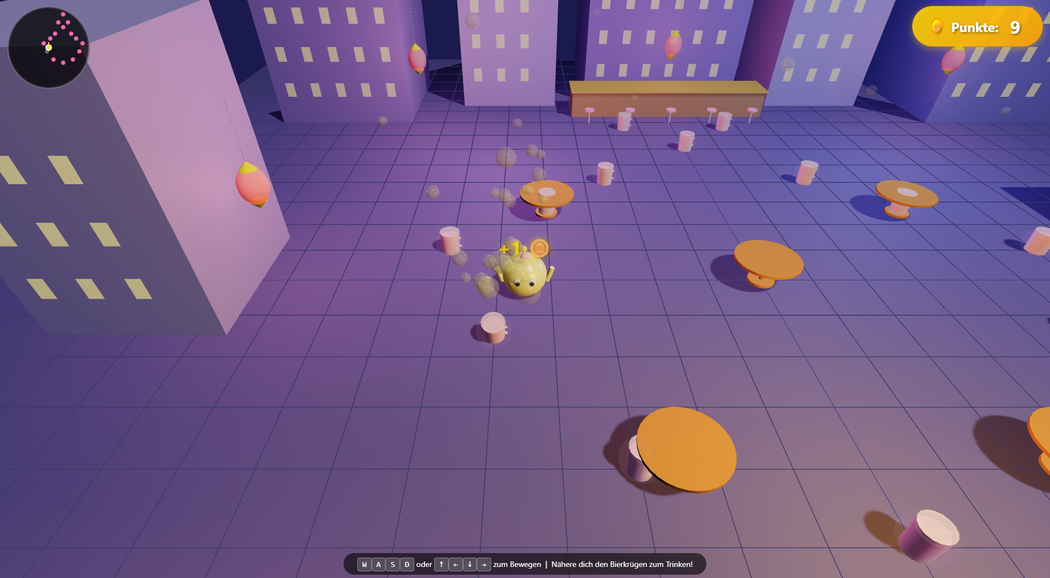}\par\medskip
  \caption{\textbf{3D collection-game comparison (2/2).} Continuation of Figure~\ref{fig:case-foam-a}. WebCatalyst shows a high-quality 3D scene with a \texttt{+1} collection effect and an updated score.}
  \label{fig:case-foam-b}
\end{figure*}

\begin{figure*}[p]
  \centering
  \caseGameHeader
  \noindent\textbf{Qwen3.6-35B-A3B}\hfill{\small\color{black!70}Terrain disappears after launch}\par\smallskip
\includegraphics[width=0.49\textwidth]{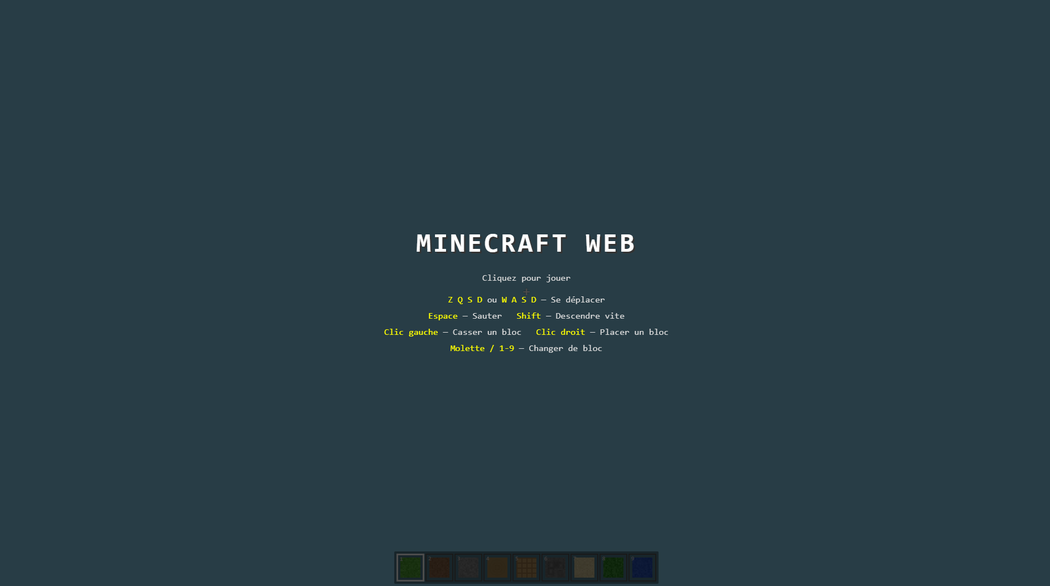}\hfill
\includegraphics[width=0.49\textwidth]{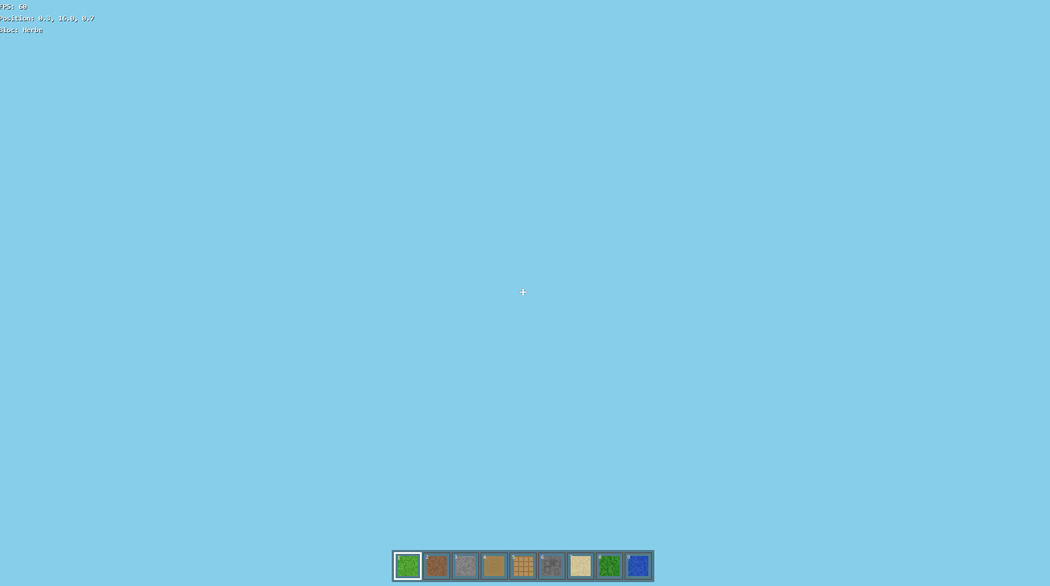}\par\medskip
  \noindent\textbf{Gemini-3.1-Pro}\hfill{\small\color{black!70}}\par\smallskip
\includegraphics[width=0.49\textwidth]{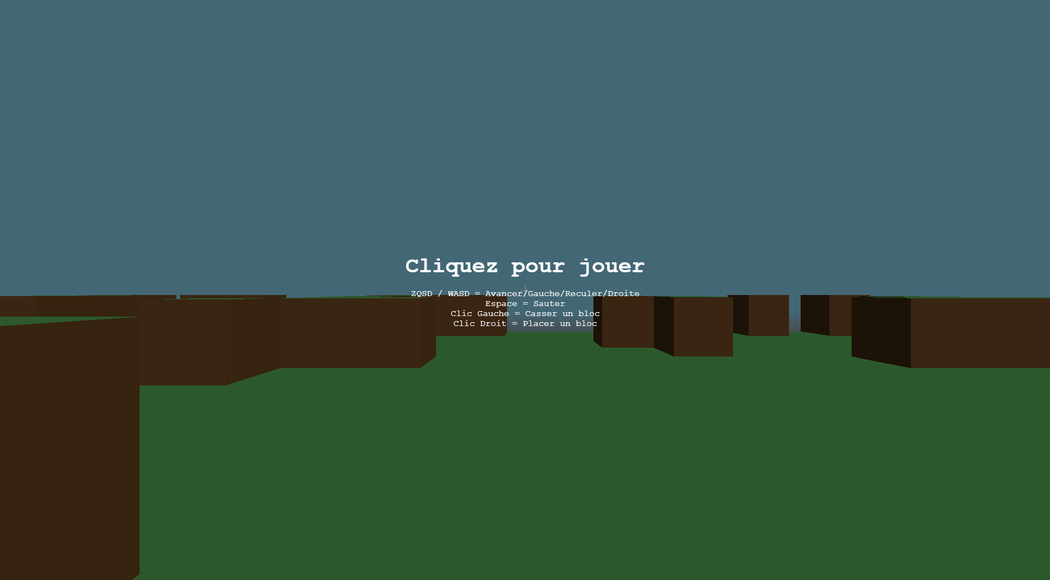}\hfill
\includegraphics[width=0.49\textwidth]{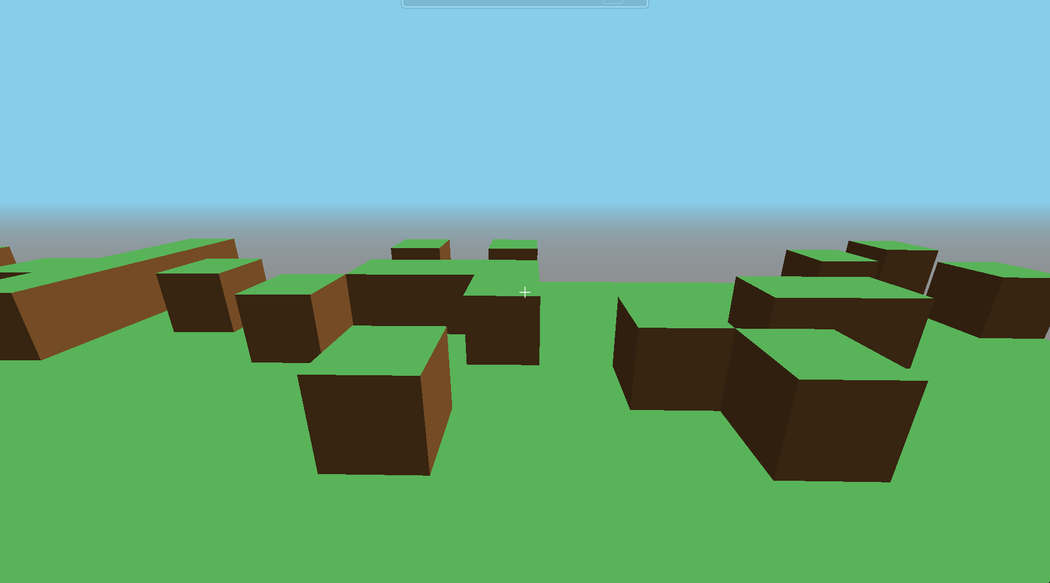}\par\medskip
  \noindent\textbf{GPT-5.3-Codex}\hfill{\small\color{black!70}Remains on the launch interface}\par\smallskip
\includegraphics[width=0.49\textwidth]{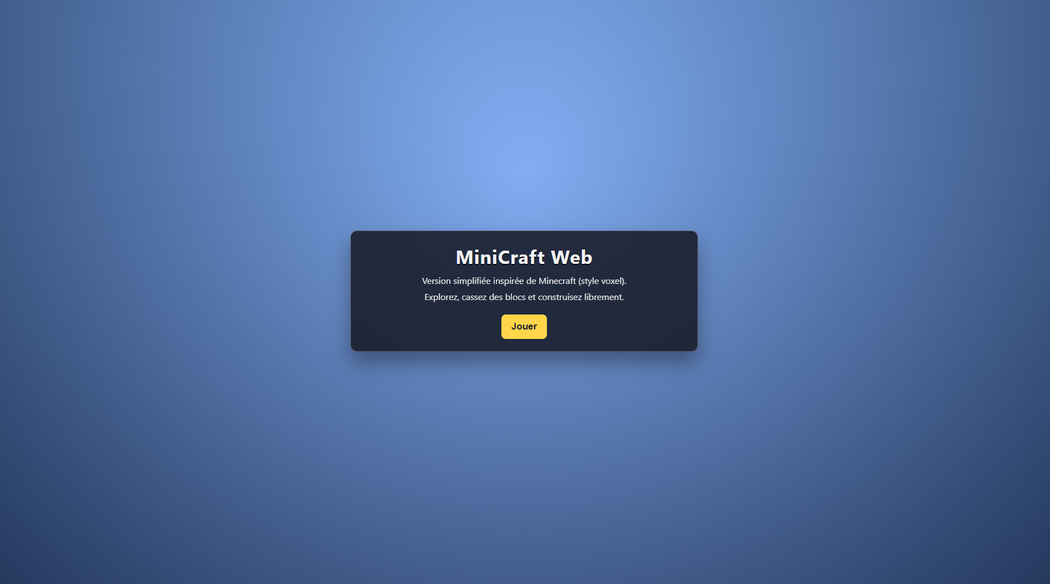}\hfill
\includegraphics[width=0.49\textwidth]{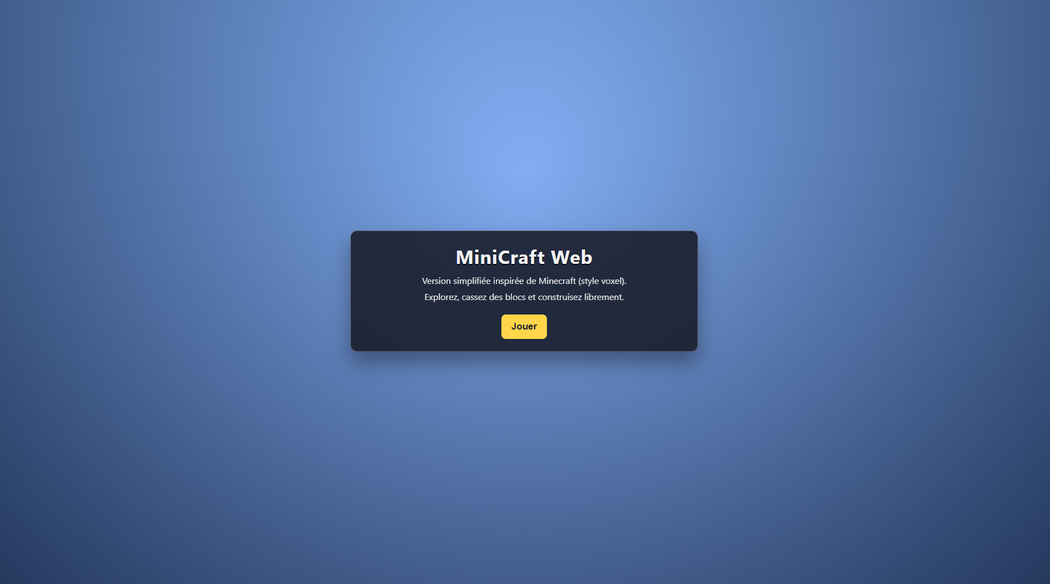}\par\medskip
  \caption{\textbf{Minecraft comparison (1/2).} The French request asks for a web version of Minecraft that preserves its visual style and core functionality. The views show the interface and the resulting voxel world after attempting to enter the game.}
  \label{fig:case-minecraft-a}
\end{figure*}

\begin{figure*}[p]
  \centering
  \caseGameHeader
  \noindent\textbf{Claude-Opus-4.7}\hfill{\small\color{black!70}}\par\smallskip
\includegraphics[width=0.49\textwidth]{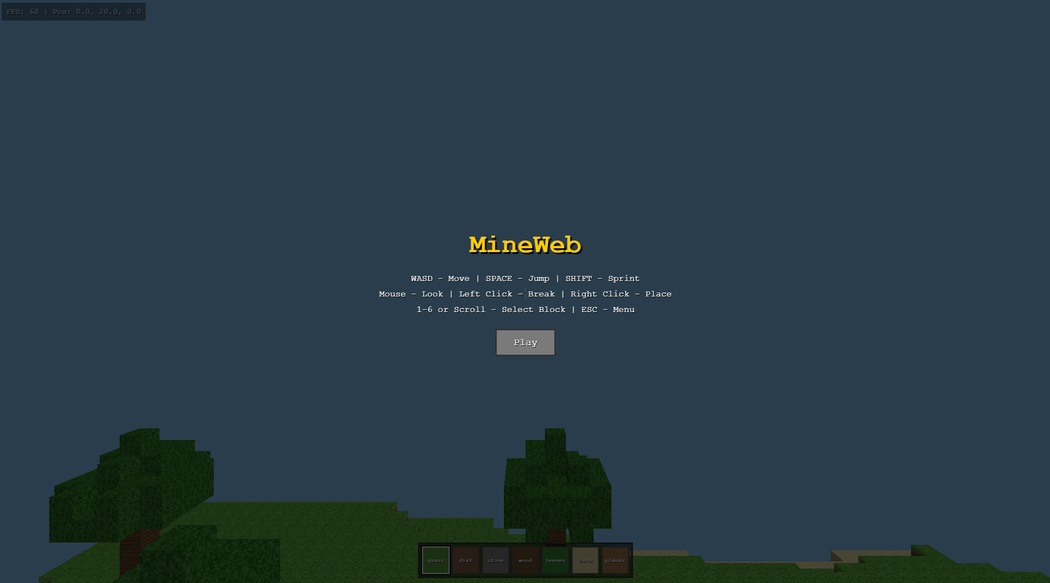}\hfill
\includegraphics[width=0.49\textwidth]{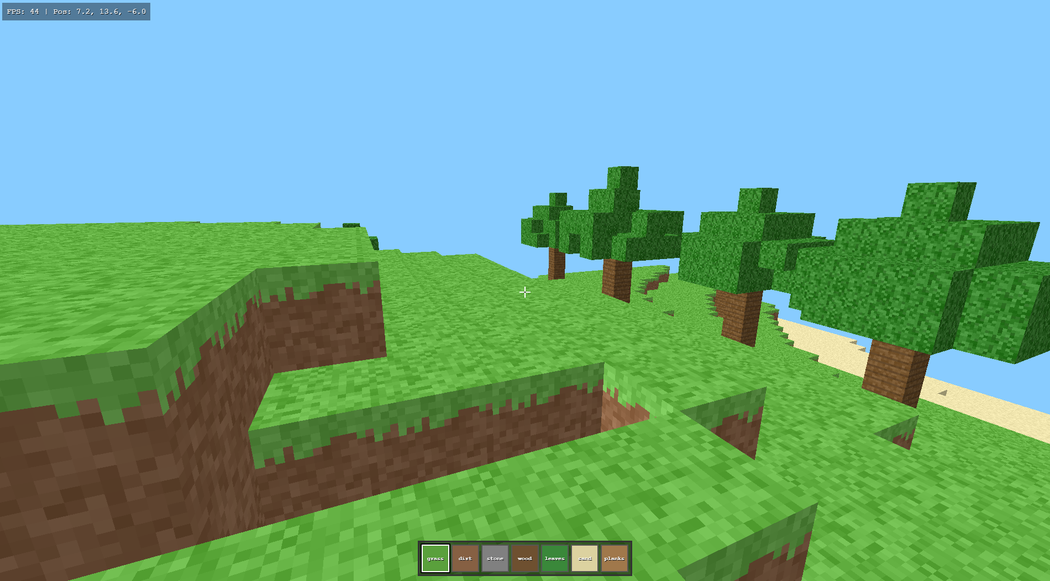}\par\medskip
  \noindent\textbf{GPT-5.4-mini}\hfill{\small\color{black!70}Character keeps falling}\par\smallskip
\includegraphics[width=0.49\textwidth]{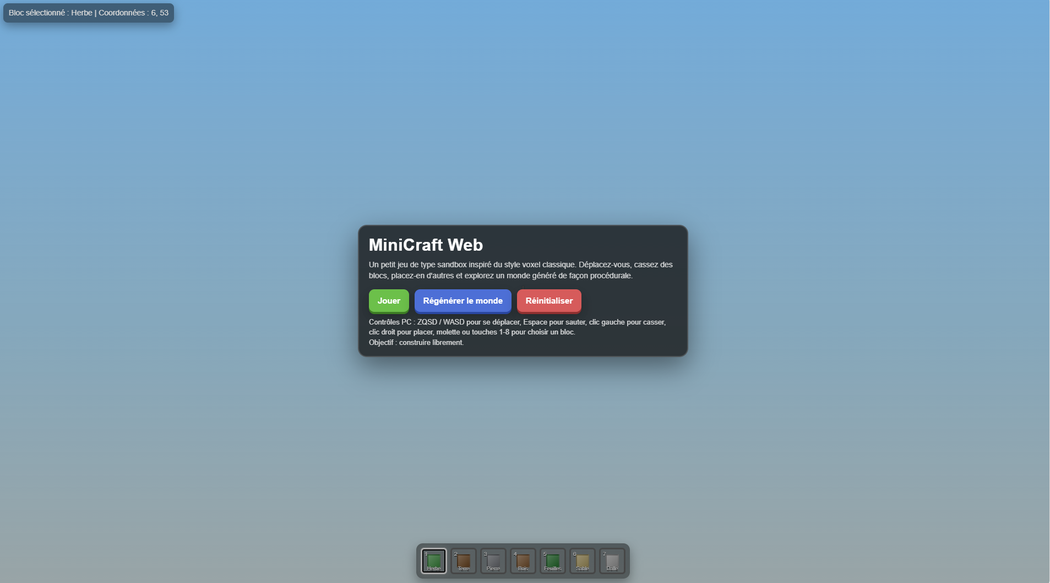}\hfill
\includegraphics[width=0.49\textwidth]{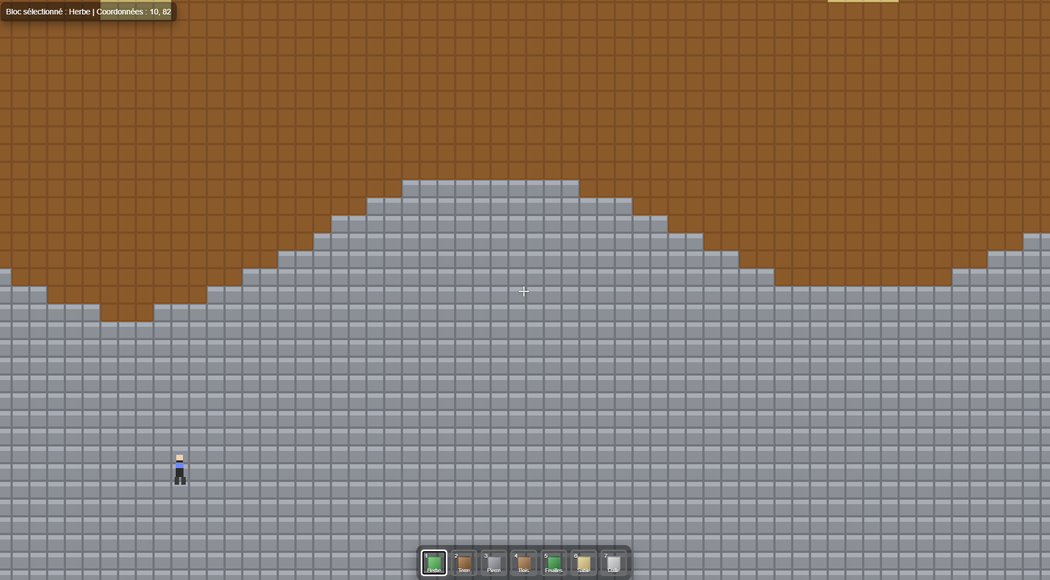}\par\medskip
  \noindent\textbf{WebCatalyst}\hfill{\small\color{black!70}Textured terrain, water, and block inventory}\par\smallskip
\includegraphics[width=0.49\textwidth]{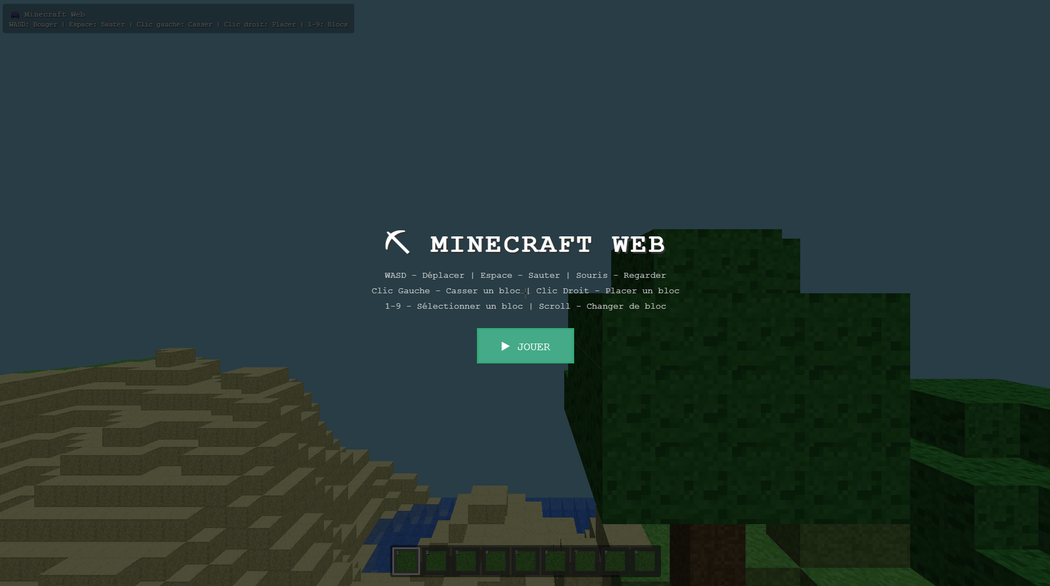}\hfill
\includegraphics[width=0.49\textwidth]{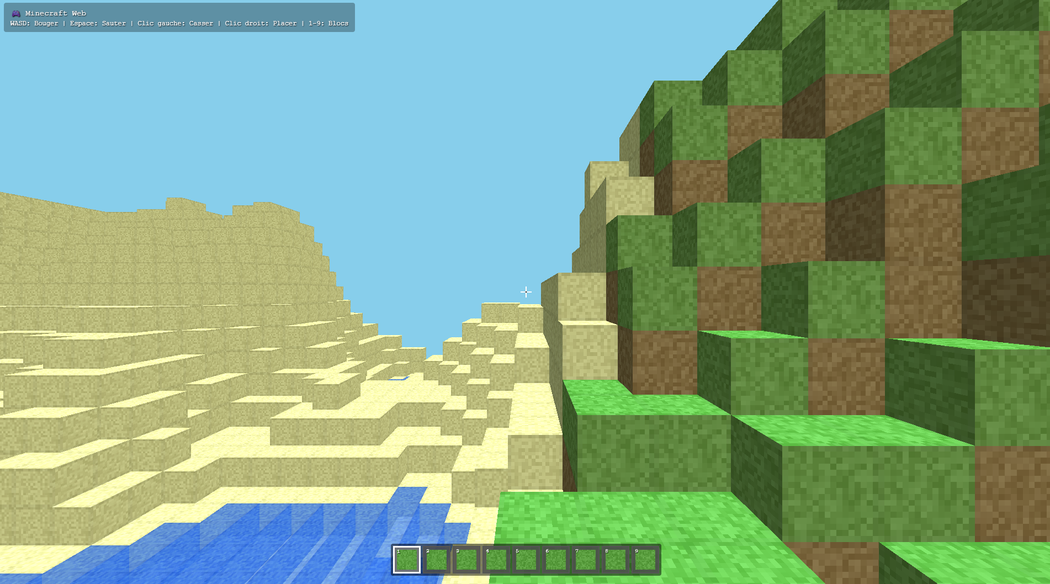}\par\medskip
  \caption{\textbf{Minecraft comparison (2/2).} Continuation of Figure~\ref{fig:case-minecraft-a}.}
  \label{fig:case-minecraft-b}
\end{figure*}

\begin{figure*}[p]
  \centering
  \caseGameHeader
  \noindent\textbf{Qwen3.6-35B-A3B}\hfill{\small\color{black!70}Movement is blocked}\par\smallskip
\includegraphics[width=0.49\textwidth]{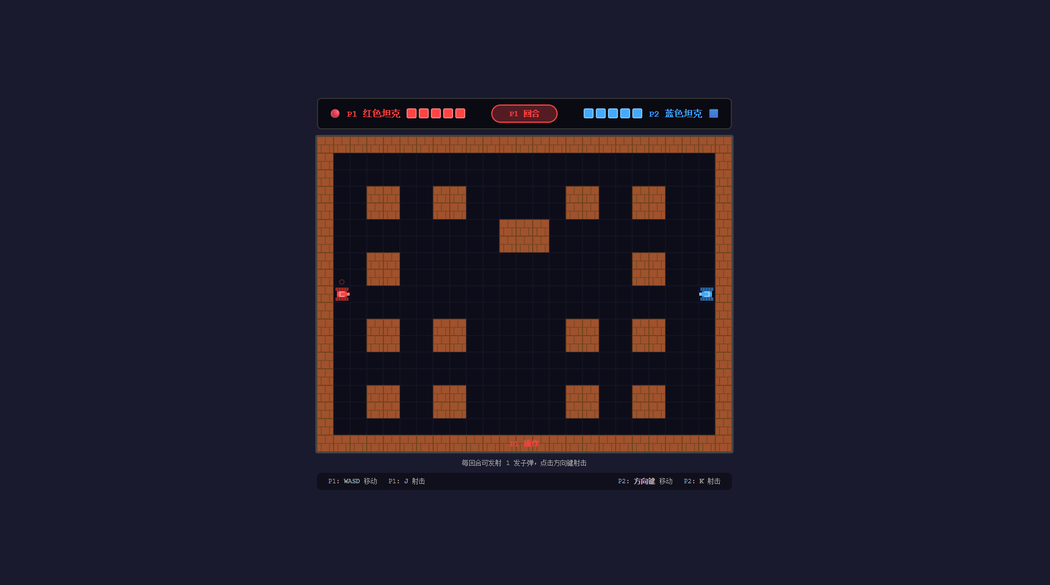}\hfill
\includegraphics[width=0.49\textwidth]{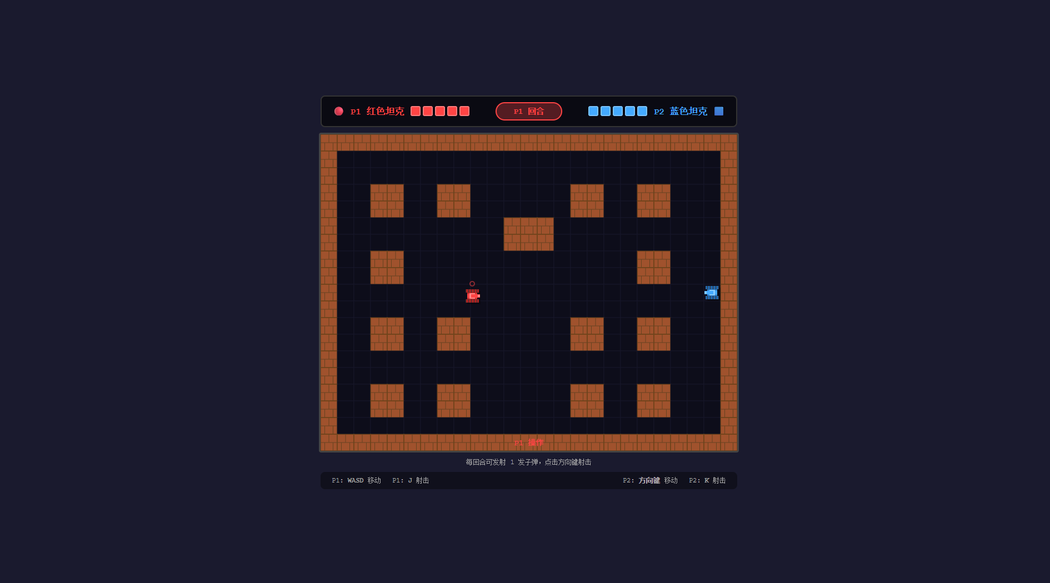}\par\medskip
  \noindent\textbf{Gemini-3.1-Pro}\hfill{\small\color{black!70}}\par\smallskip
\includegraphics[width=0.49\textwidth]{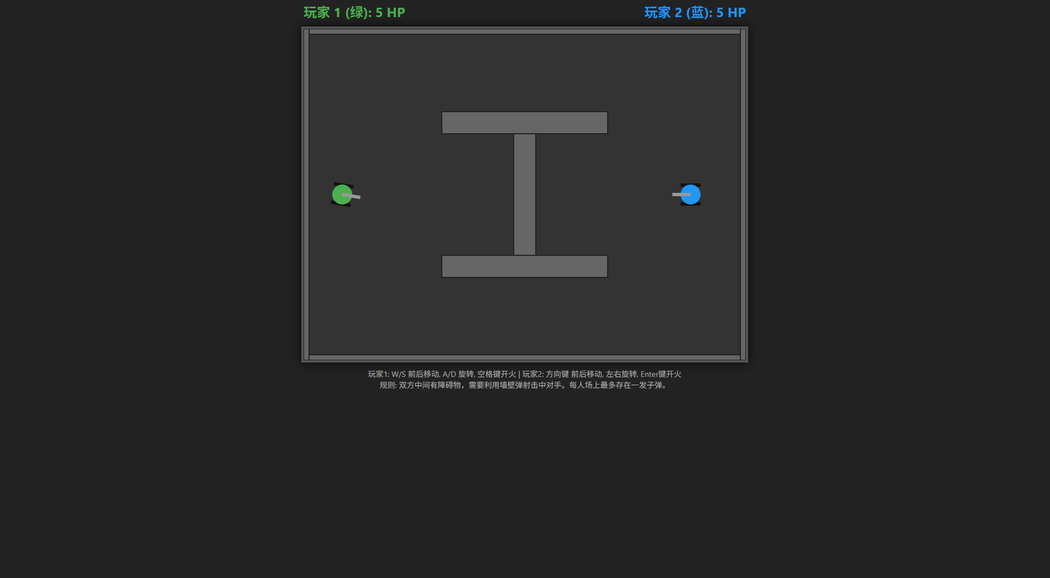}\hfill
\includegraphics[width=0.49\textwidth]{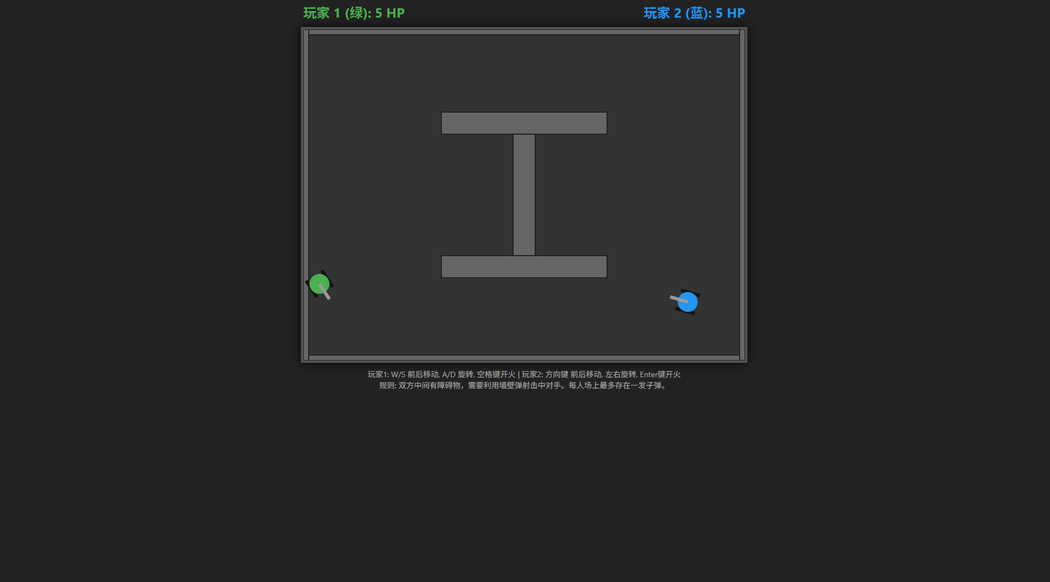}\par\medskip
  \noindent\textbf{GPT-5.3-Codex}\hfill{\small\color{black!70}}\par\smallskip
\includegraphics[width=0.49\textwidth]{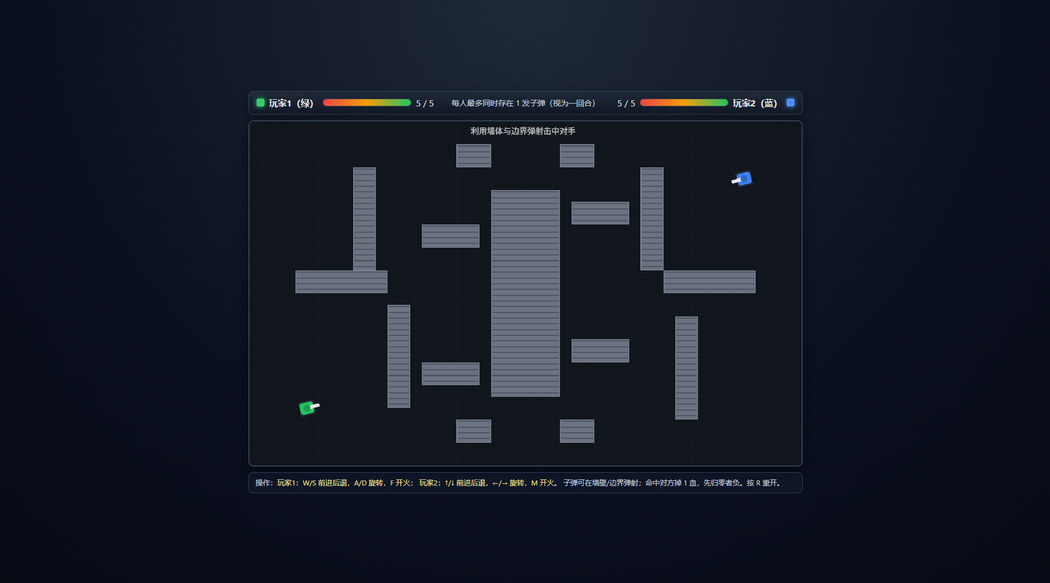}\hfill
\includegraphics[width=0.49\textwidth]{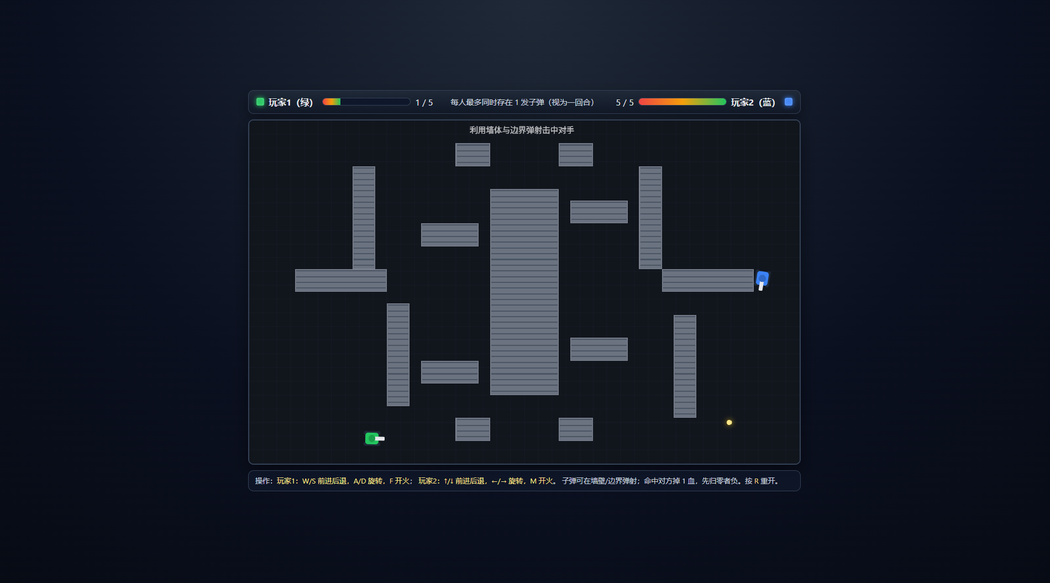}\par\medskip
  \caption{\textbf{Two-player tank-game comparison (1/2).} The Chinese request asks for a console-style tank duel with distinct player colors, five health points per player, one shot per turn, ricocheting bullets, and an arena that obstructs direct shots.}
  \label{fig:case-tanks-a}
\end{figure*}

\begin{figure*}[p]
  \centering
  \caseGameHeader
  \noindent\textbf{Claude-Opus-4.7}\hfill{\small\color{black!70}}\par\smallskip
\includegraphics[width=0.49\textwidth]{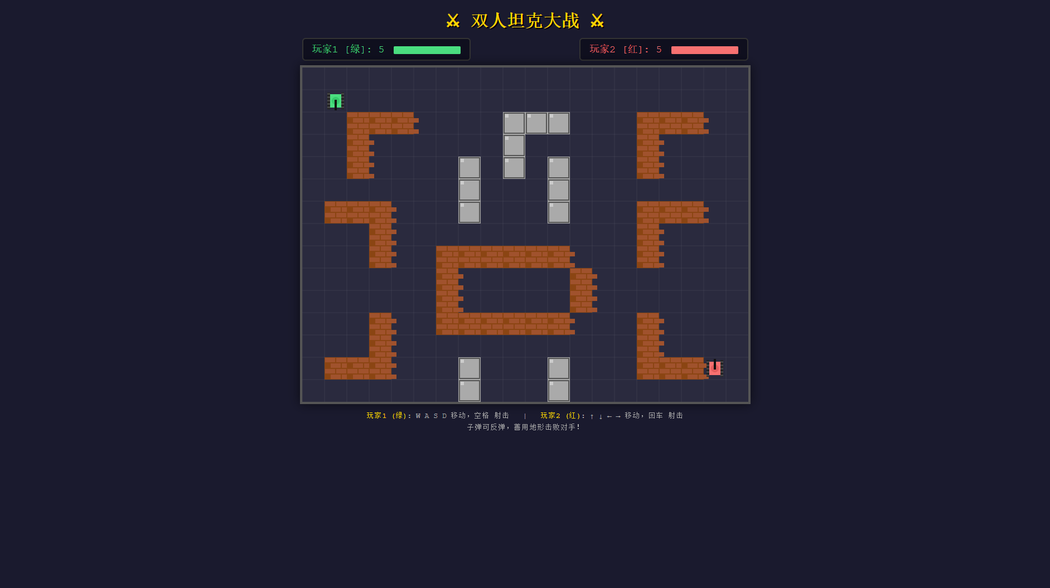}\hfill
\includegraphics[width=0.49\textwidth]{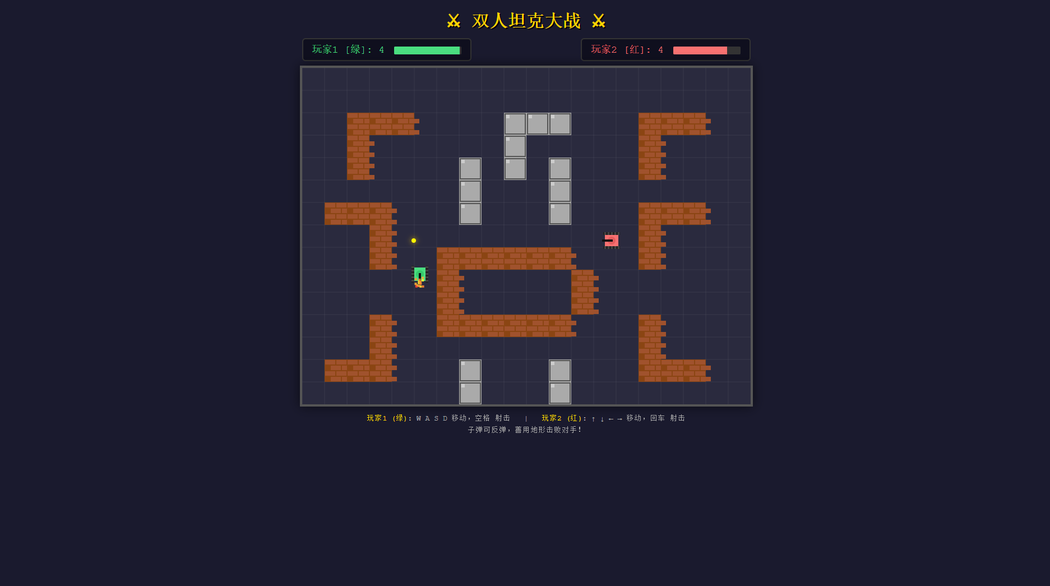}\par\medskip
  \noindent\textbf{GPT-5.4-mini}\hfill{\small\color{black!70}Cluttered map layout}\par\smallskip
\includegraphics[width=0.49\textwidth]{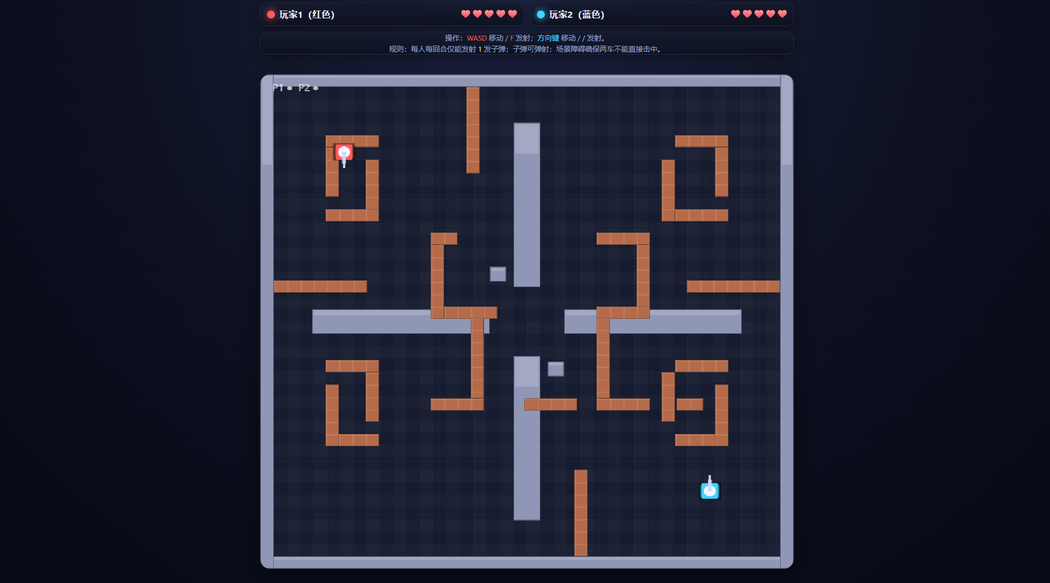}\hfill
\includegraphics[width=0.49\textwidth]{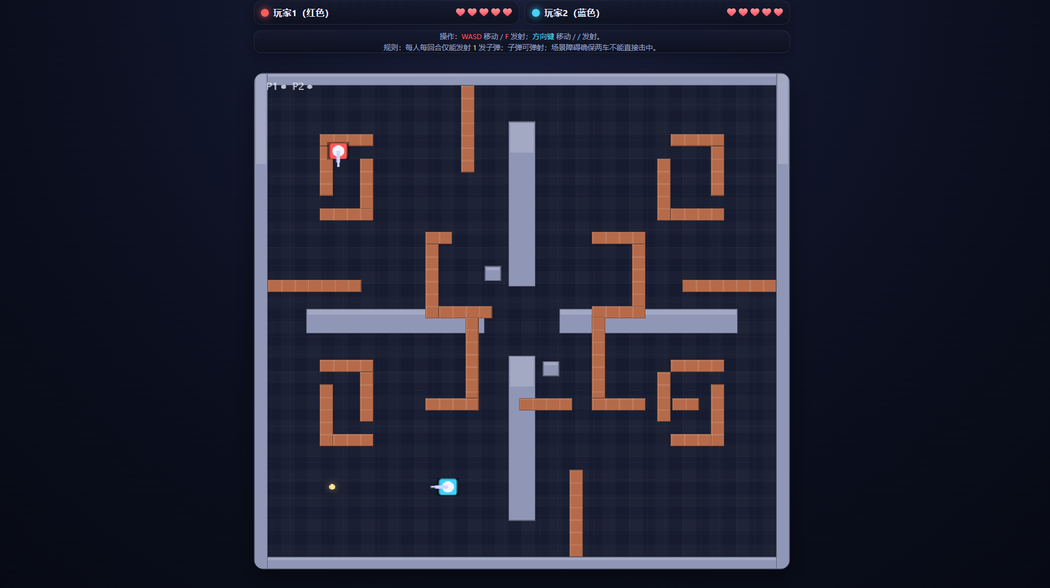}\par\medskip
  \noindent\textbf{WebCatalyst}\hfill{\small\color{black!70}Projectile and impact effects}\par\smallskip
  \includegraphics[width=0.49\textwidth]{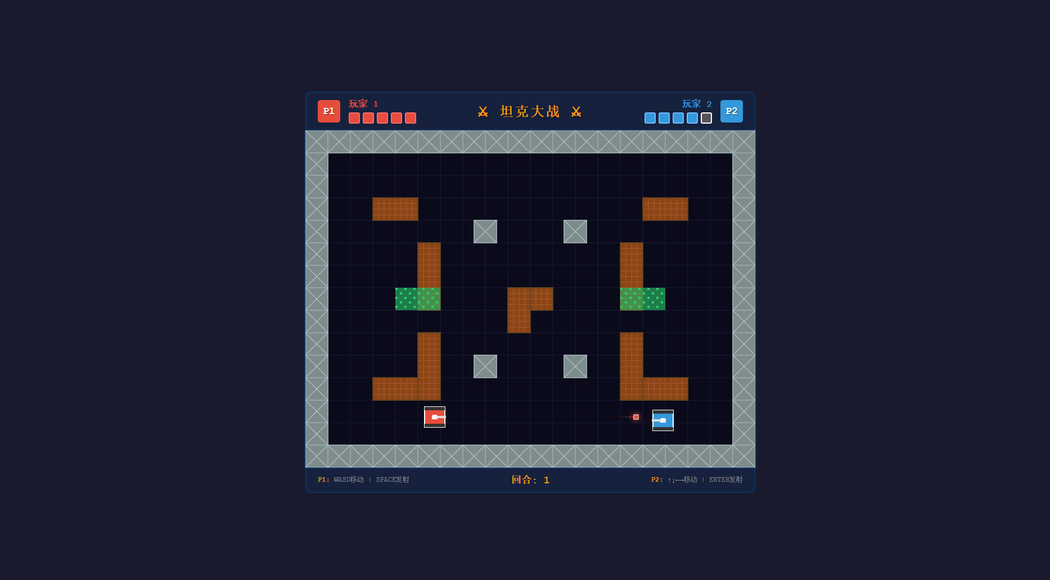}\hfill
  \includegraphics[width=0.49\textwidth]{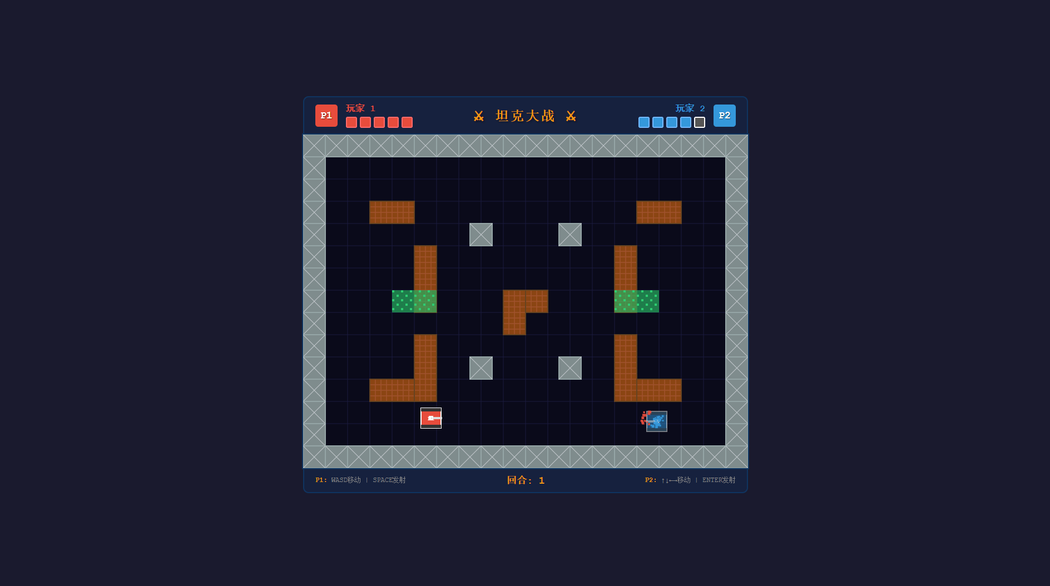}\par\medskip
  \caption{\textbf{Two-player tank-game comparison (2/2).} Continuation of Figure~\ref{fig:case-tanks-a}. WebCatalyst's paired views show a projectile and impact particles around the blue tank.}
  \label{fig:case-tanks-b}
\end{figure*}

\clearpage
\begin{figure*}[p]
  \centering
  \begin{minipage}{\textwidth}
\subsubsection{Transfer to Conventional Webpages}
\label{app:webpage-cases}

  \end{minipage}\par\medskip
  \begin{minipage}[t]{0.31\textwidth}\centering
  {\small\bfseries Qwen3.6-35B-A3B\strut}\par\smallskip
  \includegraphics[width=\linewidth]{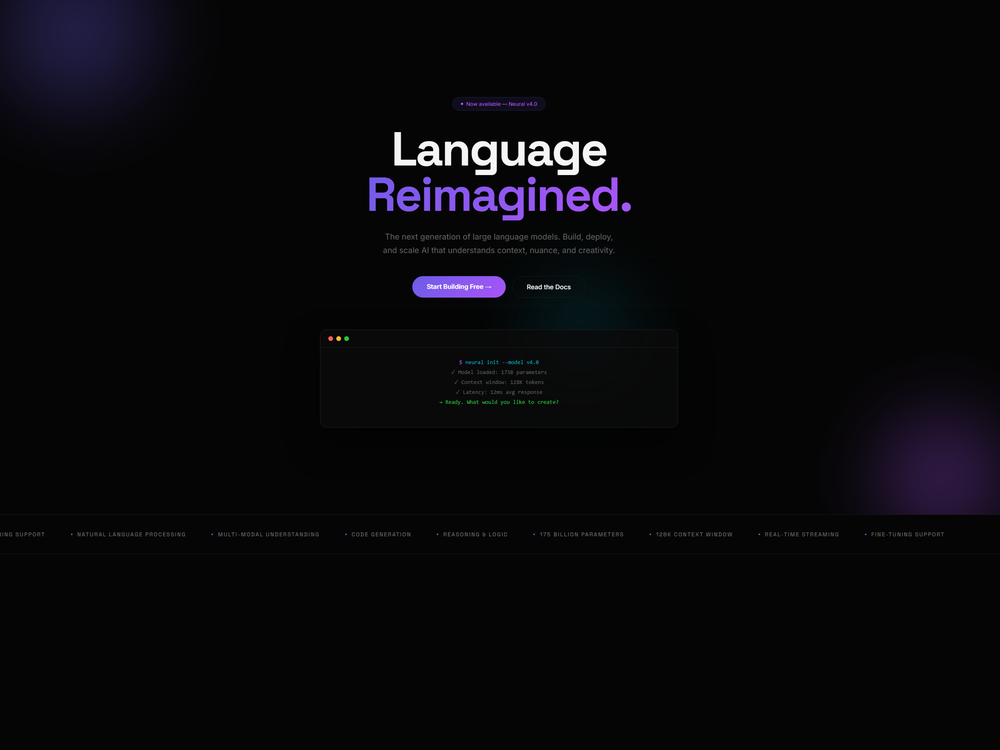}\par\smallskip
  {\footnotesize Hero detail}\par\medskip
  \begin{minipage}[t][0.49\textheight][t]{\linewidth}\centering
    \includegraphics[width=\linewidth,height=0.49\textheight,keepaspectratio]{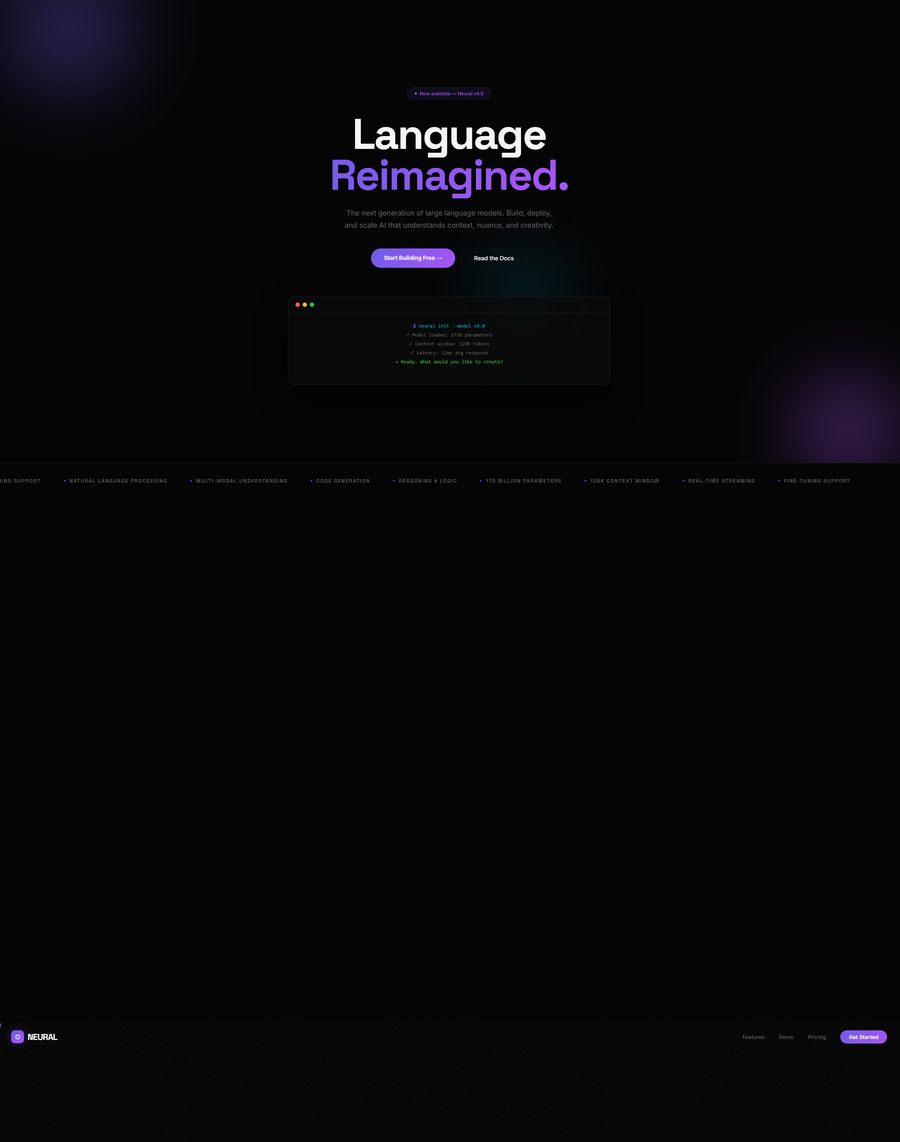}
  \end{minipage}\par\smallskip
  {\footnotesize Full-page overview}
\end{minipage}\hfill
  \begin{minipage}[t]{0.31\textwidth}\centering
  {\small\bfseries Gemini-3.1-Pro\strut}\par\smallskip
  \includegraphics[width=\linewidth]{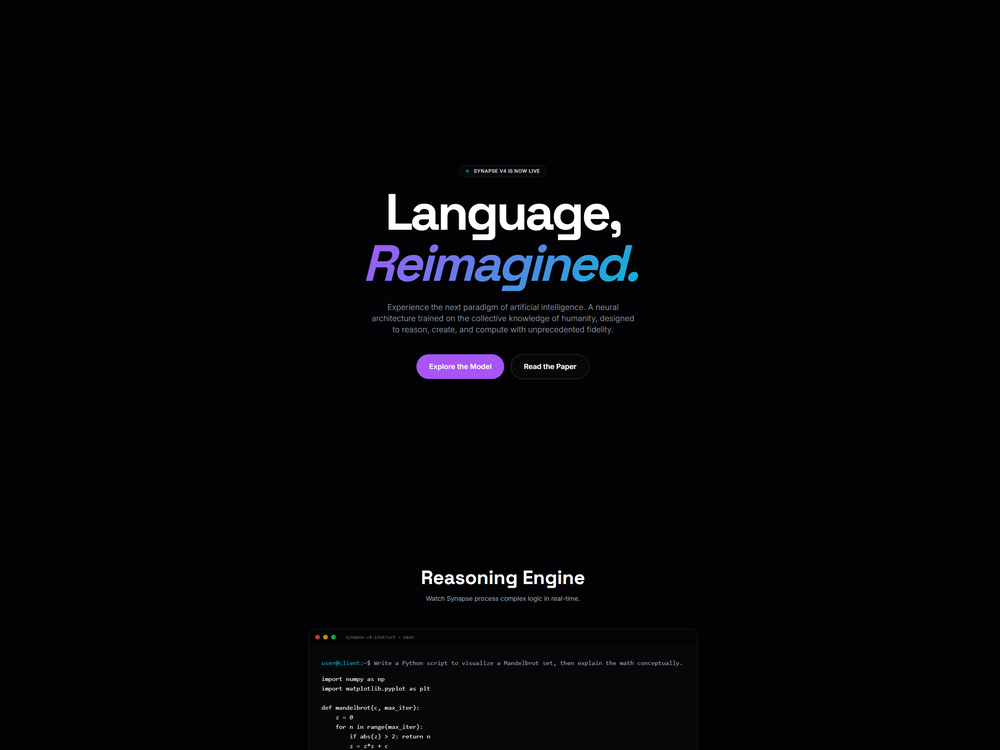}\par\smallskip
  {\footnotesize Hero detail}\par\medskip
  \begin{minipage}[t][0.49\textheight][t]{\linewidth}\centering
    \includegraphics[width=\linewidth,height=0.49\textheight,keepaspectratio]{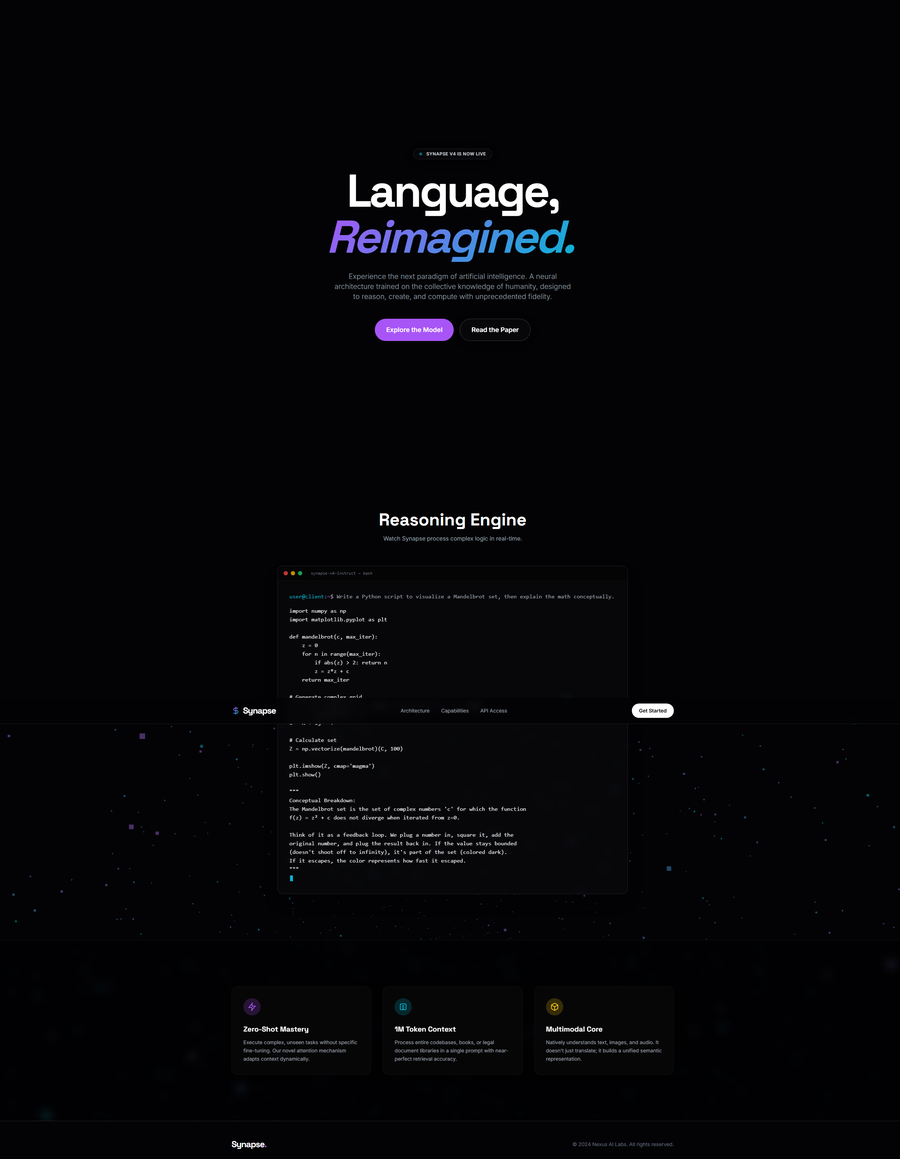}
  \end{minipage}\par\smallskip
  {\footnotesize Full-page overview}
\end{minipage}\hfill
  \begin{minipage}[t]{0.31\textwidth}\centering
  {\small\bfseries Claude-Opus-4.7\strut}\par\smallskip
  \includegraphics[width=\linewidth]{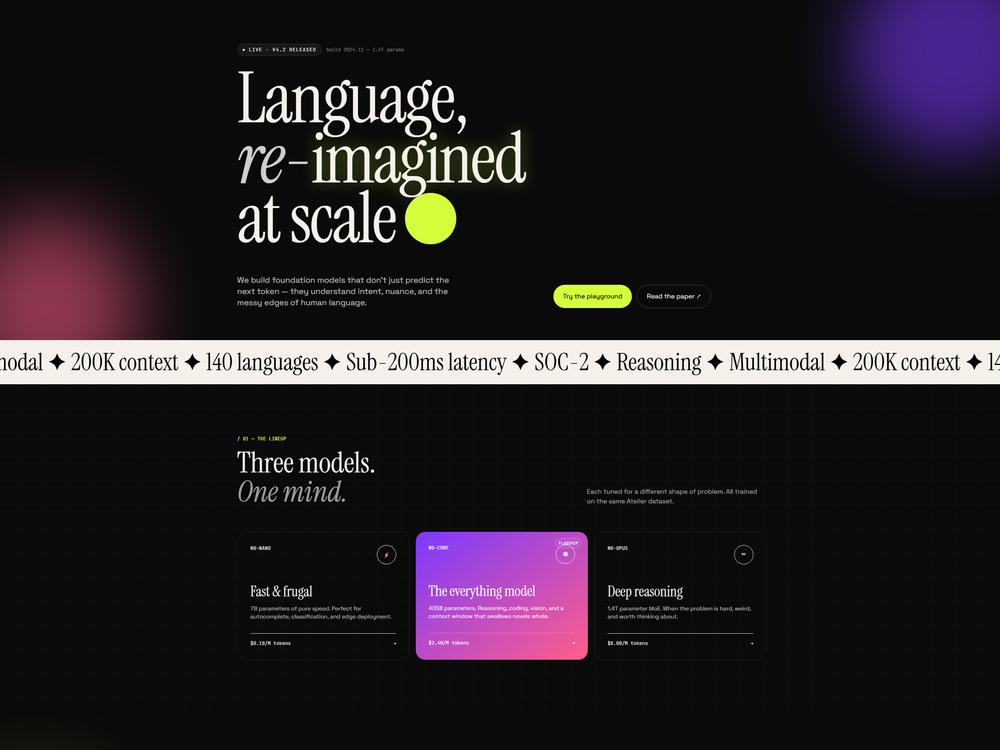}\par\smallskip
  {\footnotesize Hero detail}\par\medskip
  \begin{minipage}[t][0.49\textheight][t]{\linewidth}\centering
    \includegraphics[width=\linewidth,height=0.49\textheight,keepaspectratio]{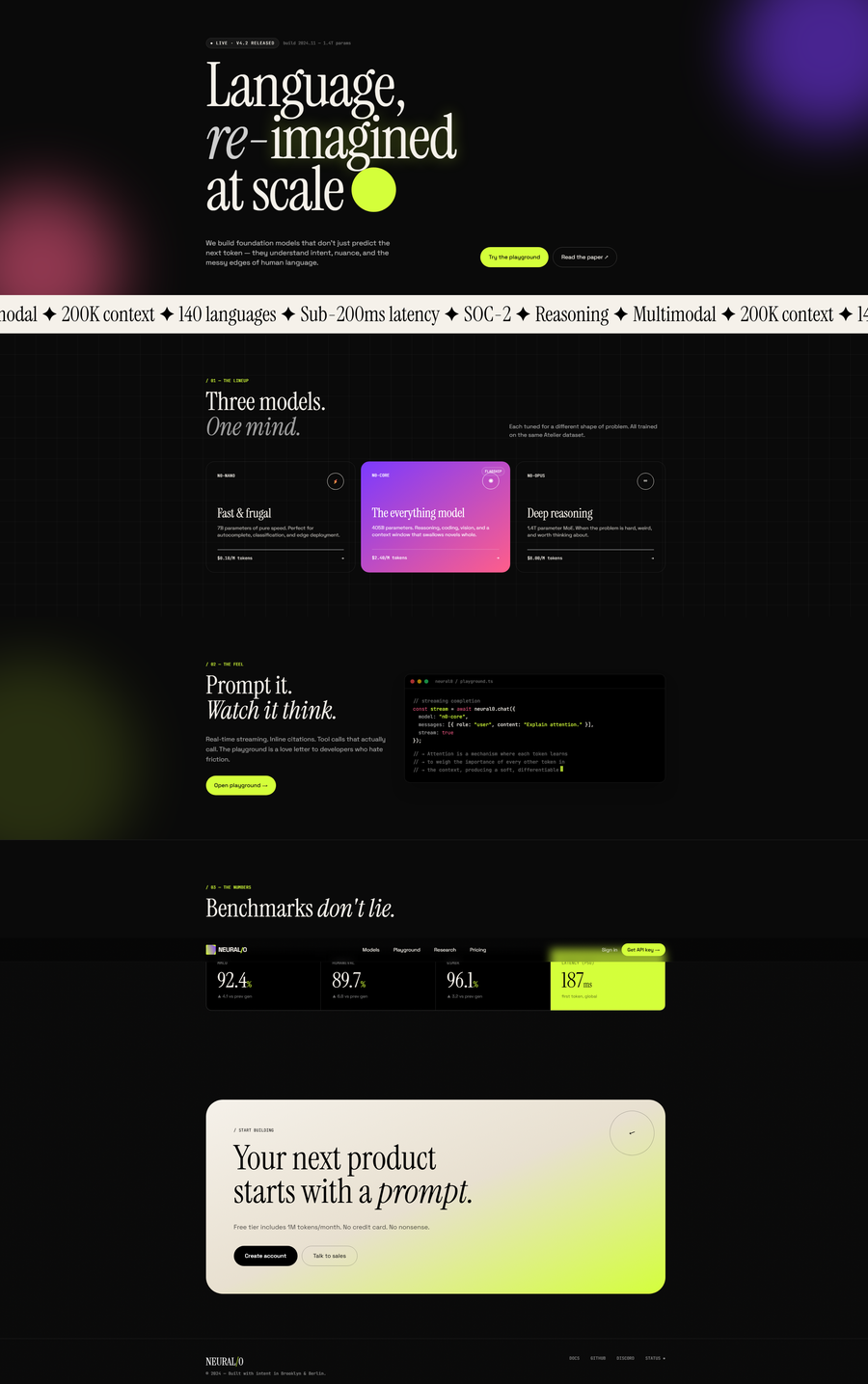}
  \end{minipage}\par\smallskip
  {\footnotesize Full-page overview}
\end{minipage}
  \caption{\textbf{LLM product-website comparison (1/2).} The enlarged top sections show typography, color, and hero composition; the full-page views show the organization and extent of the generated content.}
  \label{fig:case-llm-a}
\end{figure*}

\begin{figure*}[p]
  \centering
  \begin{minipage}[t]{0.43\textwidth}\centering
  {\small\bfseries GPT-5.4-mini\strut}\par\smallskip
  \includegraphics[width=\linewidth]{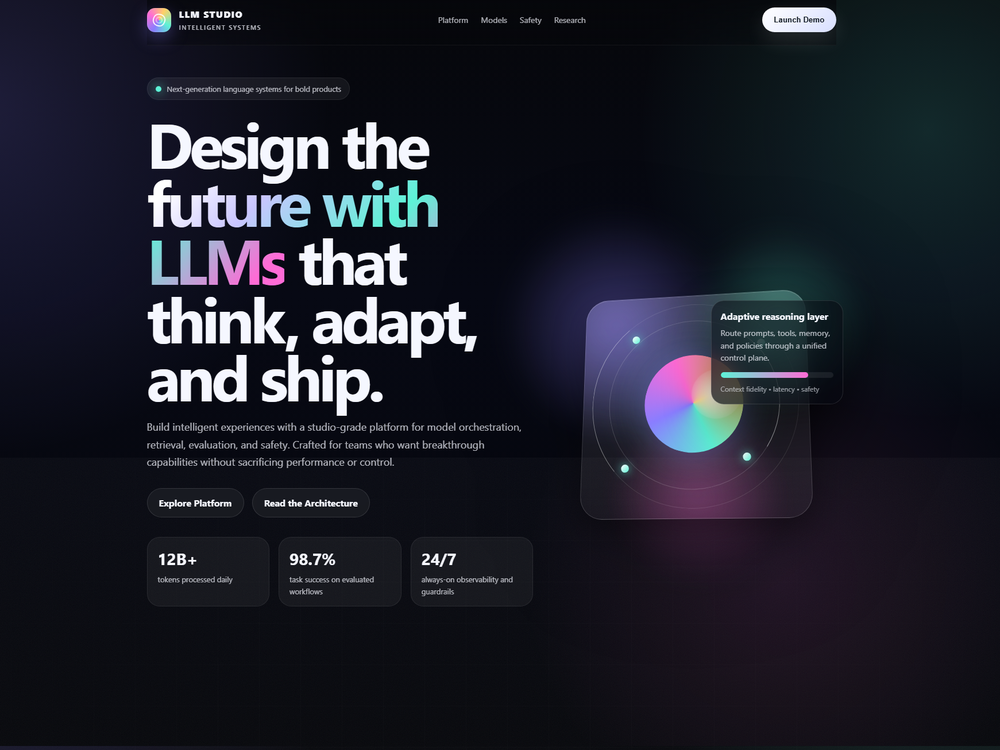}\par\smallskip
  {\footnotesize Hero detail}\par\medskip
  \begin{minipage}[t][0.49\textheight][t]{\linewidth}\centering
    \includegraphics[width=\linewidth,height=0.49\textheight,keepaspectratio]{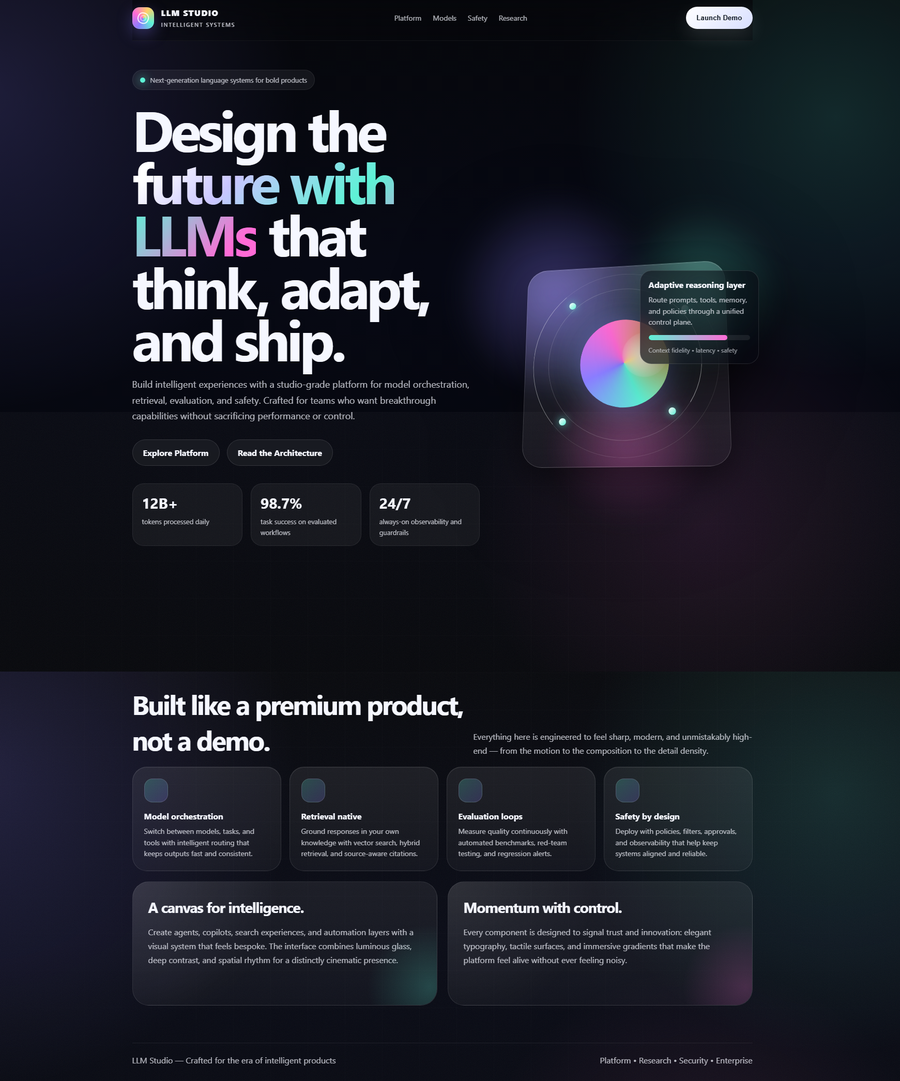}
  \end{minipage}\par\smallskip
  {\footnotesize Full-page overview}
\end{minipage}\hspace{0.035\textwidth}
  \begin{minipage}[t]{0.43\textwidth}\centering
  {\small\bfseries WebCatalyst\strut}\par\smallskip
  \includegraphics[width=\linewidth]{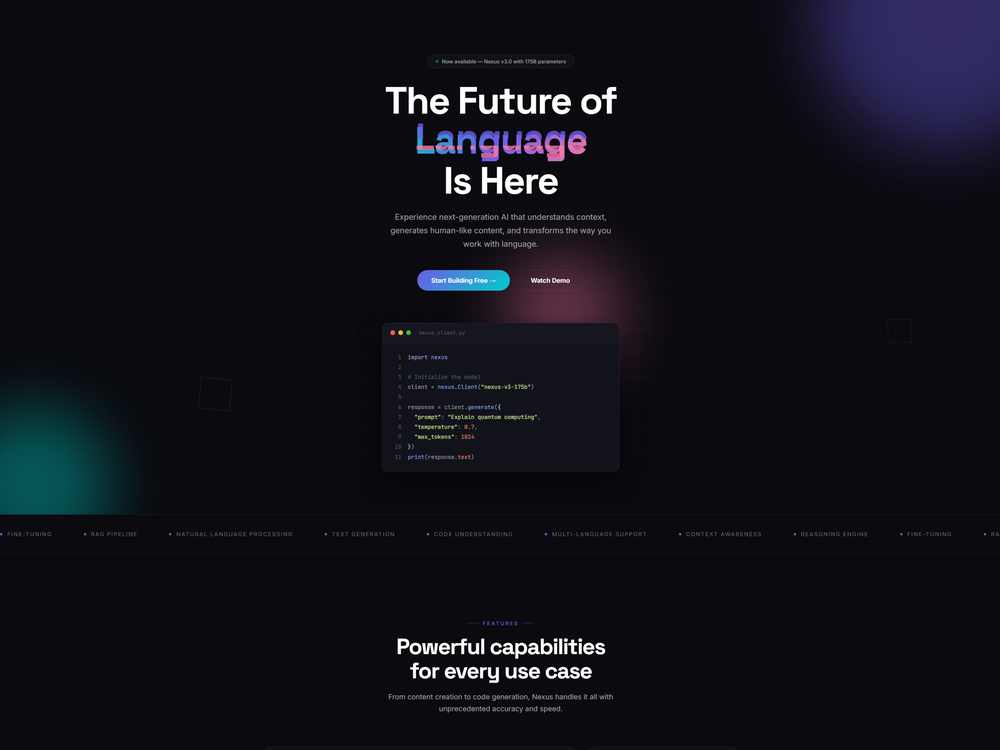}\par\smallskip
  {\footnotesize Hero detail}\par\medskip
  \begin{minipage}[t][0.49\textheight][t]{\linewidth}\centering
    \includegraphics[width=\linewidth,height=0.49\textheight,keepaspectratio]{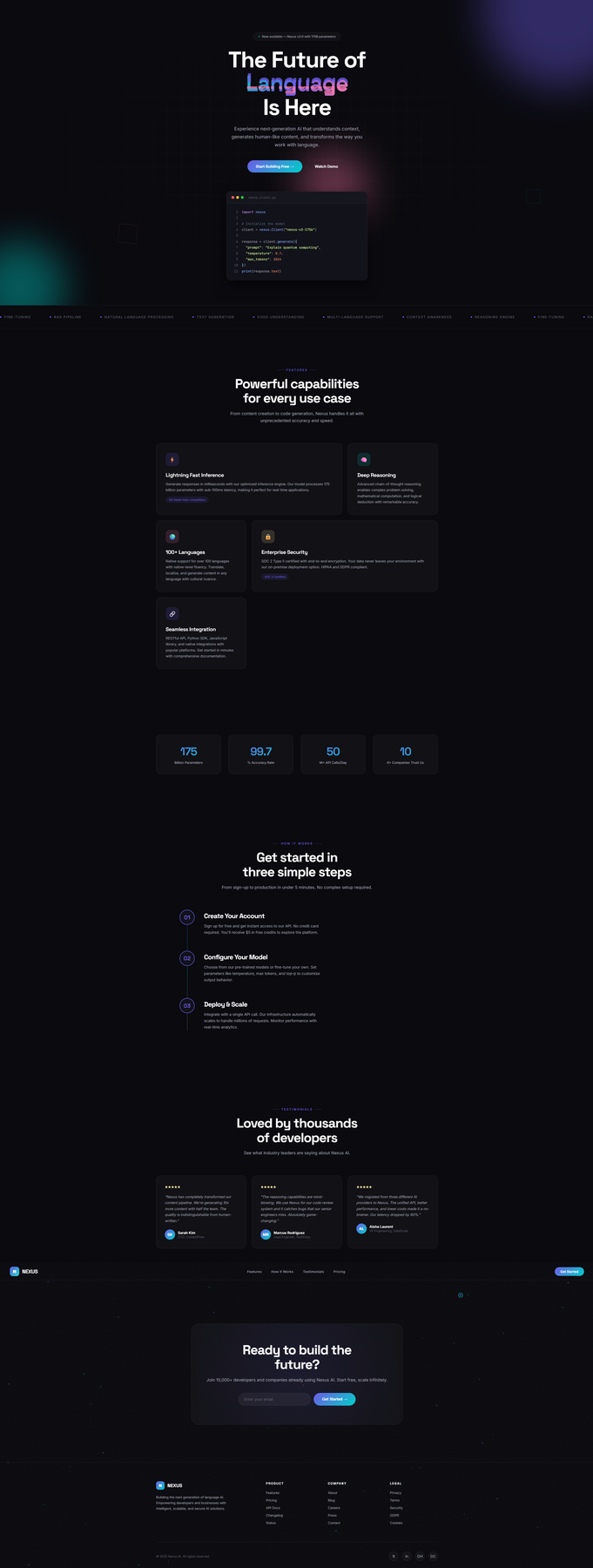}
  \end{minipage}\par\smallskip
  {\footnotesize Full-page overview}
\end{minipage}
  \caption{\textbf{LLM product-website comparison (2/2).} Continuation of Figure~\ref{fig:case-llm-a}. WebCatalyst maintains the dark theme and shows diverse features across the full page.}
  \label{fig:case-llm-b}
\end{figure*}

\begin{figure*}[p]
  \centering
  \begin{minipage}[t]{0.31\textwidth}\centering
  {\small\bfseries Qwen3.6-35B-A3B\strut}\par\smallskip
  \includegraphics[width=\linewidth]{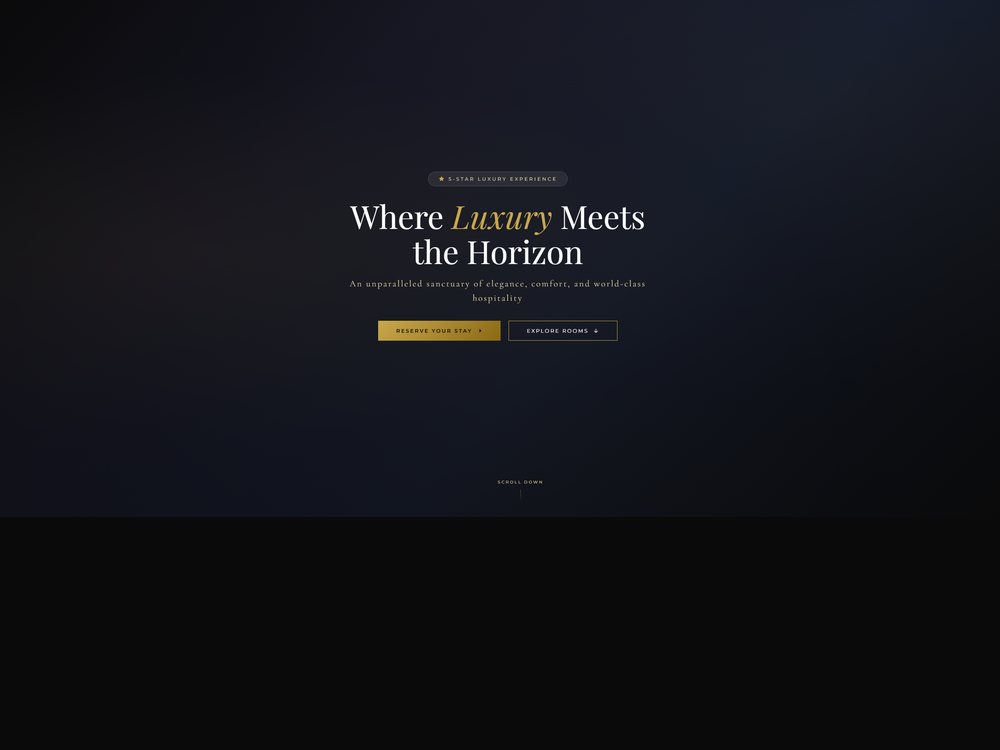}\par\smallskip
  {\footnotesize Hero detail}\par\medskip
  \begin{minipage}[t][0.49\textheight][t]{\linewidth}\centering
    \includegraphics[width=\linewidth,height=0.49\textheight,keepaspectratio]{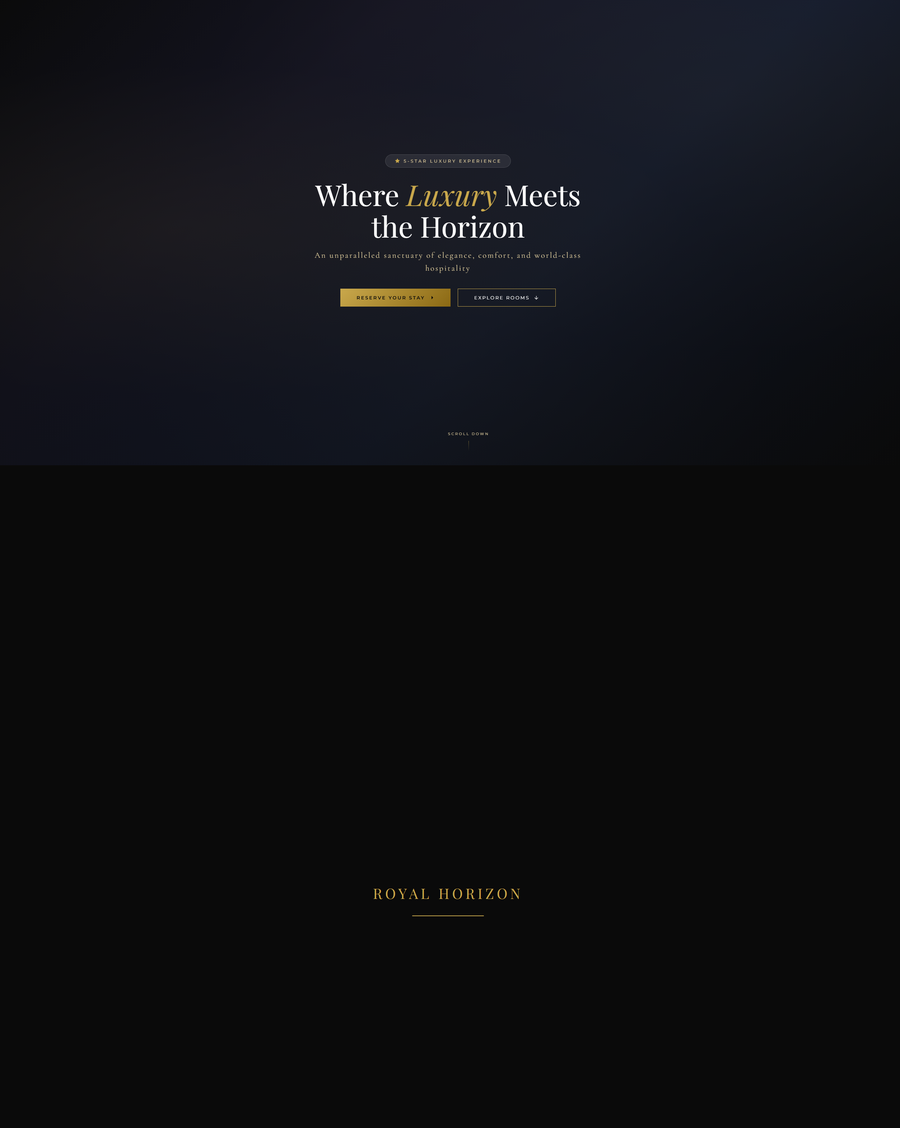}
  \end{minipage}\par\smallskip
  {\footnotesize Full-page overview}
\end{minipage}\hfill
  \begin{minipage}[t]{0.31\textwidth}\centering
  {\small\bfseries Gemini-3.1-Pro\strut}\par\smallskip
  \includegraphics[width=\linewidth]{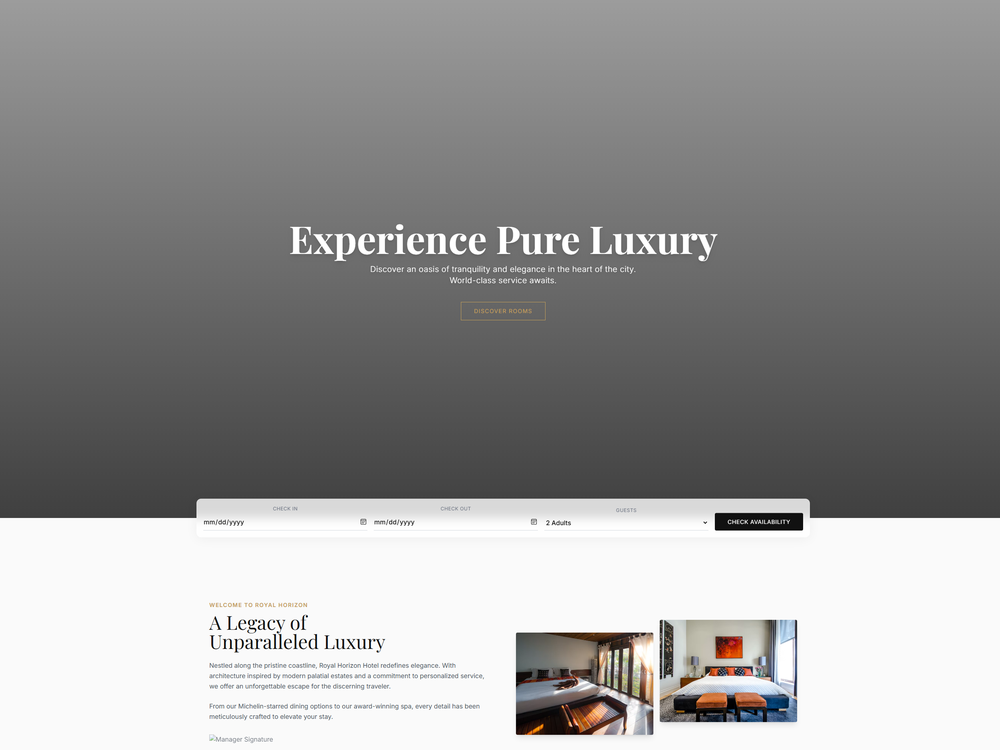}\par\smallskip
  {\footnotesize Hero detail}\par\medskip
  \begin{minipage}[t][0.49\textheight][t]{\linewidth}\centering
    \includegraphics[width=\linewidth,height=0.49\textheight,keepaspectratio]{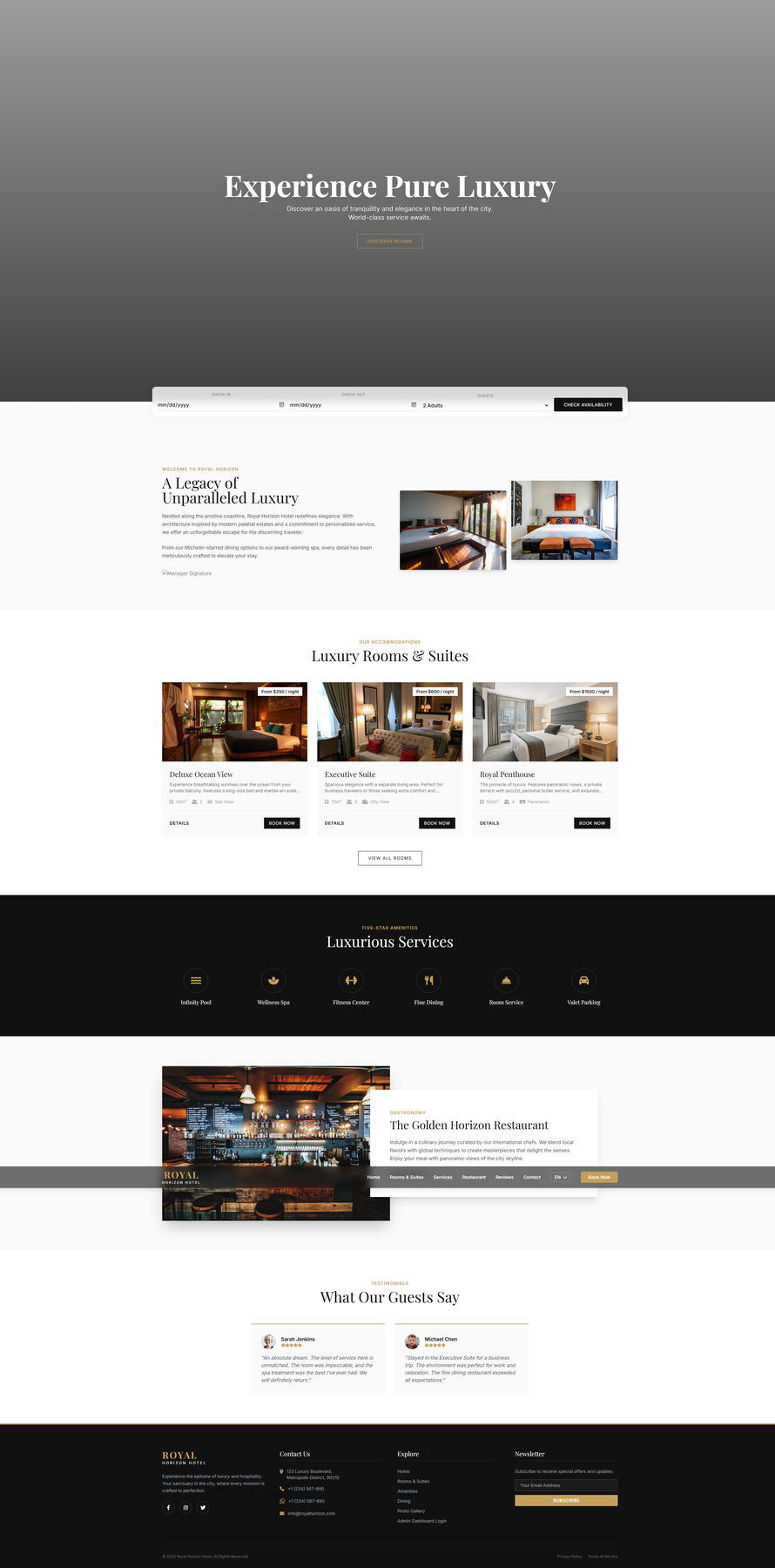}
  \end{minipage}\par\smallskip
  {\footnotesize Full-page overview}
\end{minipage}\hfill
  \begin{minipage}[t]{0.31\textwidth}\centering
  {\small\bfseries Claude-Opus-4.7\strut}\par\smallskip
  \includegraphics[width=\linewidth]{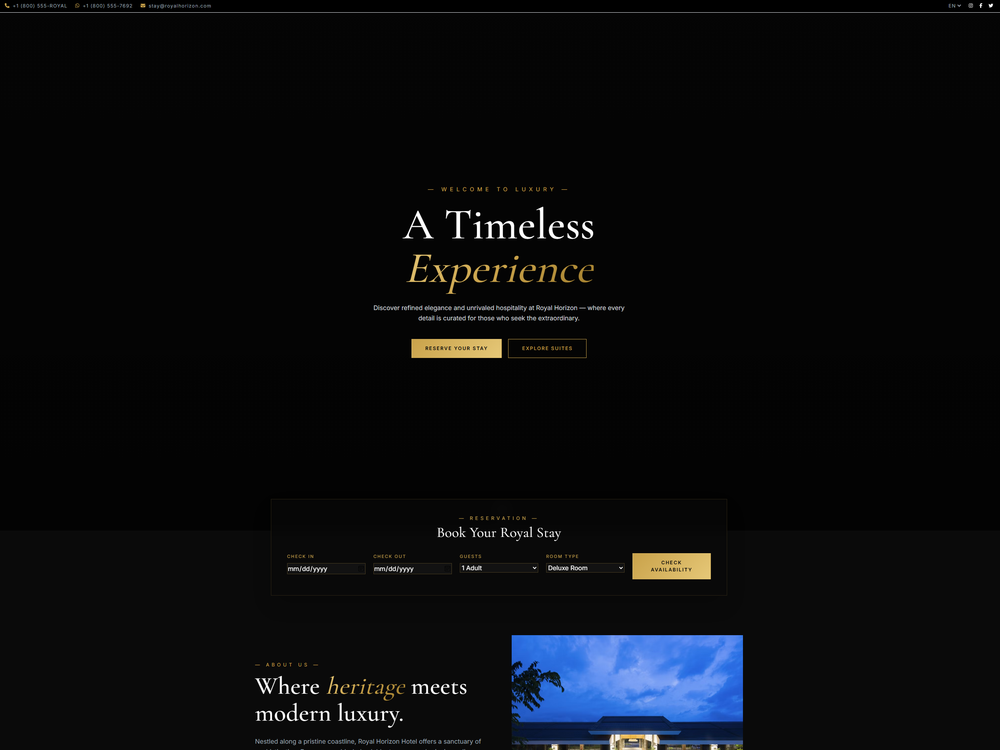}\par\smallskip
  {\footnotesize Hero detail}\par\medskip
  \begin{minipage}[t][0.49\textheight][t]{\linewidth}\centering
    \includegraphics[width=\linewidth,height=0.49\textheight,keepaspectratio]{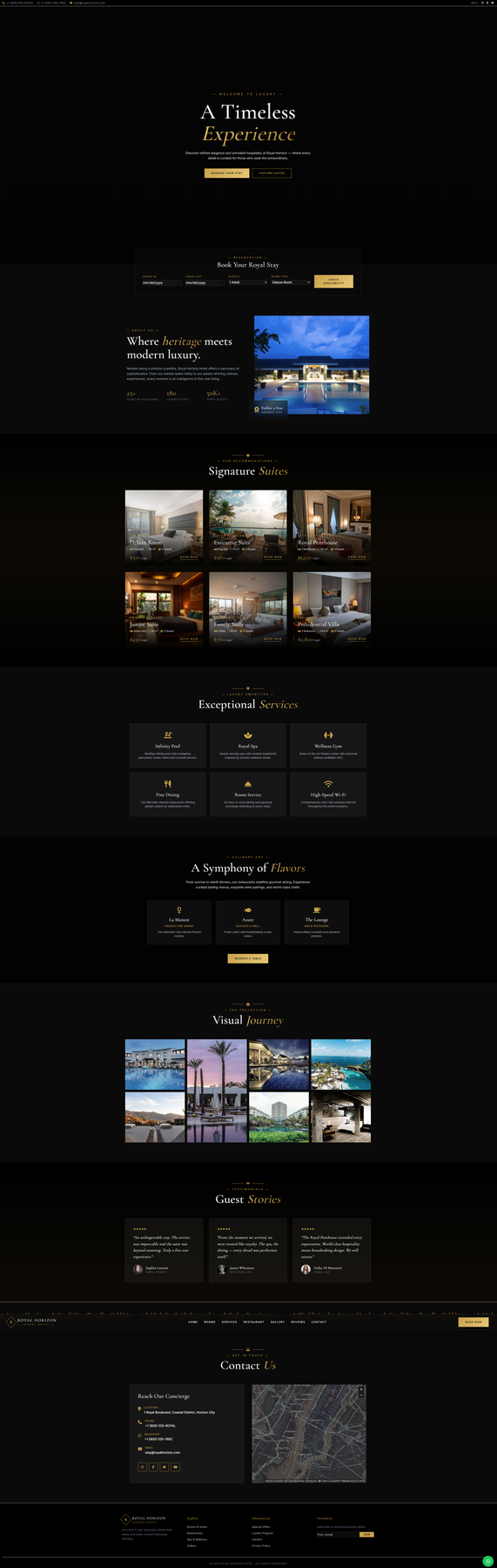}
  \end{minipage}\par\smallskip
  {\footnotesize Full-page overview}
\end{minipage}
  \caption{\textbf{Hotel-website comparison (1/2).} The hero details show the landing-page presentation; the full-page overviews show accommodations, services, visual content, and lower-page sections.}
  \label{fig:case-hotel-a}
\end{figure*}

\begin{figure*}[p]
  \centering
  \begin{minipage}[t]{0.43\textwidth}\centering
  {\small\bfseries GPT-5.4-mini\strut}\par\smallskip
  \includegraphics[width=\linewidth]{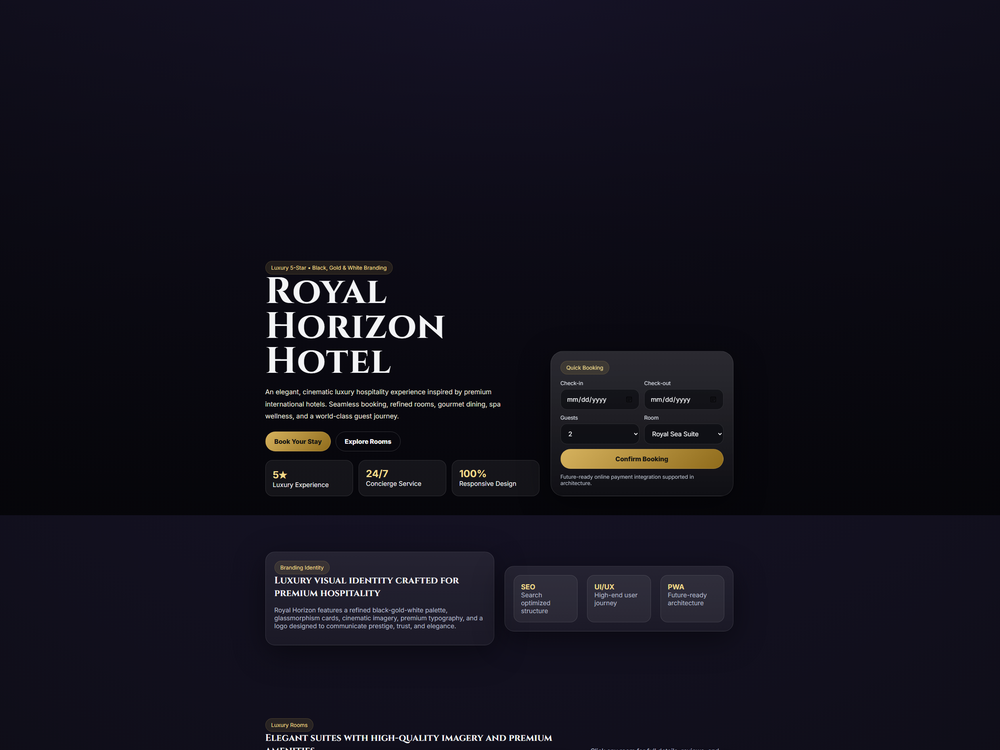}\par\smallskip
  {\footnotesize Hero detail}\par\medskip
  \begin{minipage}[t][0.49\textheight][t]{\linewidth}\centering
    \includegraphics[width=\linewidth,height=0.49\textheight,keepaspectratio]{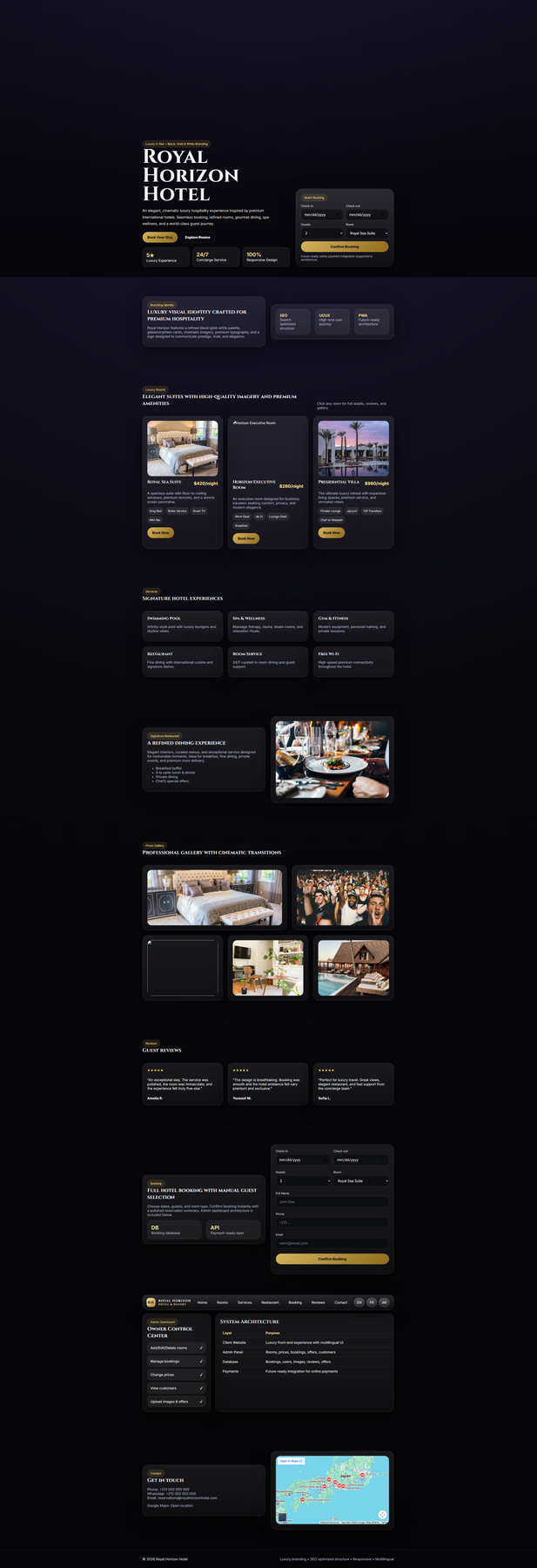}
  \end{minipage}\par\smallskip
  {\footnotesize Full-page overview}
\end{minipage}\hspace{0.035\textwidth}
  \begin{minipage}[t]{0.43\textwidth}\centering
  {\small\bfseries WebCatalyst\strut}\par\smallskip
  \includegraphics[width=\linewidth]{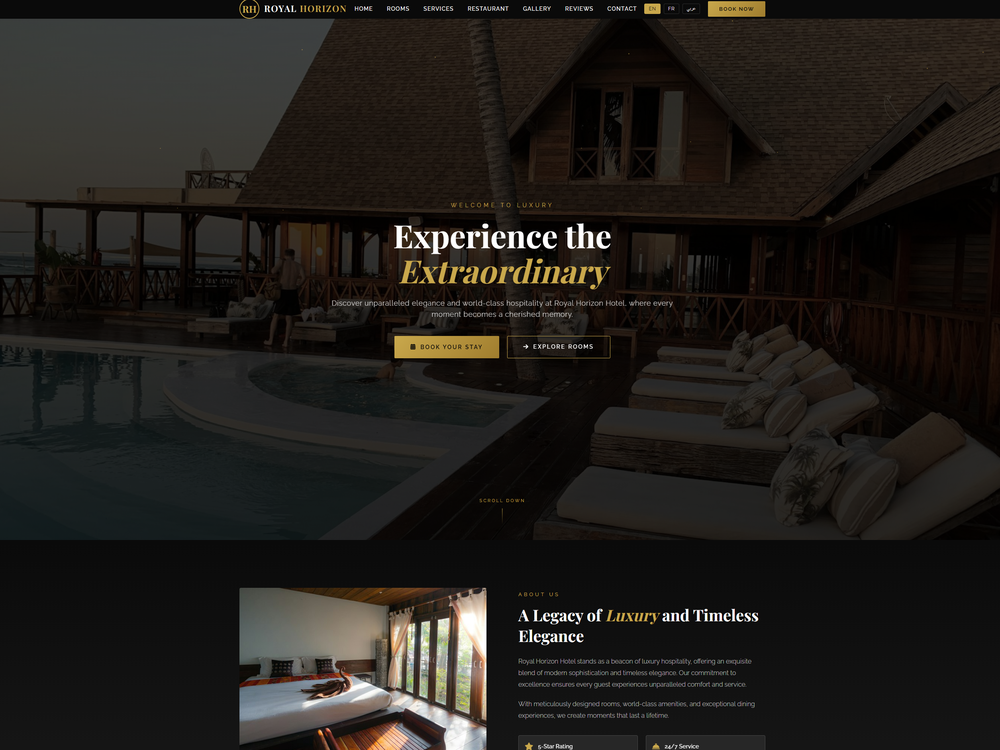}\par\smallskip
  {\footnotesize Hero detail}\par\medskip
  \begin{minipage}[t][0.49\textheight][t]{\linewidth}\centering
    \includegraphics[width=\linewidth,height=0.49\textheight,keepaspectratio]{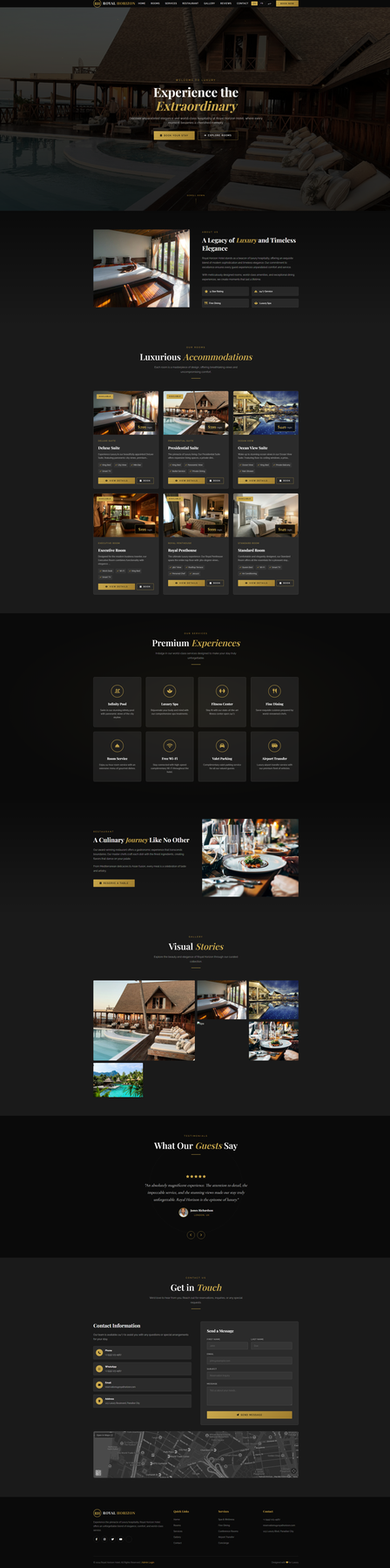}
  \end{minipage}\par\smallskip
  {\footnotesize Full-page overview}
\end{minipage}
  \caption{\textbf{Hotel-website comparison (2/2).} Continuation of Figure~\ref{fig:case-hotel-a}. WebCatalyst combines a photographic hero with image-card layouts, resulting in a visually polished and aesthetically appealing presentation.}
  \label{fig:case-hotel-b}
\end{figure*}
\clearpage
\begin{figure*}[p]
\centering
\begin{minipage}{\textwidth}
\subsubsection{An Illustrative GUI Assessor Trajectory}
\label{app:assessor-example}
\small

\textbf{Illustration:} Figures~\ref{fig:assessor-trajectory-1}--\ref{fig:assessor-trajectory-3} follow an eight-turn GUI Assessor episode on \textit{Loan Approval Race}. Each turn shows the selected action, before/after screenshots, and transition and screenshot judgements, followed by the Meta-Level assessment.
\end{minipage}\par\medskip
\begin{minipage}{\textwidth}\small\textbf{Task:} Create a complete self-contained HTML game called ``Loan Approval Race'' where I play as a banker racing against AI coworkers to review loan files, dodge paperwork chaos, collect bonuses, and reach the approval desk first. Make it browser-based with simple controls, smooth movement, fun finance-themed obstacles, scoring, and a quick restart option.\end{minipage}\par\medskip
\par\noindent\colorbox{blue!6}{\parbox{\dimexpr\textwidth-2\fboxsep\relax}{%
  \footnotesize\bfseries Turn 1\hfill\raisebox{-1pt}{\includegraphics[height=9pt]{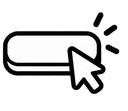}}\enspace Click Start Race}}\par\smallskip
\noindent\begin{minipage}[t]{0.645\textwidth}\vspace{0pt}
  \begin{minipage}[t]{0.485\linewidth}\centering\fontsize{8}{9}\selectfont
    \textbf{Before}\par\smallskip\includegraphics[width=\linewidth,trim=250bp 130bp 250bp 130bp,clip]{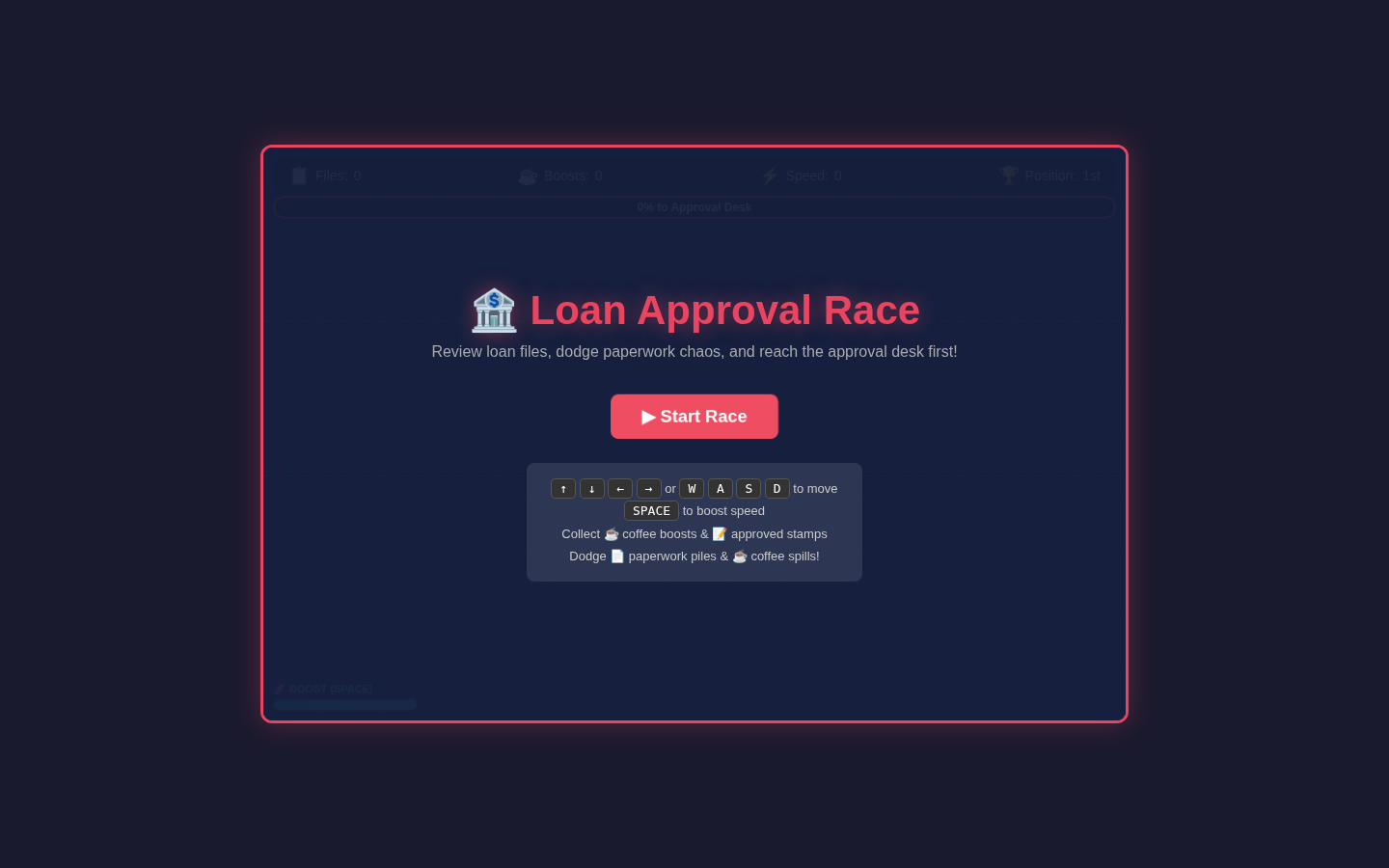}
  \end{minipage}\hfill
  \begin{minipage}[t]{0.485\linewidth}\centering\fontsize{8}{9}\selectfont
    \textbf{After}\par\smallskip\includegraphics[width=\linewidth,trim=250bp 130bp 250bp 130bp,clip]{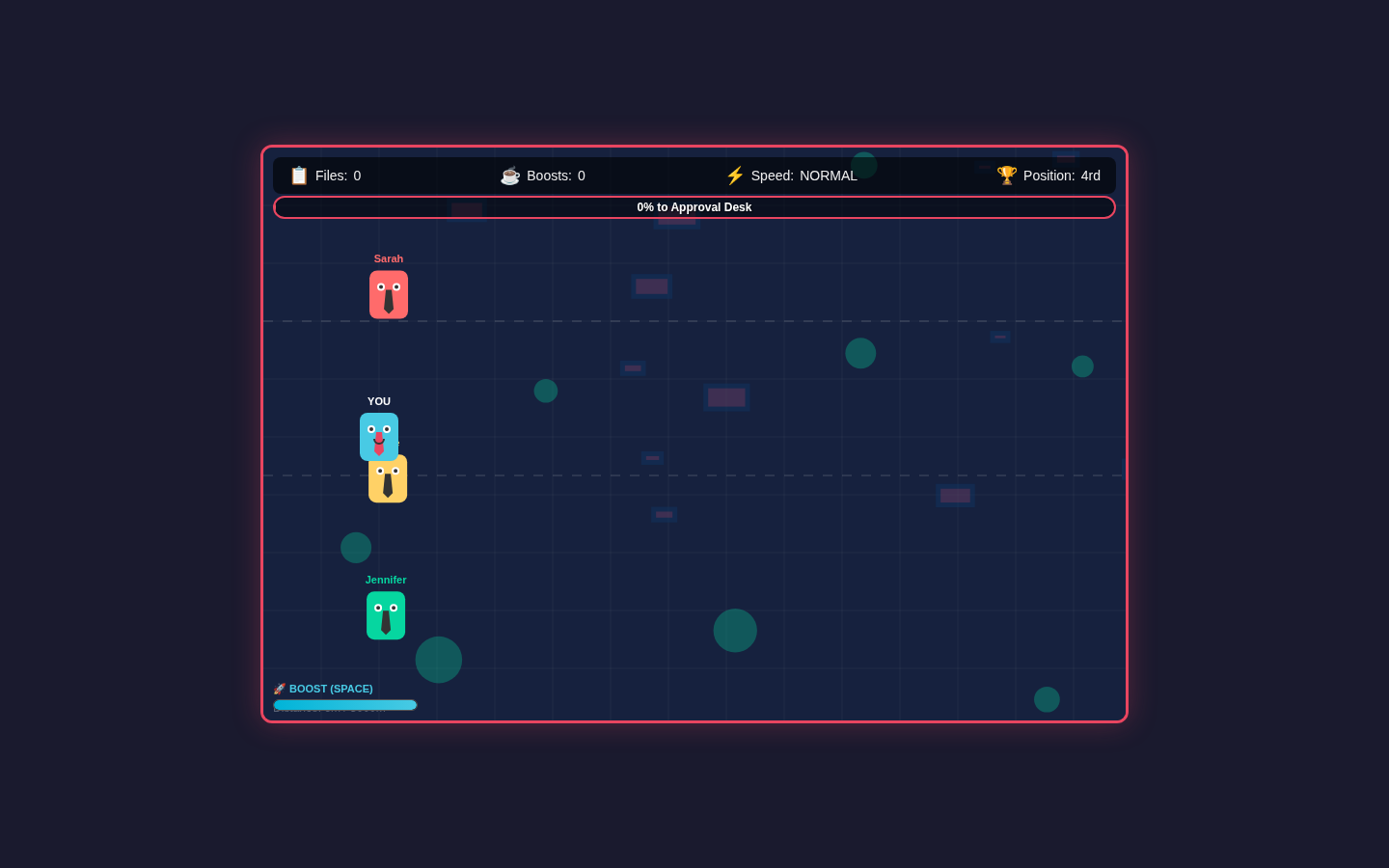}
  \end{minipage}
\end{minipage}\hfill
\begin{minipage}[t]{0.335\textwidth}\vspace{0pt}
  \raggedright\fontsize{8.5}{10}\selectfont 
\textbf{Decision:} Click Start Race to enter the active course.\par\smallskip
\raisebox{-1pt}{\includegraphics[height=9pt]{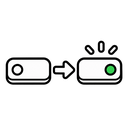}}\enspace\textbf{Transition 93/100:} The start overlay closes and reveals racers, hazards, pickups, and a complete HUD. The active race starts at 0\% progress.\par\smallskip
\raisebox{-1pt}{\includegraphics[height=9pt]{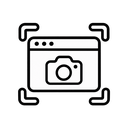}}\enspace\textbf{Screenshot 93/100:} The active course is sharply rendered, with readable resource counters, race position, progress, and boost status.\par\smallskip
\textbf{Judge note:} The HUD displays '4rd' instead of '4th'.

\end{minipage}\par\medskip
\par\noindent\colorbox{blue!6}{\parbox{\dimexpr\textwidth-2\fboxsep\relax}{%
  \footnotesize\bfseries Turn 2\hfill\raisebox{-1pt}{\includegraphics[height=9pt]{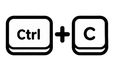}}\enspace Right + Space (300 ms)}}\par\smallskip
\noindent\begin{minipage}[t]{0.645\textwidth}\vspace{0pt}
  \begin{minipage}[t]{0.485\linewidth}\centering\fontsize{8}{9}\selectfont
    \textbf{Before}\par\smallskip\includegraphics[width=\linewidth,trim=250bp 130bp 250bp 130bp,clip]{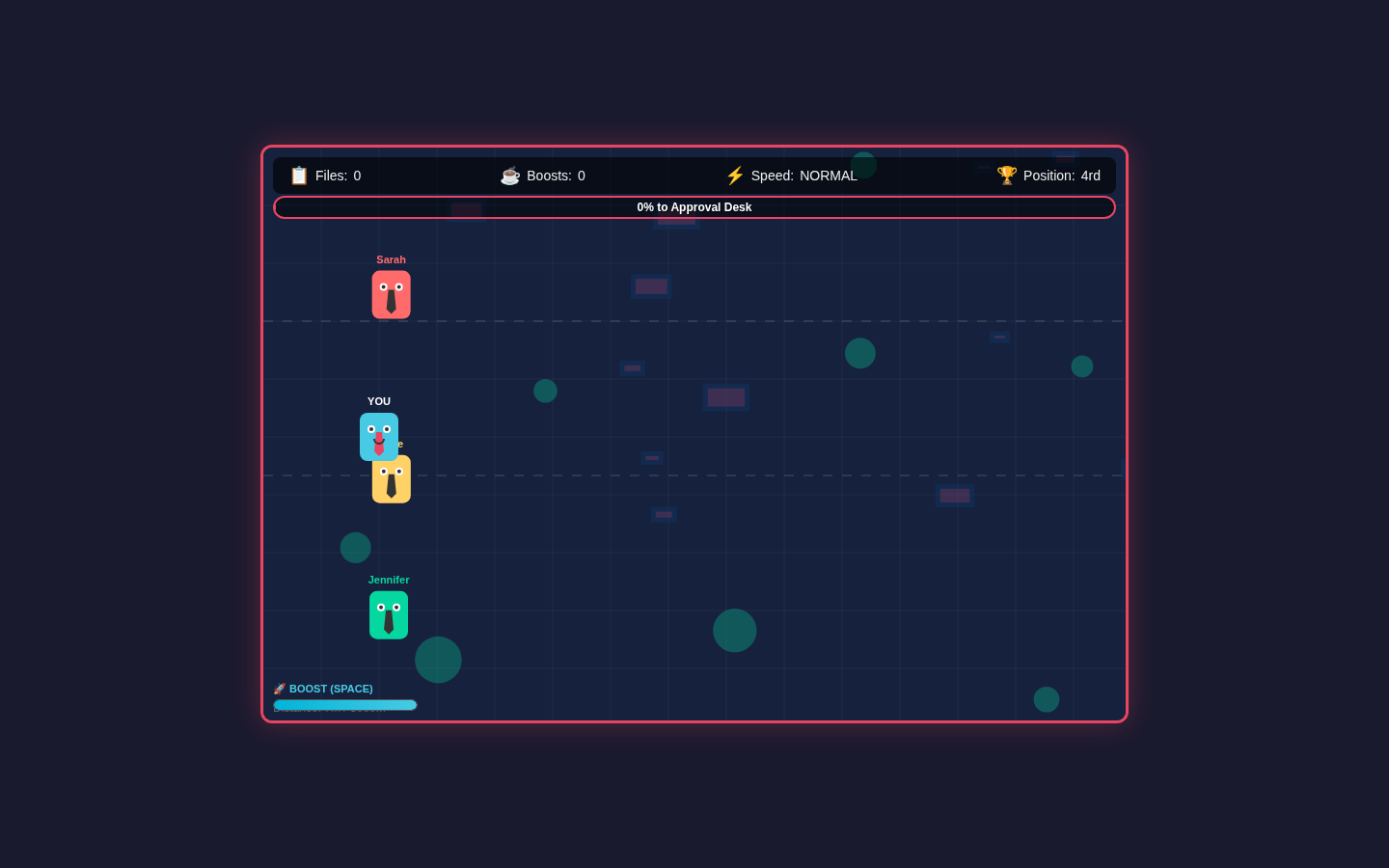}
  \end{minipage}\hfill
  \begin{minipage}[t]{0.485\linewidth}\centering\fontsize{8}{9}\selectfont
    \textbf{After}\par\smallskip\includegraphics[width=\linewidth,trim=250bp 130bp 250bp 130bp,clip]{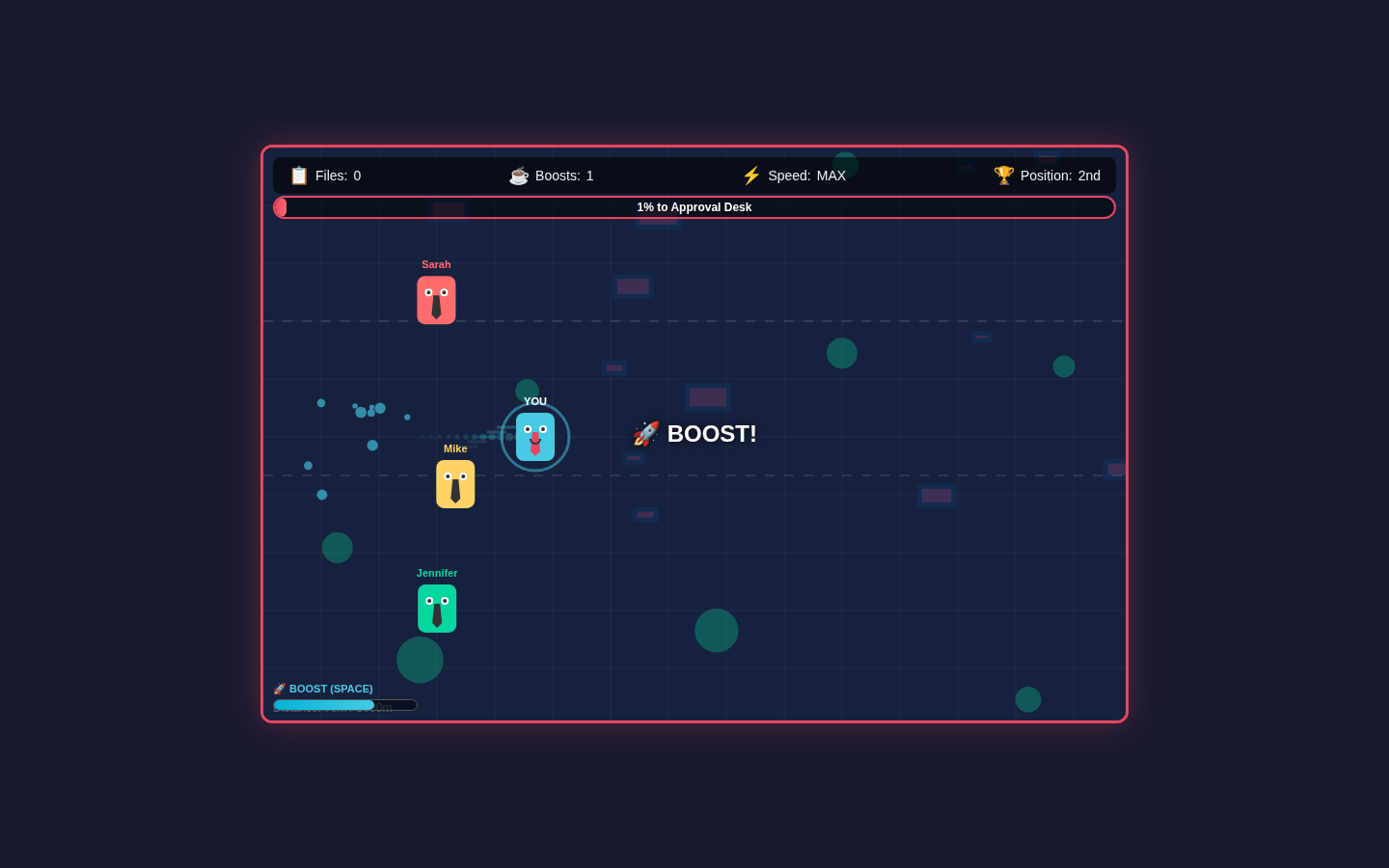}
  \end{minipage}
\end{minipage}\hfill
\begin{minipage}[t]{0.335\textwidth}\vspace{0pt}
  \raggedright\fontsize{8.5}{10}\selectfont 
\textbf{Decision:} From fourth place, move right and boost to gain ground on the AI coworkers.\par\smallskip
\raisebox{-1pt}{\includegraphics[height=9pt]{appendix_assets/icons/transition-judge.png}}\enspace\textbf{Transition 95/100:} The player advances under MAX speed with a boost notification, cyan trail, and particles. Position improves from fourth to second as progress rises to 1\%.\par\smallskip
\raisebox{-1pt}{\includegraphics[height=9pt]{appendix_assets/icons/screenshot-judge.png}}\enspace\textbf{Screenshot 95/100:} The boosted player, rivals, obstacles, movement effects, and updated HUD are clear and complete.\par\smallskip
\textbf{Judge note:} The boost counter rises during activation, making its resource meaning ambiguous.

\end{minipage}\par\medskip
\par\noindent\colorbox{blue!6}{\parbox{\dimexpr\textwidth-2\fboxsep\relax}{%
  \footnotesize\bfseries Turn 3\hfill\raisebox{-1pt}{\includegraphics[height=9pt]{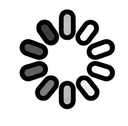}}\enspace Wait (1 s)}}\par\smallskip
\noindent\begin{minipage}[t]{0.645\textwidth}\vspace{0pt}
  \begin{minipage}[t]{0.485\linewidth}\centering\fontsize{8}{9}\selectfont
    \textbf{Before}\par\smallskip\includegraphics[width=\linewidth,trim=250bp 130bp 250bp 130bp,clip]{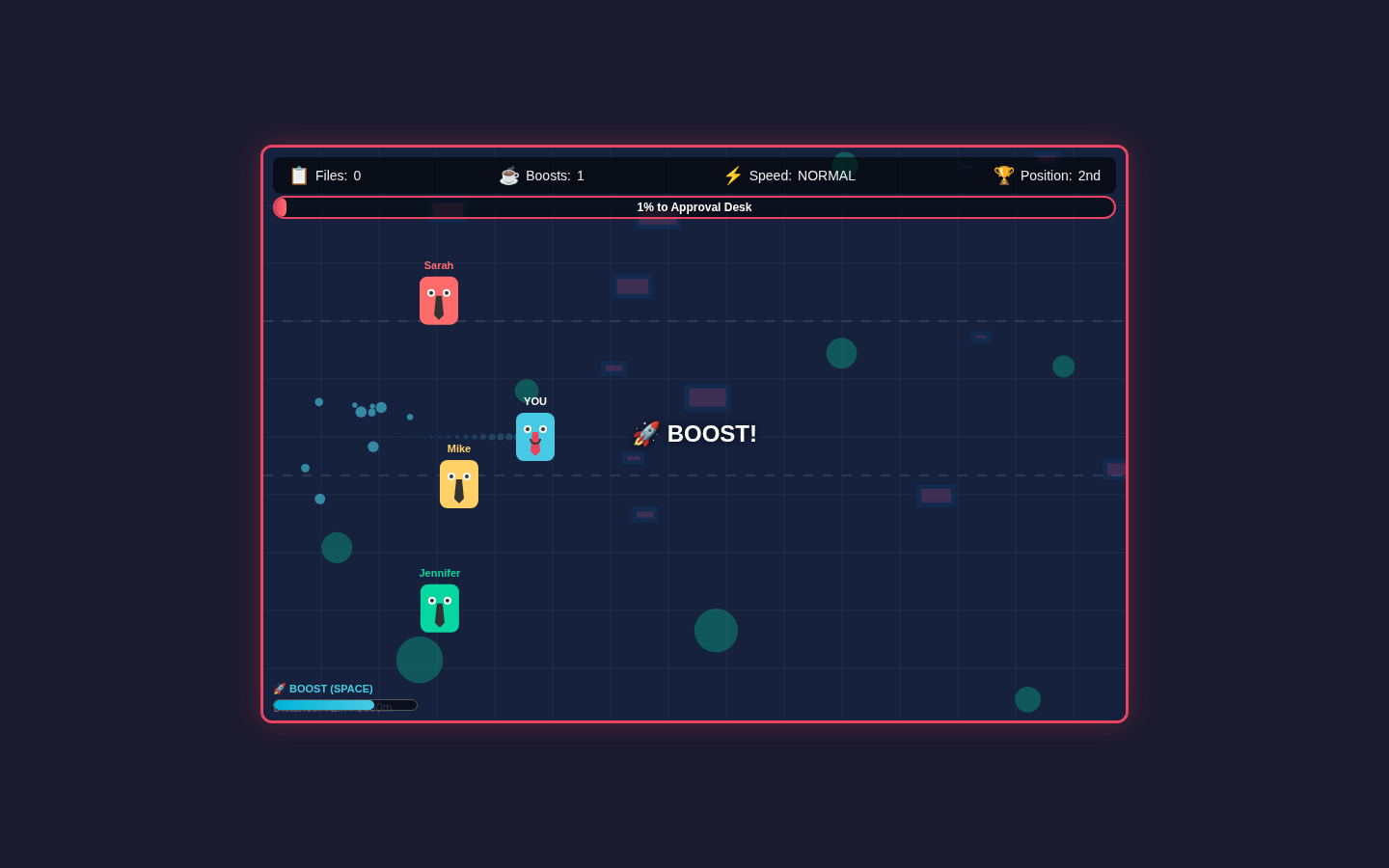}
  \end{minipage}\hfill
  \begin{minipage}[t]{0.485\linewidth}\centering\fontsize{8}{9}\selectfont
    \textbf{After}\par\smallskip\includegraphics[width=\linewidth,trim=250bp 130bp 250bp 130bp,clip]{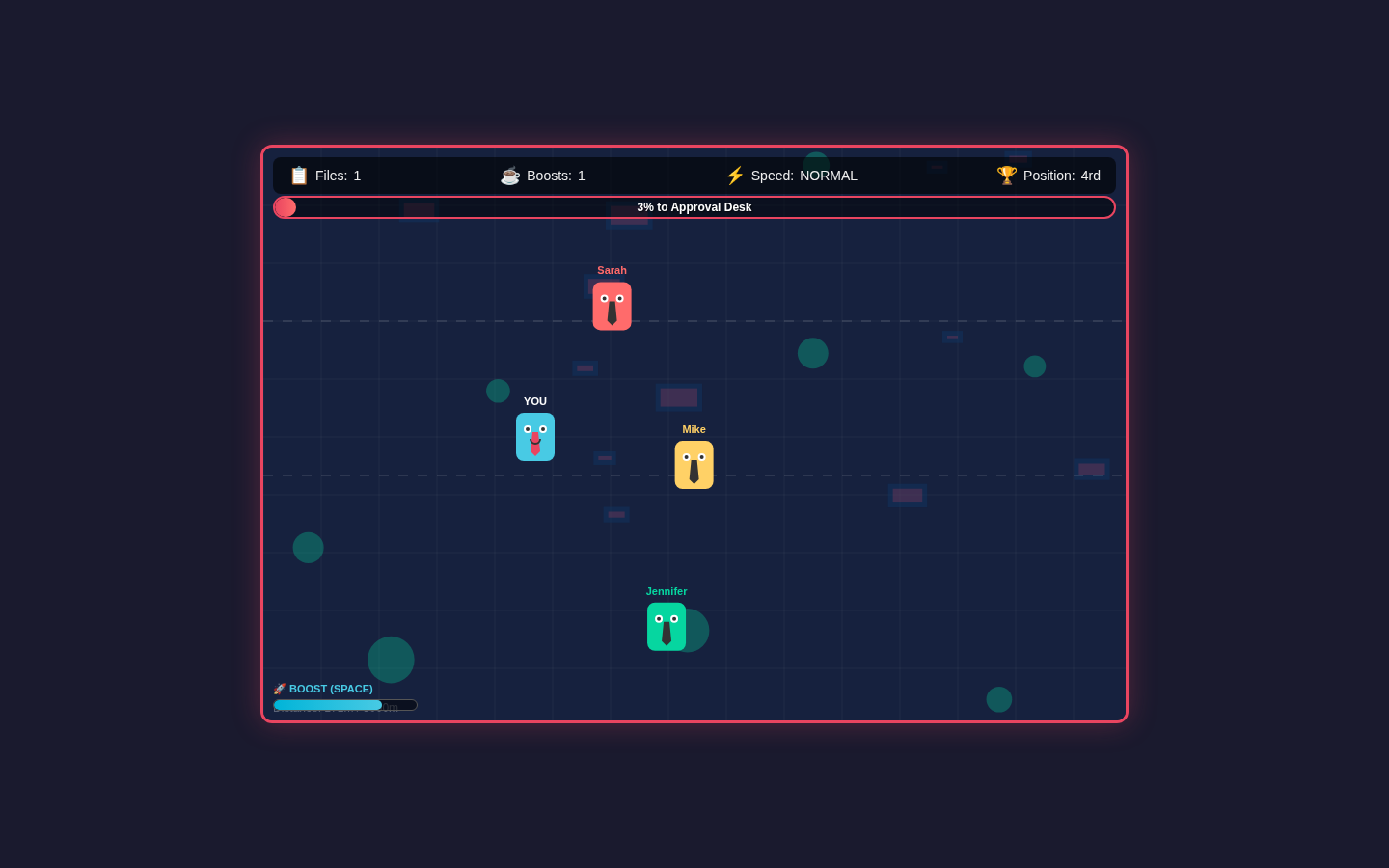}
  \end{minipage}
\end{minipage}\hfill
\begin{minipage}[t]{0.335\textwidth}\vspace{0pt}
  \raggedright\fontsize{8.5}{10}\selectfont 
\textbf{Decision:} Wait in the current lane for boost energy to recover while the race advances.\par\smallskip
\raisebox{-1pt}{\includegraphics[height=9pt]{appendix_assets/icons/transition-judge.png}}\enspace\textbf{Transition 89/100:} Progress rises from 1\% to 3\%, the boost notification disappears, and Files increases to 1. The player falls from second to fourth place.\par\smallskip
\raisebox{-1pt}{\includegraphics[height=9pt]{appendix_assets/icons/screenshot-judge.png}}\enspace\textbf{Screenshot 94/100:} The HUD and racers remain readable; subtle obstacles and the erroneous fourth-place ordinal reduce presentation quality.\par\smallskip
\textbf{Judge note:} The file gain and change in race position have unclear visual causes.

\end{minipage}\par\medskip
\caption{GUI Assessor trajectory on Loan Approval Race, turns 1--3.}
\label{fig:assessor-trajectory-1}
\end{figure*}

\begin{figure*}[p]
\centering
\par\noindent\colorbox{blue!6}{\parbox{\dimexpr\textwidth-2\fboxsep\relax}{%
  \footnotesize\bfseries Turn 4\hfill\raisebox{-1pt}{\includegraphics[height=9pt]{appendix_assets/icons/combo.png}}\enspace Down + Space (1 s)}}\par\smallskip
\noindent\begin{minipage}[t]{0.645\textwidth}\vspace{0pt}
  \begin{minipage}[t]{0.485\linewidth}\centering\fontsize{8}{9}\selectfont
    \textbf{Before}\par\smallskip\includegraphics[width=\linewidth,trim=250bp 130bp 250bp 130bp,clip]{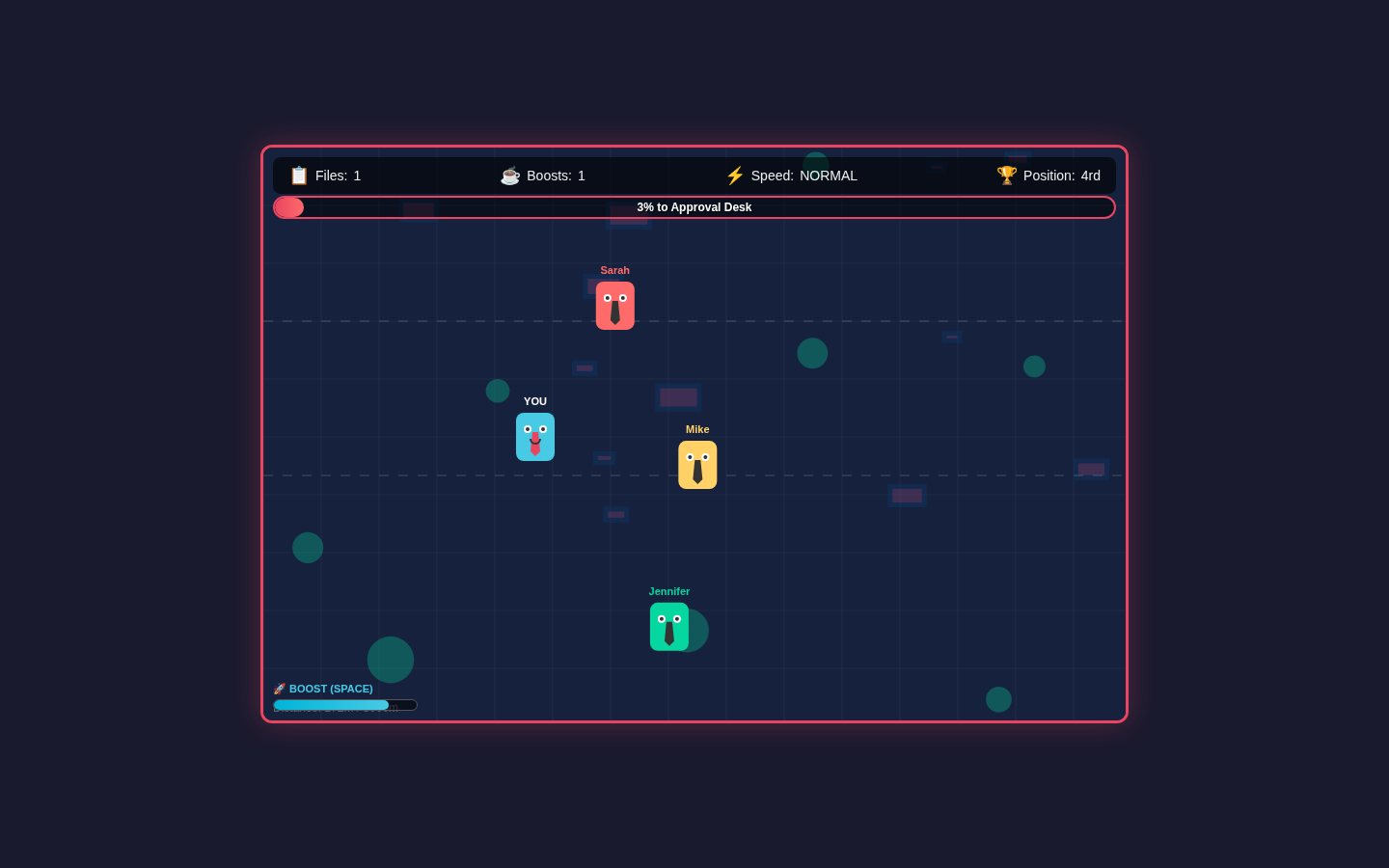}
  \end{minipage}\hfill
  \begin{minipage}[t]{0.485\linewidth}\centering\fontsize{8}{9}\selectfont
    \textbf{After}\par\smallskip\includegraphics[width=\linewidth,trim=250bp 130bp 250bp 130bp,clip]{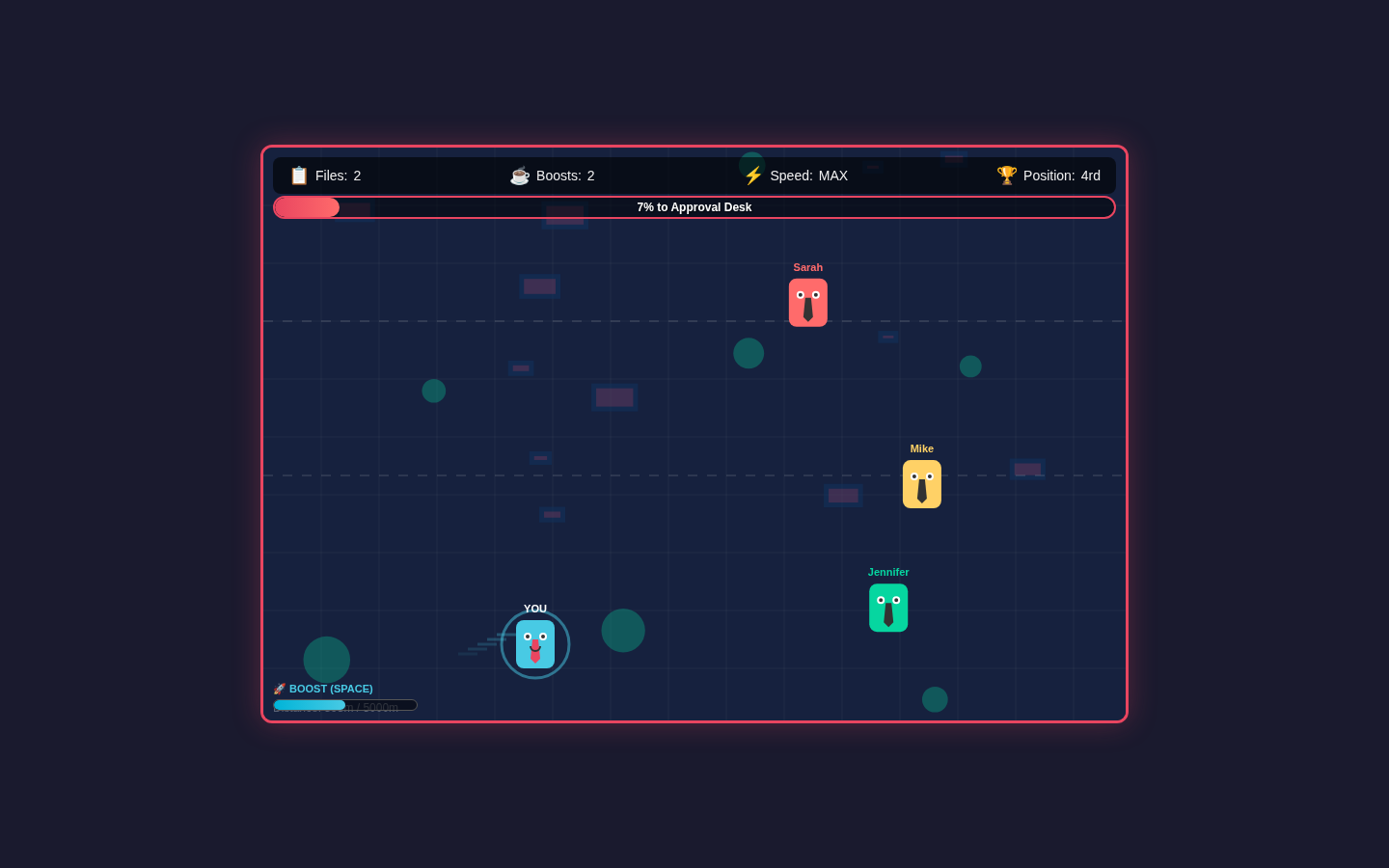}
  \end{minipage}
\end{minipage}\hfill
\begin{minipage}[t]{0.335\textwidth}\vspace{0pt}
  \raggedright\fontsize{8.5}{10}\selectfont 
\textbf{Decision:} Move into the lower lane and boost past approaching paperwork.\par\smallskip
\raisebox{-1pt}{\includegraphics[height=9pt]{appendix_assets/icons/transition-judge.png}}\enspace\textbf{Transition 94/100:} The player moves into the lower lane with an active boost trail. Speed switches to MAX, the meter decreases, and progress advances from 3\% to 7\%.\par\smallskip
\raisebox{-1pt}{\includegraphics[height=9pt]{appendix_assets/icons/screenshot-judge.png}}\enspace\textbf{Screenshot 92/100:} The boosted player, competitors, hazards, progress, and complete HUD remain sharp and stable.\par\smallskip
\textbf{Judge note:} The small boost-meter annotation is faint.

\end{minipage}\par\medskip
\par\noindent\colorbox{blue!6}{\parbox{\dimexpr\textwidth-2\fboxsep\relax}{%
  \footnotesize\bfseries Turn 5\hfill\raisebox{-1pt}{\includegraphics[height=9pt]{appendix_assets/icons/combo.png}}\enspace Right + Space (1 s)}}\par\smallskip
\noindent\begin{minipage}[t]{0.645\textwidth}\vspace{0pt}
  \begin{minipage}[t]{0.485\linewidth}\centering\fontsize{8}{9}\selectfont
    \textbf{Before}\par\smallskip\includegraphics[width=\linewidth,trim=250bp 130bp 250bp 130bp,clip]{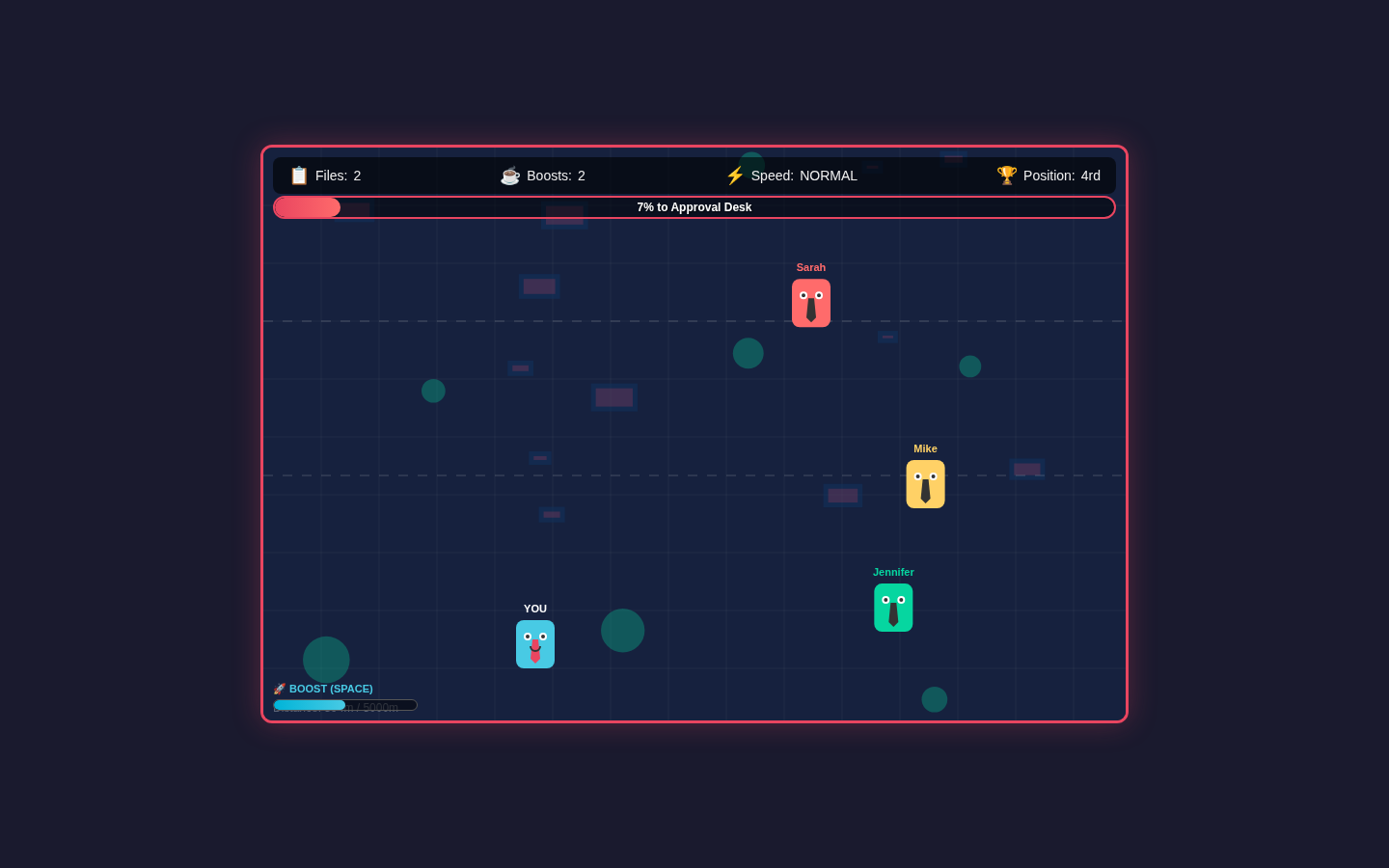}
  \end{minipage}\hfill
  \begin{minipage}[t]{0.485\linewidth}\centering\fontsize{8}{9}\selectfont
    \textbf{After}\par\smallskip\includegraphics[width=\linewidth,trim=250bp 130bp 250bp 130bp,clip]{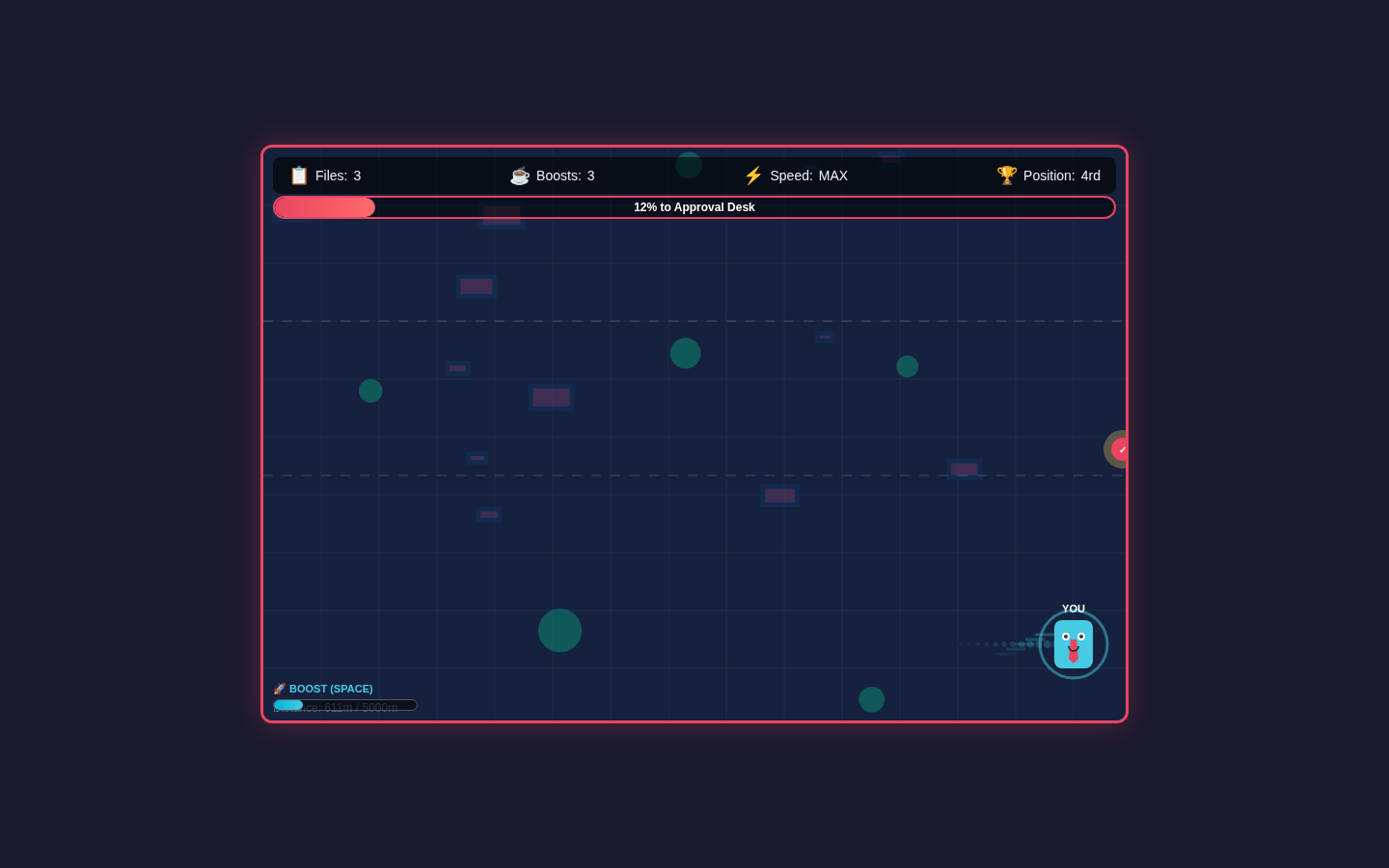}
  \end{minipage}
\end{minipage}\hfill
\begin{minipage}[t]{0.335\textwidth}\vspace{0pt}
  \raggedright\fontsize{8.5}{10}\selectfont 
\textbf{Decision:} Boost right from the lower lane to close the gap on the coworkers.\par\smallskip
\raisebox{-1pt}{\includegraphics[height=9pt]{appendix_assets/icons/transition-judge.png}}\enspace\textbf{Transition 91/100:} The player advances to the lower-right under MAX speed with a visible boost trail. Progress rises from 7\% to 12\%; Files and Boosts each reach 3.\par\smallskip
\raisebox{-1pt}{\includegraphics[height=9pt]{appendix_assets/icons/screenshot-judge.png}}\enspace\textbf{Screenshot 91/100:} The course, player, boost effect, and HUD remain readable.\par\smallskip
\textbf{Judge note:} A far-right object is partially clipped, and resource relationships remain unclear.

\end{minipage}\par\medskip
\par\noindent\colorbox{blue!6}{\parbox{\dimexpr\textwidth-2\fboxsep\relax}{%
  \footnotesize\bfseries Turn 6\hfill\raisebox{-1pt}{\includegraphics[height=9pt]{appendix_assets/icons/wait.png}}\enspace Wait (1 s)}}\par\smallskip
\noindent\begin{minipage}[t]{0.645\textwidth}\vspace{0pt}
  \begin{minipage}[t]{0.485\linewidth}\centering\fontsize{8}{9}\selectfont
    \textbf{Before}\par\smallskip\includegraphics[width=\linewidth,trim=250bp 130bp 250bp 130bp,clip]{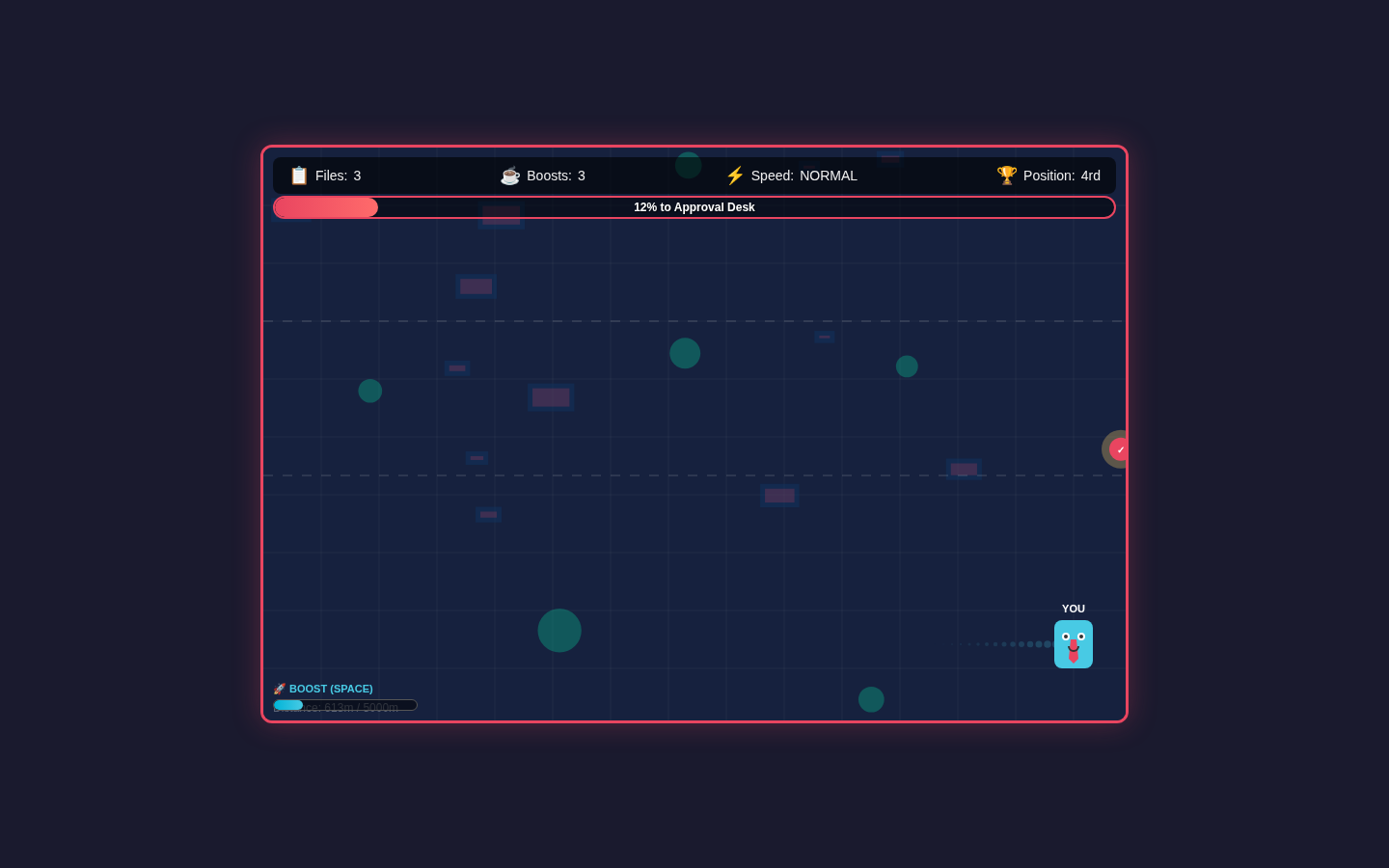}
  \end{minipage}\hfill
  \begin{minipage}[t]{0.485\linewidth}\centering\fontsize{8}{9}\selectfont
    \textbf{After}\par\smallskip\includegraphics[width=\linewidth,trim=250bp 130bp 250bp 130bp,clip]{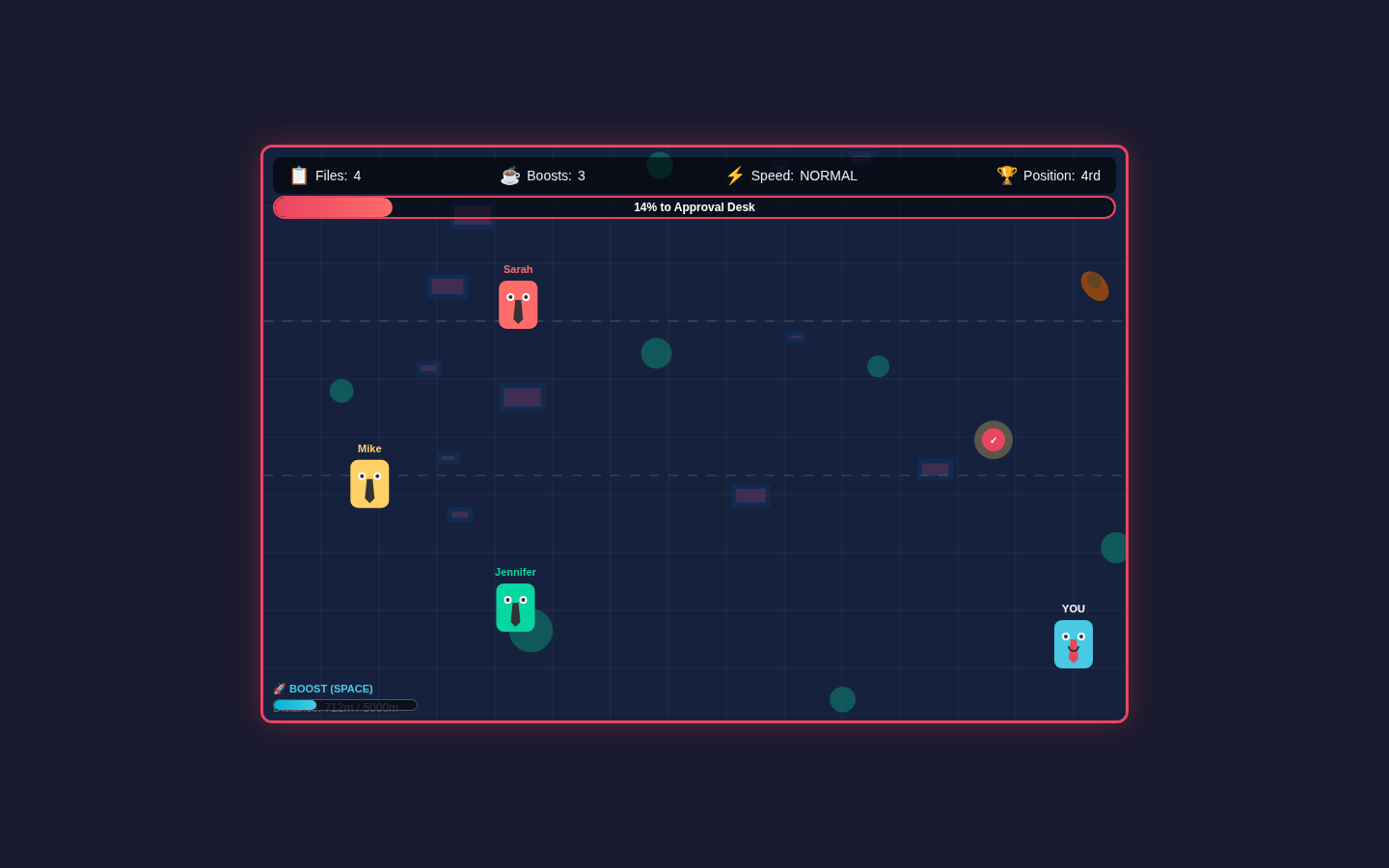}
  \end{minipage}
\end{minipage}\hfill
\begin{minipage}[t]{0.335\textwidth}\vspace{0pt}
  \raggedright\fontsize{8.5}{10}\selectfont 
\textbf{Decision:} Wait in the lower lane for the boost meter to recharge before another burst.\par\smallskip
\raisebox{-1pt}{\includegraphics[height=9pt]{appendix_assets/icons/transition-judge.png}}\enspace\textbf{Transition 92/100:} The race advances from 12\% to 14\% while the player stays in the lower lane. The boost meter refills and Files increases to 4.\par\smallskip
\raisebox{-1pt}{\includegraphics[height=9pt]{appendix_assets/icons/screenshot-judge.png}}\enspace\textbf{Screenshot 91/100:} The three-lane course, racers, and status HUD stay clear; fine meter text and dark field details are harder to read.\par\smallskip
\textbf{Judge note:} The file-count increase during the wait has no explicit pickup feedback.

\end{minipage}\par\medskip
\caption{GUI Assessor trajectory, turns 4--6.}
\label{fig:assessor-trajectory-2}
\end{figure*}

\begin{figure*}[p]
\centering
\par\noindent\colorbox{blue!6}{\parbox{\dimexpr\textwidth-2\fboxsep\relax}{%
  \footnotesize\bfseries Turn 7\hfill\raisebox{-1pt}{\includegraphics[height=9pt]{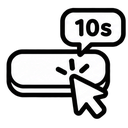}}\enspace Space (1 s)}}\par\smallskip
\noindent\begin{minipage}[t]{0.645\textwidth}\vspace{0pt}
  \begin{minipage}[t]{0.485\linewidth}\centering\fontsize{8}{9}\selectfont
    \textbf{Before}\par\smallskip\includegraphics[width=\linewidth,trim=250bp 130bp 250bp 130bp,clip]{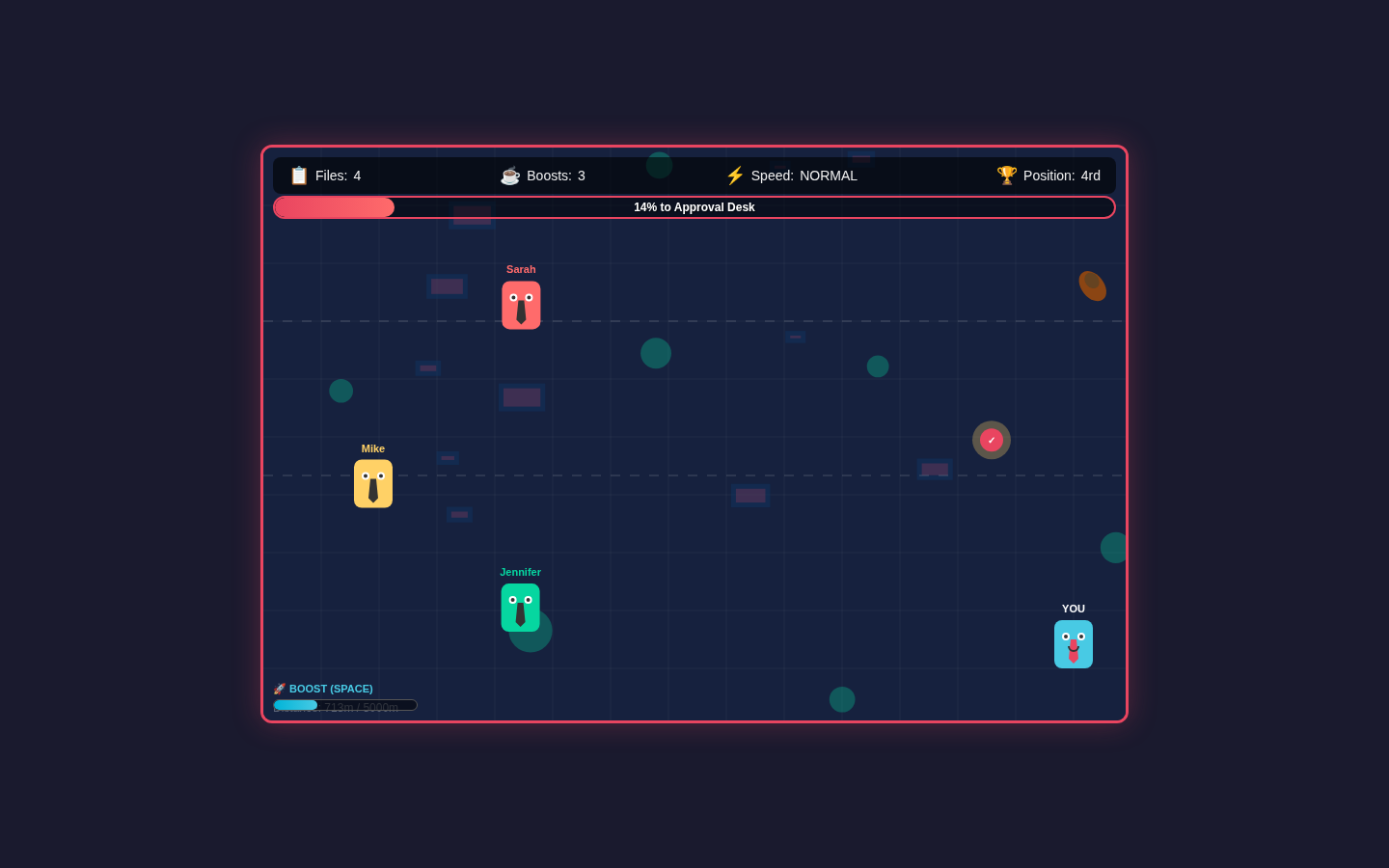}
  \end{minipage}\hfill
  \begin{minipage}[t]{0.485\linewidth}\centering\fontsize{8}{9}\selectfont
    \textbf{After}\par\smallskip\includegraphics[width=\linewidth,trim=250bp 130bp 250bp 130bp,clip]{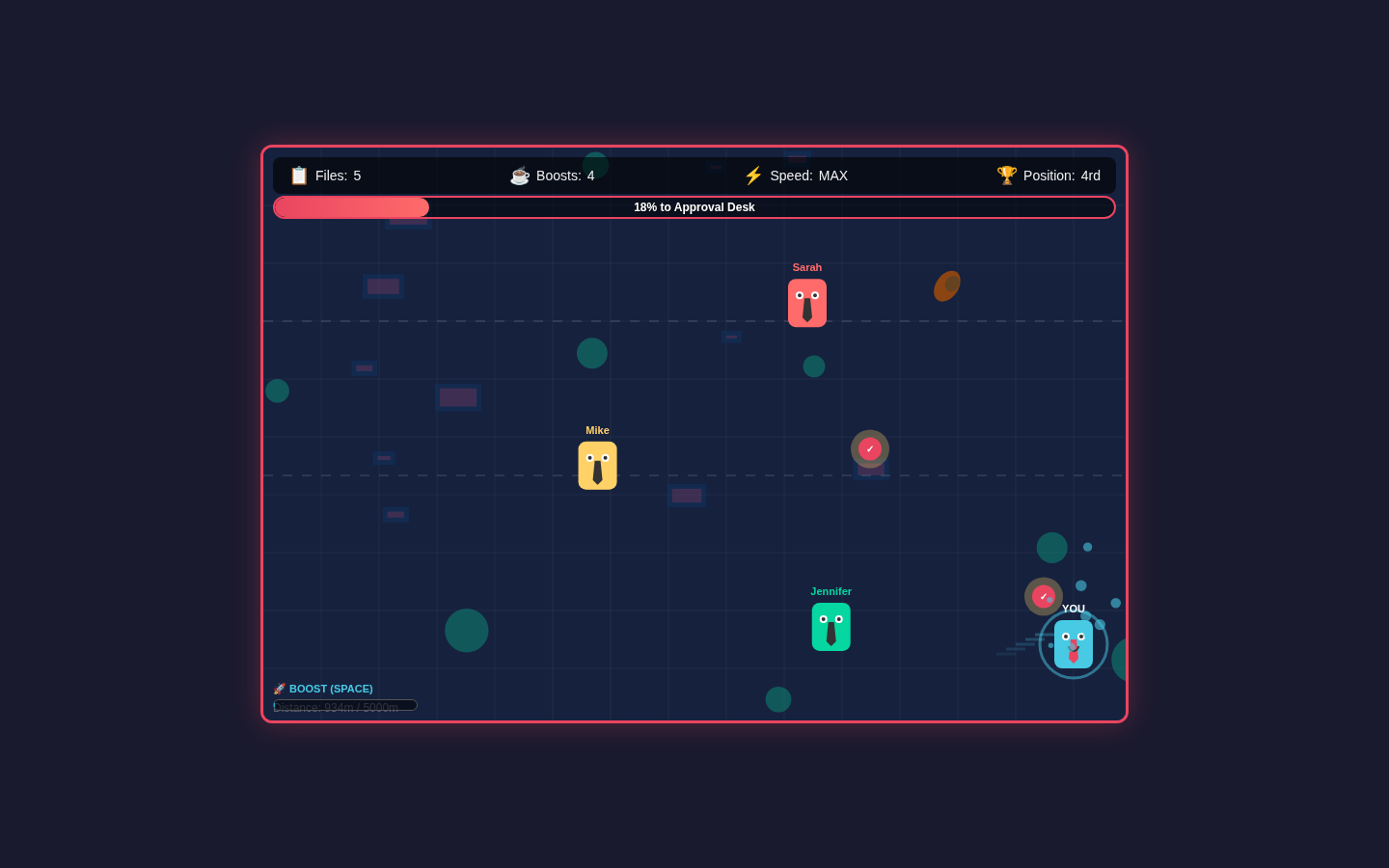}
  \end{minipage}
\end{minipage}\hfill
\begin{minipage}[t]{0.335\textwidth}\vspace{0pt}
  \raggedright\fontsize{8.5}{10}\selectfont 
\textbf{Decision:} Use a sustained boost from the clear lower lane to accelerate race progress.\par\smallskip
\raisebox{-1pt}{\includegraphics[height=9pt]{appendix_assets/icons/transition-judge.png}}\enspace\textbf{Transition 93/100:} Speed switches from NORMAL to MAX with visible acceleration effects. Progress reaches 18\%, Files reaches 5, and Boosts reaches 4 while rank remains fourth.\par\smallskip
\raisebox{-1pt}{\includegraphics[height=9pt]{appendix_assets/icons/screenshot-judge.png}}\enspace\textbf{Screenshot 95/100:} Speed, resources, progress, rank, hazards, and the player boost animation are sharp and easy to identify.\par\smallskip
\textbf{Judge note:} The boost counter increases during boosting without a clear explanation.

\end{minipage}\par\medskip
\par\noindent\colorbox{blue!6}{\parbox{\dimexpr\textwidth-2\fboxsep\relax}{%
  \footnotesize\bfseries Turn 8\hfill\raisebox{-1pt}{\includegraphics[height=9pt]{appendix_assets/icons/press.png}}\enspace Space (1 s)}}\par\smallskip
\noindent\begin{minipage}[t]{0.645\textwidth}\vspace{0pt}
  \begin{minipage}[t]{0.485\linewidth}\centering\fontsize{8}{9}\selectfont
    \textbf{Before}\par\smallskip\includegraphics[width=\linewidth,trim=250bp 130bp 250bp 130bp,clip]{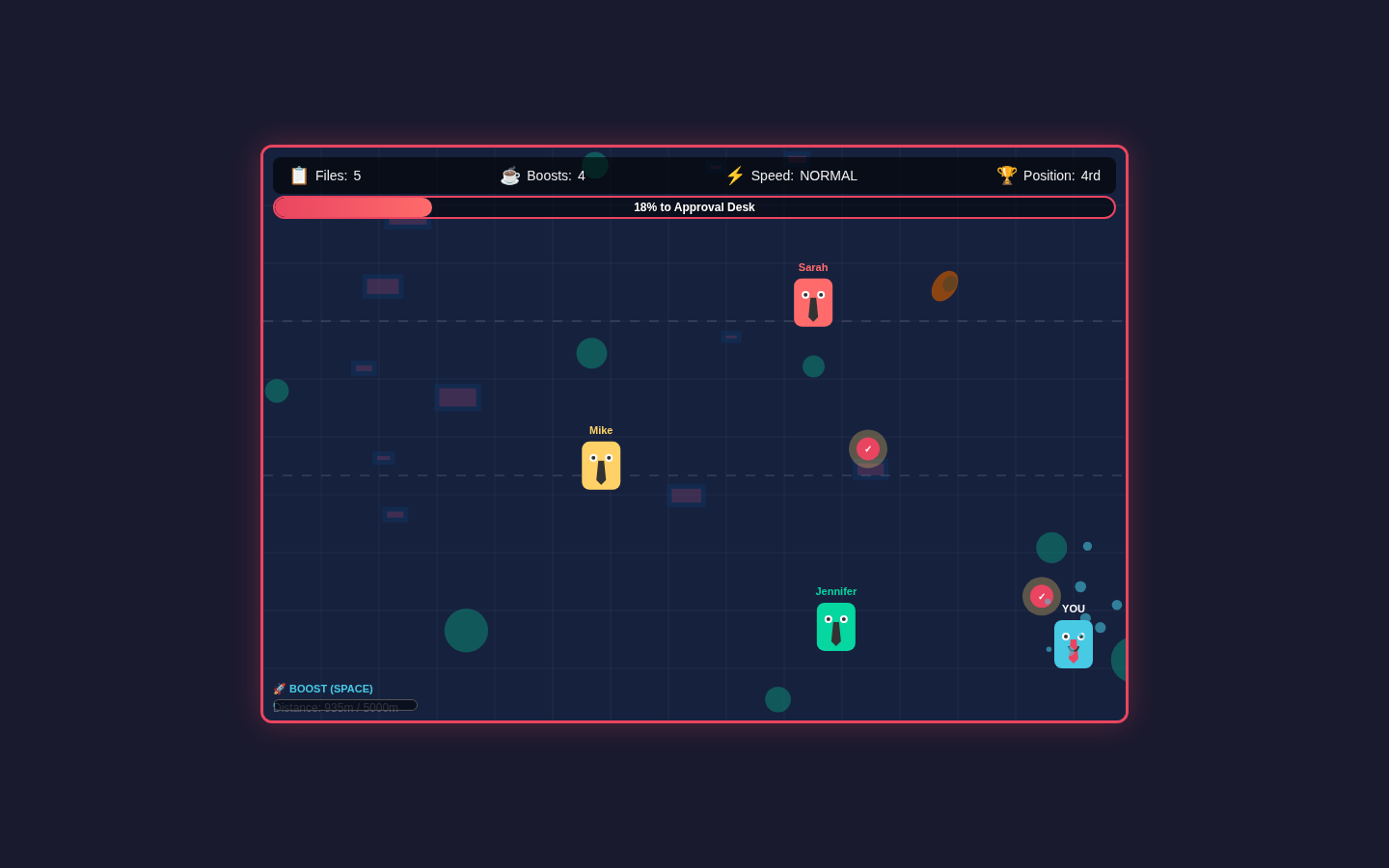}
  \end{minipage}\hfill
  \begin{minipage}[t]{0.485\linewidth}\centering\fontsize{8}{9}\selectfont
    \textbf{After}\par\smallskip\includegraphics[width=\linewidth,trim=250bp 130bp 250bp 130bp,clip]{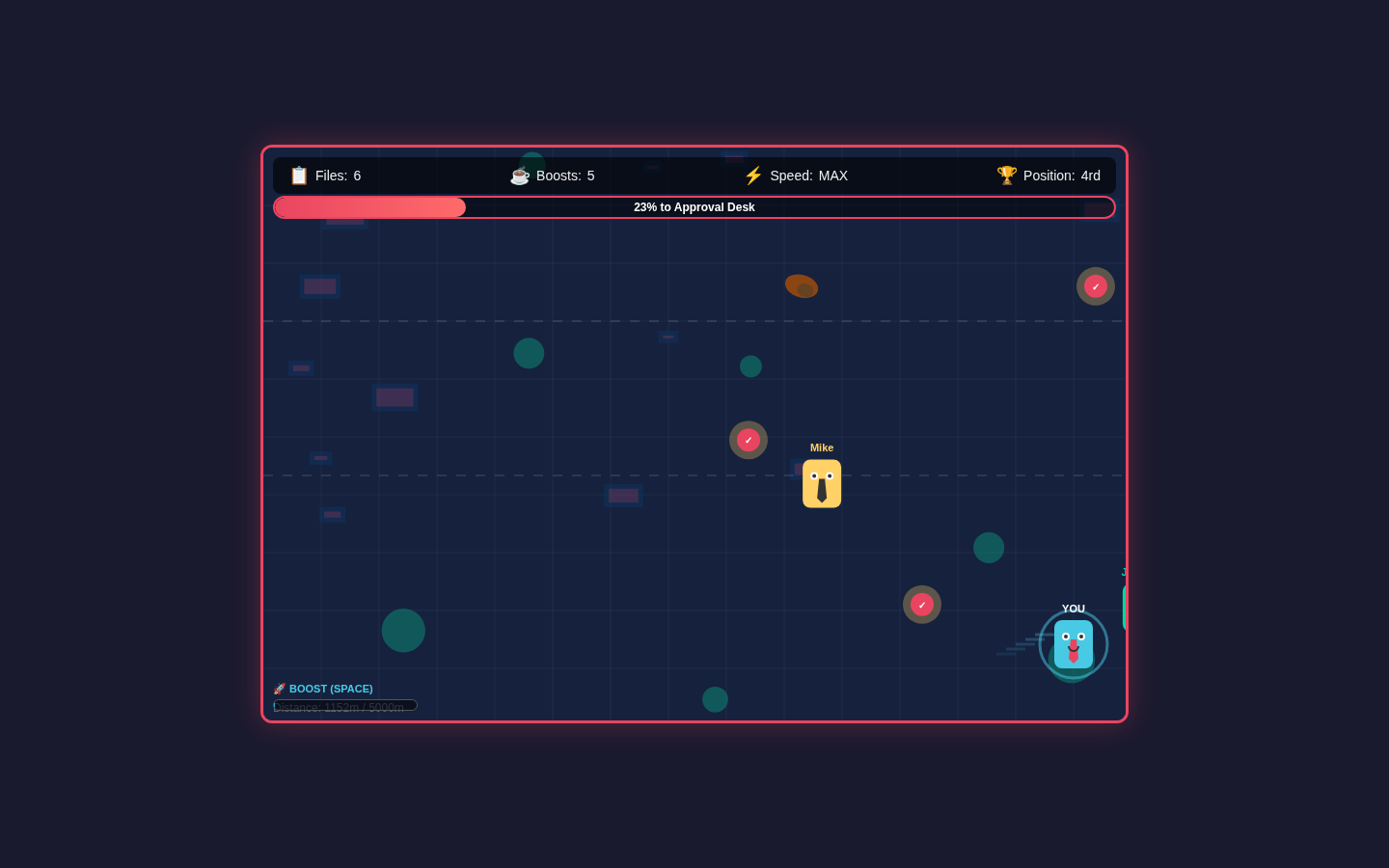}
  \end{minipage}
\end{minipage}\hfill
\begin{minipage}[t]{0.335\textwidth}\vspace{0pt}
  \raggedright\fontsize{8.5}{10}\selectfont 
\textbf{Decision:} At 18\% progress, use another sustained boost from the lower lane to advance.\par\smallskip
\raisebox{-1pt}{\includegraphics[height=9pt]{appendix_assets/icons/transition-judge.png}}\enspace\textbf{Transition 93/100:} A cyan trail accompanies MAX speed. Distance advances from 935 m to 1152 m and progress from 18\% to 23\%; both resource counters rise by one.\par\smallskip
\raisebox{-1pt}{\includegraphics[height=9pt]{appendix_assets/icons/screenshot-judge.png}}\enspace\textbf{Screenshot 93/100:} The HUD exposes resources, speed, rank, and progress; the cyan player and boost trail remain clearly visible.\par\smallskip
\textbf{Judge note:} A rival is partially clipped at the right edge.

\end{minipage}\par\medskip
\noindent\colorbox{blue!6}{\parbox{\dimexpr\textwidth-2\fboxsep\relax}{\footnotesize\bfseries
\raisebox{-1pt}{\includegraphics[height=9pt]{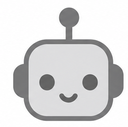}}\enspace Meta-Level Judge}}\par\smallskip
\begin{minipage}{\textwidth}\fontsize{9}{11}\selectfont
Steering and boosting produce visible speed, trail, and HUD changes. The episode reaches 23\% progress. The judge identifies unclear resource counters, small obstacle details, and edge clipping.
\par\medskip
\centering\footnotesize
\setlength{\tabcolsep}{4pt}\renewcommand{\arraystretch}{1.12}
\begin{tabular}{@{}ccccccc@{}}
\toprule
\makecell{Visual\\stability} & \makecell{Screenshot\\aesthetics} & \makecell{Feedback\\clarity} & \makecell{Interaction\\reliability} & \makecell{State\\legibility} & \makecell{Prompt\\alignment} & \makecell{Task\\progress} \\
\midrule
95 & 86.38 & 92.38 & 100 & 93.5 & 92 & 72 \\
\bottomrule
\end{tabular}
\end{minipage}\par\smallskip
\caption{GUI Assessor trajectory with the Meta-Level assessment and its dimension scores.}
\label{fig:assessor-trajectory-3}
\end{figure*}

\clearpage
\end{document}